\documentclass[10pt,twocolumn,letterpaper]{article}

\usepackage{iccv}              % To produce the CAMERA-READY version
\usepackage{graphicx}
\usepackage{amsmath}
\usepackage{amssymb}
\usepackage{booktabs}
\usepackage{animate}
\usepackage{enumitem}
\usepackage{multirow}
\usepackage{diagbox}
\usepackage{amsmath}
\usepackage{accents}
\usepackage{graphicx}
\usepackage{amsmath}
\usepackage{amssymb}
\usepackage{booktabs}
\usepackage{float}
\usepackage{diagbox}
\usepackage{bbding}
\usepackage{utfsym}
\usepackage{pifont}
\usepackage{multirow}
\usepackage{enumitem}
\usepackage{tabularx}
\usepackage{microtype}
\usepackage{soul}
\usepackage{threeparttable}
\usepackage{MnSymbol}
\usepackage{algorithm}
\usepackage{listings}
\usepackage{xcolor}

\makeatletter
\@namedef{ver@everyshi.sty}{}
\makeatother
\usepackage{tikz}
\usepackage{makecell}
\usepackage{wrapfig}
\usepackage{xcolor,colortbl}
\usepackage{color}
\usepackage[accsupp]{axessibility}  % Improves PDF readability for those with disabilities.
\definecolor{iccvblue}{rgb}{0.21,0.49,0.74}
\usepackage[pagebackref,breaklinks,colorlinks,allcolors=iccvblue]{hyperref}

\usepackage[capitalize]{cleveref}
\crefname{section}{Sec.}{Secs.}
\Crefname{section}{Section}{Sections}
\Crefname{table}{Table}{Tables}
\crefname{table}{Tab.}{Tabs.}

\def\etal{\textit{et al}.}

\newcommand{\rev}[1]{\textcolor{black}{#1}}

\def\paperID{2808} % *** Enter the Paper ID here
\def\confName{ICCV}
\def\confYear{2025}

\definecolor{Gray}{gray}{0.5}
\definecolor{LightCyan}{rgb}{0.88,1,1}

\newcolumntype{a}{>{\columncolor{Gray}}c}
\newcolumntype{b}{>{\columncolor{white}}c}

\newlength{\dhatheight}
\newcommand{\doublehat}[1]{%
    \settoheight{\dhatheight}{\ensuremath{\hat{#1}}}%
    \addtolength{\dhatheight}{-0.35ex}%
    \hat{\vphantom{\rule{1pt}{\dhatheight}}%
    \smash{\hat{#1}}}}

\begin{document}

\title{Cross-Subject Mind Decoding from Inaccurate Representations}

\author{Yangyang Xu$^{1*}$,
       Bangzhen Liu$^2$,
       Wenqi Shao$^3$,
       Yong Du$^4$\thanks{Corresponding authors: Yangyang Xu (xuyangyang@hit.edu.cn) and Yong Du (csyongdu@ouc.edu.cn).},
       Shengfeng He$^5$,
       Tingting Zhu$^{1}$%\thanks{Corresponding author (xuyangyang@hit.edu.cn).}
       \\
       \normalsize{$^1$The University of Oxford \hspace{5mm} $^2$South China University of Technology \hspace{5mm} $^3$Shanghai AI Lab} \\ 
       \normalsize{$^4$Ocean University of China \hspace{5mm} $^5$Singapore Management University \hspace{5mm} \vspace{-1mm}}}

% \textsl{\textbf{Yangyang Xu}, Bangzhen Liu, Wenqi Shao, Yong Du, Shengfeng He, and Tingting Zhu}  \\

\teaser{%\vspace{5mm}
    \centering
        \captionsetup[subfloat]{justification=centering}
     \hspace{-2.25mm}
     \subfloat[BrainDiffuser]{\label{x}
     \begin{minipage}{0.12\linewidth}
     \includegraphics[width=\linewidth]{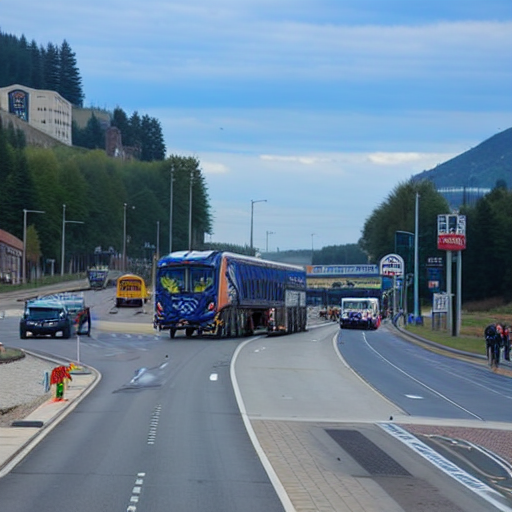}
     \includegraphics[width=\linewidth]{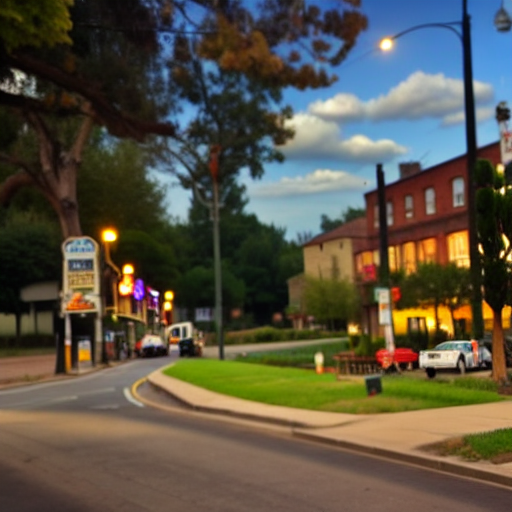}
     \end{minipage}
     }
     \hspace{-2.25mm}
     \subfloat[MindEye1]{\label{xx}
     \begin{minipage}{0.12\linewidth}
     \includegraphics[width=\linewidth]{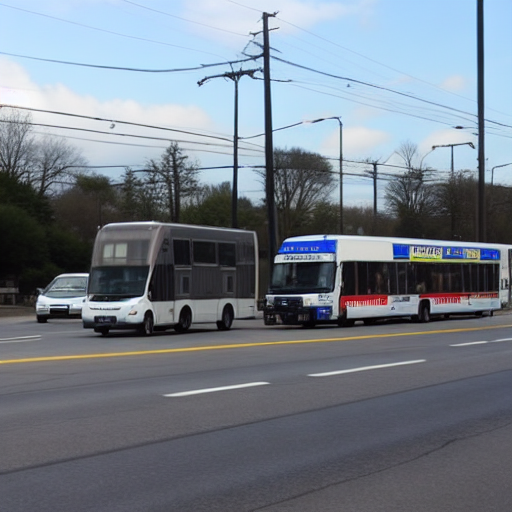}
     \includegraphics[width=\linewidth]{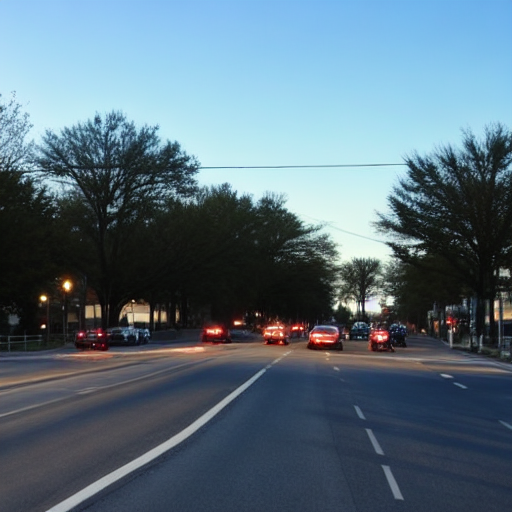}
     \end{minipage}
     }
     \hspace{-2.25mm}
     \subfloat[MindBridge]{\label{xxx}
     \begin{minipage}{0.12\linewidth}
     \includegraphics[width=\linewidth]{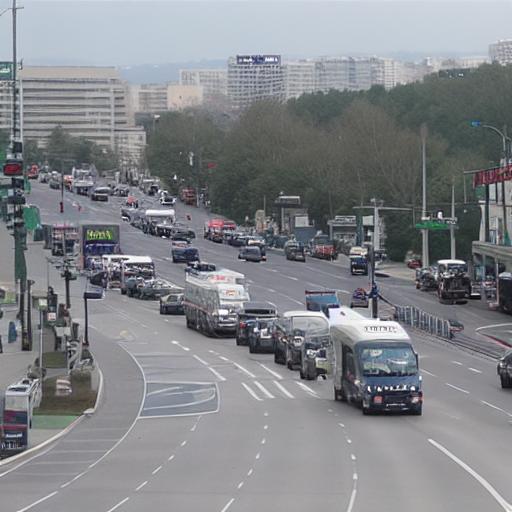}
     \includegraphics[width=\linewidth]{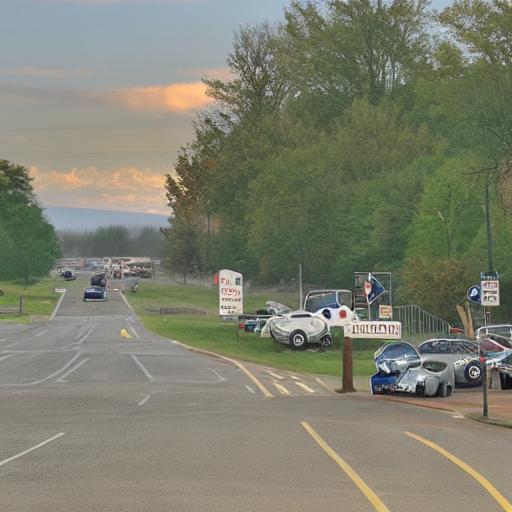}
     \end{minipage}
     }
     \hspace{-2.25mm}
     \subfloat[MindEye2]{\label{xxxx}
     \begin{minipage}{0.12\linewidth}
     \includegraphics[width=\linewidth]{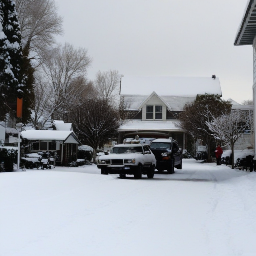}
     \includegraphics[width=\linewidth]{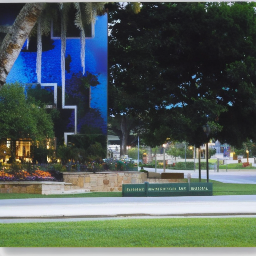}
     \end{minipage}
     }
     \hspace{-2.25mm}
     \subfloat[Neuropictor]{\label{x1}
     \begin{minipage}{0.12\linewidth}
     \includegraphics[width=\linewidth]{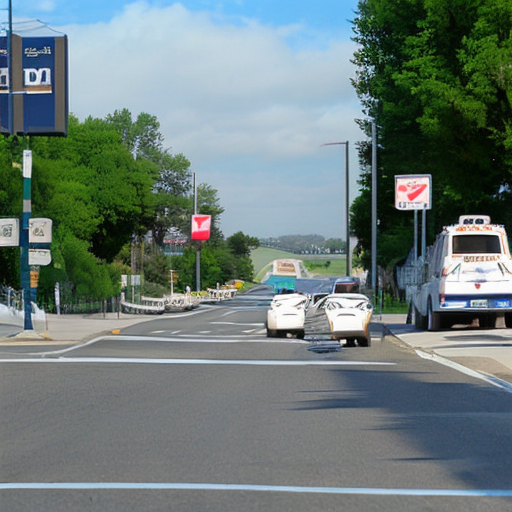}
     \includegraphics[width=\linewidth]{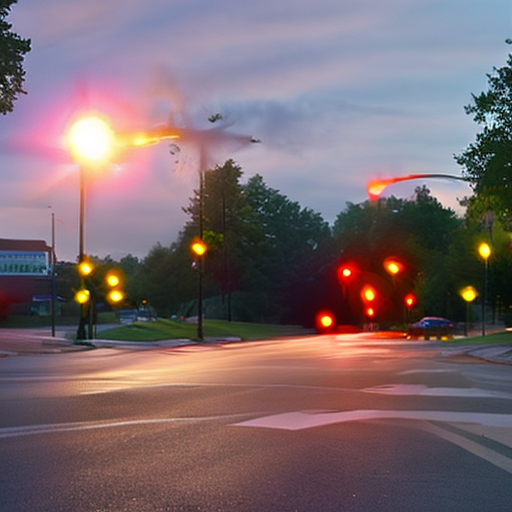}
     \end{minipage}
     }
     \hspace{-2.25mm}
     \subfloat[Direct Decoding]{\label{oddlabel}
     \begin{minipage}{0.12\linewidth}
     \includegraphics[width=\linewidth]{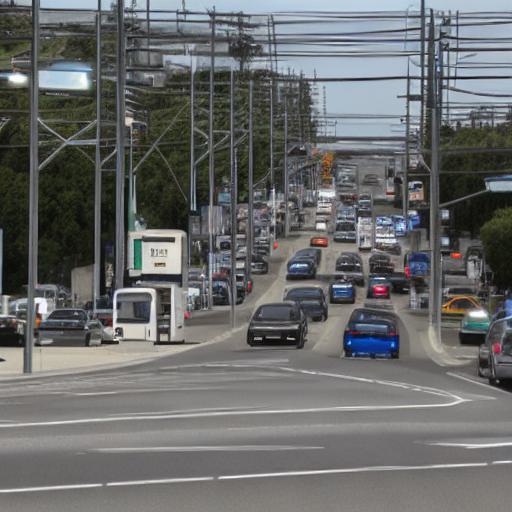}
     \includegraphics[width=\linewidth]{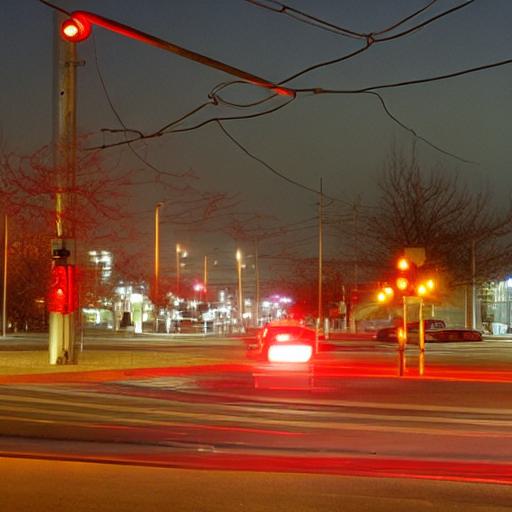}
     \end{minipage}
     }
     \hspace{-2.25mm}
     \subfloat[Ours]{\label{xxxxxaax}
     \begin{minipage}{0.12\linewidth}
     \includegraphics[width=\linewidth]{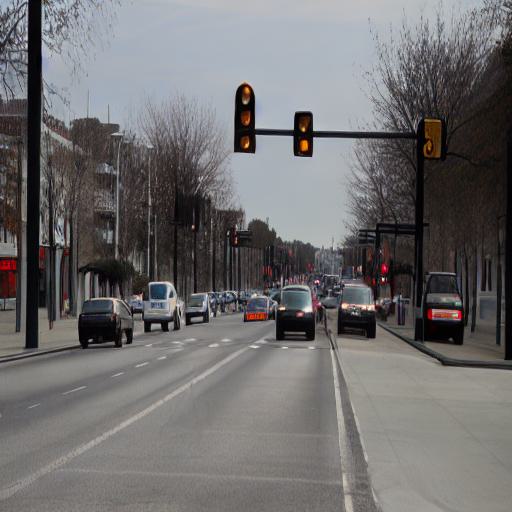}
     \includegraphics[width=\linewidth]{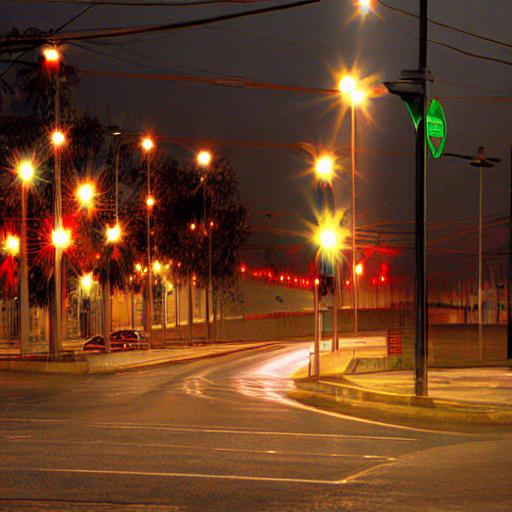}
     \end{minipage}
     }
     \hspace{-2.25mm}
     \subfloat[Stimulus]{\label{xxxxxx}
     \begin{minipage}{0.12\linewidth}
     \includegraphics[width=\linewidth]{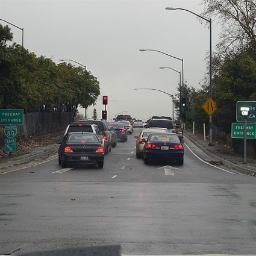}
     \includegraphics[width=\linewidth]{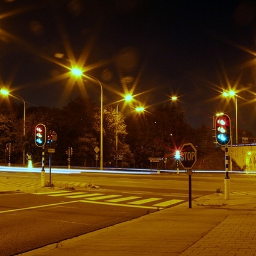}
     \end{minipage}
     }
% \vspace{-2mm}
\caption{We propose a framework for faithful cross-subject mind decoding. Unlike prior approaches that struggle with cross-subject generalization and accumulate errors in representation prediction, our method ensures more accurate and high-fidelity image reconstruction from fMRI signals.}
\label{fig:teaser}
\vspace{-4mm}
}

\maketitle

%%%%%%%%% ABSTRACT
\begin{abstract}
% Decoding stimulus images from fMRI signals has advanced with pre-trained generative models, yet existing methods struggle with cross-subject mappings due to cognitive variability and subject-specific differences. This challenge stems from sequential errors, where unidirectional mappings produce partially inaccurate representations that, when fed into diffusion models, compound errors and degrade reconstruction fidelity.
% To address this, we propose the Bidirectional Autoencoder Intertwining framework for the accurate mind representation prediction. Our approach unifies multiple subjects through the Subject Bias Modulation Module while leveraging bidirectional mapping to better capture the data distribution for accurate representation prediction. To further improve fidelity when decoding the representations to stimulus images, we introduce a Semantic Refinement Module to enhance semantic representations and a Visual Coherence Module to mitigate the impact of inaccurate visual representations. Integrated with ControlNet and Stable Diffusion, our method surpasses state-of-the-art approaches on benchmark datasets in both qualitative and quantitative evaluations. Moreover, our framework demonstrates strong adaptability to new subjects with minimal training samples.

Decoding stimulus images from fMRI signals has advanced with pre-trained generative models. However, existing methods struggle with cross-subject mappings due to cognitive variability and subject-specific differences. This challenge arises from sequential errors, where unidirectional mappings generate partially inaccurate representations that, when fed into diffusion models, accumulate errors and degrade reconstruction fidelity.
To address this, we propose the Bidirectional Autoencoder Intertwining framework for accurate decoded representation prediction. Our approach unifies multiple subjects through a Subject Bias Modulation Module while leveraging bidirectional mapping to better capture data distributions for precise representation prediction. To further enhance fidelity when decoding representations into stimulus images, we introduce a Semantic Refinement Module to improve semantic representations and a Visual Coherence Module to mitigate the effects of inaccurate visual representations. Integrated with ControlNet and Stable Diffusion, our method outperforms state-of-the-art approaches on benchmark datasets in both qualitative and quantitative evaluations. Moreover, our framework exhibits strong adaptability to new subjects with minimal training samples.

\end{abstract}

\section{Introduction}
\label{secintro}

The human visual cortex processes sensory stimuli and encodes them into brain signals, playing a fundamental role in shaping perceptual experience~\cite{sergent1992functional,kanwisher2002fusiform,jain2023selectivity}. Decoding these brain signals back into the original stimuli has become a significant focus in neuroscience and computer science, presenting a challenging inverse problem with potential applications in brain-computer interfaces (BCIs) and cognitive science~\cite{du2022fmri}. The functional Magnetic Resonance Imaging (fMRI), which captures changes in blood oxygenation, is widely used in BCIs for mind decoding due to its ability to reflect dynamic brain activity patterns~\cite{chen2023seeing,ozcelik2023natural,takagi2023high,lu2023minddiffuser,scotti2024reconstructing}.

Early methods for decoding fMRI data map voxel signals to the feature space of pre-trained Convolutional Neural Networks (CNNs) to classify object categories~\cite{horikawa2017generic}. However, these methods were limited in reconstructing complex visual stimuli. To improve the realism of reconstructions, researchers introduced Generative Adversarial Networks (GANs)~\cite{brock2018large,karras2020analyzing,zhou2022towards} and diffusion models~\cite{rombach2022high,xu2023versatile} for mind decoding, where fMRI data is typically mapped to image representations within generative models to guide image synthesis.

Despite these advances, current approaches are limited by individual cognitive variability and often rely on subject-specific models that require separate training for each individual~\cite{wang2024mindbridge,scotti2024mindeye2}. Such subject-specific models generally lack generalizability across individuals, reducing their scalability and applicability. Recently, cross-subject mind decoding frameworks~\cite{scotti2024mindeye2,wang2024mindbridge,quan2024psychometry} have attempted to map fMRI voxels to shared representations across subjects. However, these frameworks face two sequential sources of error: \romannumeral1) unidirectional mappings often fail to capture the complex variability across subjects, producing partially inaccurate representations, and \romannumeral2) these inaccurate representations are subsequently fed into pre-trained diffusion models without adjustments for errors, leading to compounded inaccuracies and frequently resulting in reconstructions with low fidelity and unrealistic details. As shown in Fig.~\ref{fig:teaser}, prior works fail to decode the night street scene.

To achieve high-fidelity mind decoding, we argue that both sources of inaccurate must be addressed. First, enhancing the accuracy of image representations during the initial mapping stage is essential to minimize inaccuracies. Second, the framework must be resilient to inaccurate in image representations during downstream processing to prevent error propagation and ensure that final reconstructions maintain high fidelity.

To address these challenges, we first propose a Bidirectional Autoencoder Intertwining (BAI) framework, which learns a bidirectional mapping between fMRI voxels and semantic/visual representations, capturing complex cross-subject relationships between these domains. By intertwining transformations in both directions, it not only improves fidelity in fMRI-to-image decoding but also enables the synthesis of fMRI-like data from semantic/visual inputs, yielding more reliable representations. However, as shown in Fig.~\ref{oddlabel}, decoding the representations predicted by BAI directly cannot guarantee the fidelity reconstruction, as it ignores the error in second step. To support inaccurate-tolerant decoding, we introduce the Semantic Refinement Module (SRM) and Visual Coherence Module (VCM), which mitigate the impact of representation errors on image reconstruction. Specifically, BAI employs two intertwined autoencoders for fMRI voxels and representations, achieving bidirectional mapping by swapping decoders. To reduce subject-specific biases, we incorporate a Subject Bias Modulation Module (SBMM) in the fMRI autoencoder, which applies statistical modulation. The Semantic Refinement Module refines the predicted semantic embedding, while the Visual Coherence Module optimally integrates visual representations to ensure output fidelity~\cite{liu2024smartcontrol}. Combined with ControlNet~\cite{zhang2023adding}, these modules reduce dependence on precise representations, preserving both semantic and visual consistency with the original stimuli. We evaluate our approach on the Natural Scenes Dataset~\cite{allen2022massive}, and results demonstrate that our framework outperforms state-of-the-art methods. Furthermore, our framework adapts effectively to new subjects with minimal additional samples.

In summary, our contributions are threefold: 
\begin{itemize}[leftmargin=*,nosep] 
\item We propose a cross-subject mind decoding framework that learns bidirectional mappings between fMRI voxels and semantic/visual representations, capturing complex cross-subject relationships between these domains. 
\item We design Semantic Refinement and Visual Coherence modules to enhance reconstruction accuracy and consistency, reducing dependence on exact representations for high-fidelity mind decoding. 
\item Extensive experiments demonstrate that our framework significantly outperforms state-of-the-art methods in cross-subject mind decoding and adapts effectively to new subjects with minimal additional samples. 
\end{itemize}

\section{Related Works}
\label{sec:related}

\begin{figure*}
\centering
\includegraphics[width=0.949\textwidth]{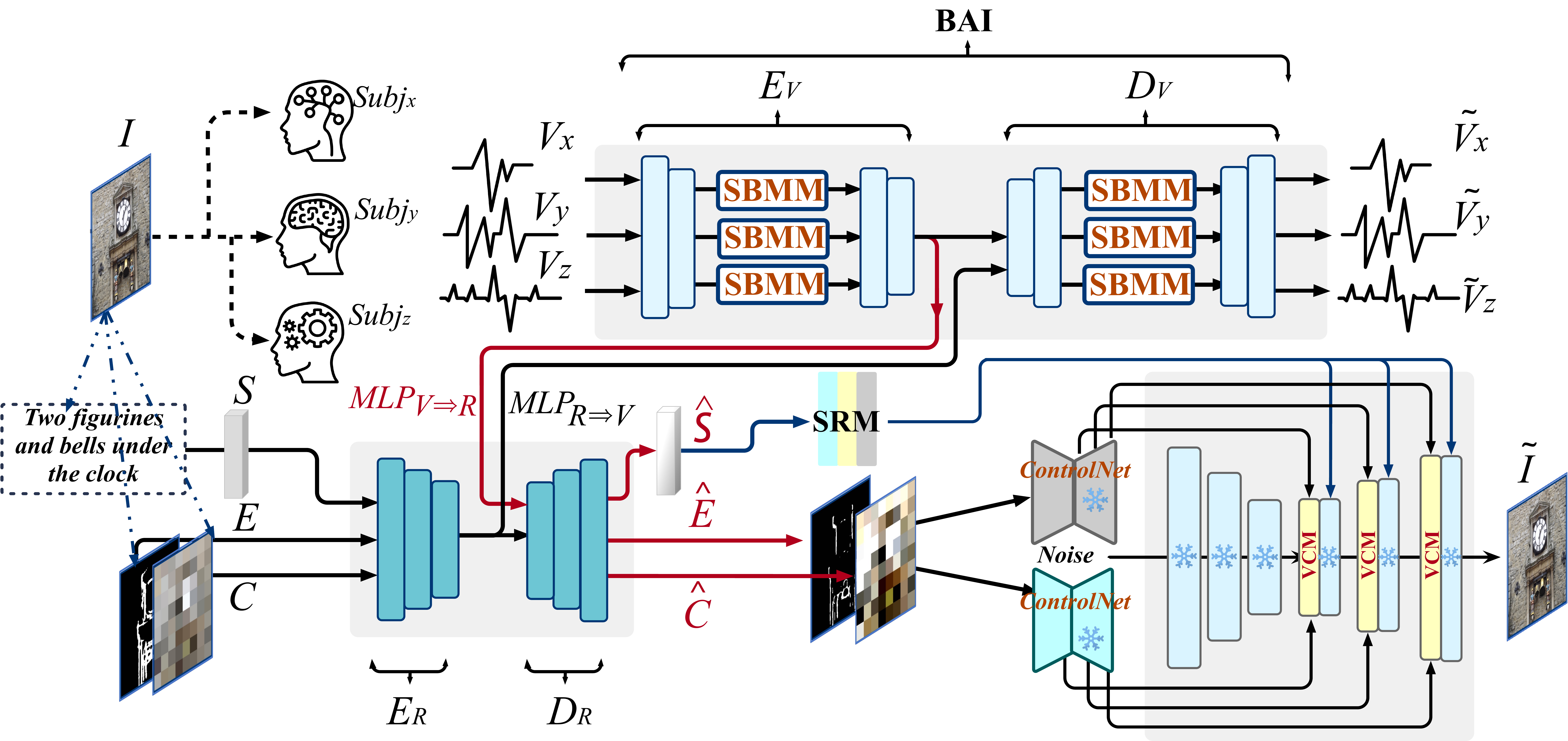}
\vspace{-2mm}
\caption{Overview of our framework. It consists of two autoencoders: one for fMRI voxels \( V_{x} \) and another for image representations (\( S \), \( E \), and \( C \)), supporting both reconstruction and bidirectional translation by swapping decoders. The Subject Bias Modulation Module (SBMM) is integrated into the fMRI encoder \( \mathcal{E}_V \) and decoder \( \mathcal{D}_V \) to eliminate inter-subject variance. Red lines illustrate the translation pipeline from fMRI voxels to image representations. To enhance reconstruction accuracy from inaccurate representations, we introduce the Semantic Refinement Module (SRM) and Visual Coherence Module (VCM) within the frozen ControlNet and Stable Diffusion (SD) components, reducing dependency on precise representations. Finally, the stimulus image \( \tilde{I} \) is reconstructed using the DDIM sampler.}
\label{fig:overview}
\vspace{-4mm}
\end{figure*}

\textbf{\quad Brain Decoding.}
Brain decoding aims to reconstruct stimuli from brain signals~\cite{xia2024umbrae,lahner2024modeling,Brainflex}. Early studies demonstrate that coarse visual information could be decoded from fMRI~\cite{thirion2006inverse,cox2003functional,kamitani2005decoding}. With advancements in deep learning, Tomoyasu~\etal~\cite{horikawa2017generic} map fMRI signals to CNN features. Recently, pre-trained generative models have shown powerful generative capabilities, and several studies leverage these models for brain decoding. Furkan~\etal~\cite{ozcelik2022reconstruction} and Milad~\etal~\cite{mozafari2020reconstructing} extract representative features from fMRI and fine-tuned pre-trained BigGAN\cite{brock2018large} for stimuli reconstruction. MindReader~\cite{lin2022mind} maps fMRI signals to CLIP embeddings~\cite{radford2021learning}, then decoded stimuli images using conditional StyleGAN2~\cite{karras2020analyzing,zhou2022towards}. Recent works introduce diffusion models~\cite{rombach2022high,xu2023versatile} for mind decoding by mapping fMRI signals to intermediate representations~\cite{lu2023minddiffuser,takagi2023high,chen2023seeing,scotti2024reconstructing,xia2024dream}. While promising results have been obtained, these approaches typically require separate model training for different subjects, limiting their broader applicability. Some recent works~\cite{wang2024mindbridge,scotti2024mindeye2,quan2024psychometry} have introduced cross-subject frameworks that unify different subjects within a single model. However, these approaches often suffer from inaccurate predicted representations due to the unidirectional mappings.

\textbf{Diffusion Models in Brain Decoding.}
Diffusion models~\cite{rombach2022high,saharia2022photorealistic,ramesh2022hierarchical,xu2023versatile,ren2024ultrapixel,ren2025turbo2k} have made significant progress in generating diverse and realistic images and videos. Many studies leverage the generative power of diffusion models for brain decoding. Specifically, versatile diffusion~\cite{xu2023versatile} unifies text-to-image and image-to-text synthesis within a single framework, and various works~\cite{wang2024mindbridge,scotti2024reconstructing,scotti2024mindeye2} employ versatile diffusion as a visual decoder by mapping fMRI signals into embeddings. Stable Diffusion (SD)\cite{rombach2022high} performs denoising in latent space, enabling high-quality text-to-image synthesis. Takagi~\etal~\cite{takagi2023high} map fMRI signals into SD’s latent representation for stimuli reconstruction but produces blurry results. MindVis~\cite{chen2023seeing} learns discriminative features from fMRI and projects them into two conditions, controlling the generation process of SD via a cross-attention mechanism. DREAM~\cite{xia2024dream} decodes semantics, depth, and color conditions from fMRI and guides the output of SD with T2I-adapter~\cite{mou2024t2i}. However, all these works overlook the impact of inaccurate representations on reconstruction fidelity. In this paper, we learn bidirectional mappings between fMRI voxels and semantic/visual representations, enabling more accurate representation predictions. Additionally, we introduce two modules to reduce dependency on exact representations, further enhancing fidelity in reconstruction.

\textbf{Controllable Diffusion Models.}
To better leverage the generative capabilities of diffusion models, various works~\cite{zhu2025stable,zhang2023adding,zhao2024uni,xu2024dreamanime,yu2024beyond,liu2024drag} enhance the controllability of pre-trained diffusion models. ControlNet~\cite{zhang2023adding} introduces zero-initialized layers to control the generation process under conditions, such as pose, edge, depth, and more. T2i-adapter~\cite{mou2024t2i} incorporates multiple adapters for pre-trained SD. Uni-ControlNet~\cite{zhao2024uni}, UniControl~\cite{qin2023unicontrol}, and ControlNet++\cite{controlnetpp} unify various control conditions within a single framework, ensuring that output images strictly adhere to multiply conditions. LooseControl\cite{bhat2024loosecontrol} introduces rougher conditions to foster greater creative flexibility, while SmartControl~\cite{liu2024smartcontrol} analyzes ControlNet’s control mechanisms and proposes a module that resolves conflicts between text prompts and multiple control conditions. In our work, we aim to reconstruct realistic and faithful stimuli images despite the presence of both inaccurate semantic and visual conditions.

\section{Methodology}
\label{sec:tsn}

Given fMRI voxels \( V_{x} \in \mathbb{R}^d \) collected from subject \( x \) viewing a stimulus image \( I \in \mathbb{R}^{h \times w \times 3} \), our goal is to develop a model that reconstructs the visual stimulus image \( \hat{I} \) from \( V_{x} \), independent of the specific subject. We follow the pipeline~\cite{xia2024dream,ozcelik2023natural,scotti2024reconstructing,wang2024mindbridge} that map fMRI voxels to the representations of an image, and decode the representations using a pre-trained diffusion model. Our framework improves the reconstruction fidelity by improving the precision of mapped representations in the first stage, and the tolerance to inaccurate representations in next decoding stage. Inspired by DREAM~\cite{xia2024dream}, we divide a natural image into three representations: semantic embedding $S$ for high-level information, edge map $E$ for structure, and color palette $C$ for low-level appearance. The overview of our method is shown in Fig.~\ref{fig:overview}, we first introduce a bidirectional mapping between fMRI voxels and semantic/visual representations, and then introduce two modules that handle inaccurate representations. Finally, the stimulus image is reconstructed with ControlNet and SD.

\subsection{Bidirectional Autoencoder Intertwining}

We first predict three representations from fMRI voxels using the bidirectional autoencoder intertwining framework, which supports bidirectional mapping by intertwining transformations in both directions. In contrast to the widely used unidirectional mapping, bidirectional mapping encourages the model to capture complex relationships between fMRI voxels and image representations, resulting in more accurate predictions. Additionally, it introduces unsupervised cycle consistency between the two domains, providing additional supervision~\cite{zhu2017unpaired}. Specifically, our framework consists of two autoencoders: one for fMRI voxels and one for image representations. It supports both translation and reconstruction between the two domains. The reconstruction pipelines are represented as \( V_{x} \Rightarrow \tilde{V}_{x} \), and \( \{S, E, C\} \Rightarrow \{\tilde{S}, \tilde{E}, \tilde{C}\} \), where \( \{\tilde{\mathbf{\cdot}}\} \) denotes the reconstructed results. The translation pipelines are represented as \( V_{x} \Rightarrow \{\hat{S}, \hat{E}, \hat{C}\} \) and \( \{S, E, C\} \Rightarrow \hat{V}_{x} \), where \( \{\hat{\mathbf{\cdot}}\} \) denotes the predicted results.

\textbf{Reconstruction.} Given fMRI voxels \( V_{x} \) from subject \( x \), they are encoded into a latent space shared across different subjects~\cite{wang2024mindbridge} by the fMRI encoder \( \mathcal{E}_{V} \). Features in this shared latent space are subject-invariant, meaning they can be used not only for voxel reconstruction but also for translation to image representations. The reconstruction is performed by decoding the features with the decoder \( \mathcal{D}_{V}(\cdot) \):
\begin{equation}
\tilde{V}_{x}=\mathcal{D}_V(\mathcal{E}_{V}(V_{x})).
\end{equation}

\setlength{\columnsep}{2pt} 
\begin{wrapfigure}{r}{0.225\textwidth}
    \centering
    \vspace{-4mm}
    \includegraphics[width=0.2\textwidth]{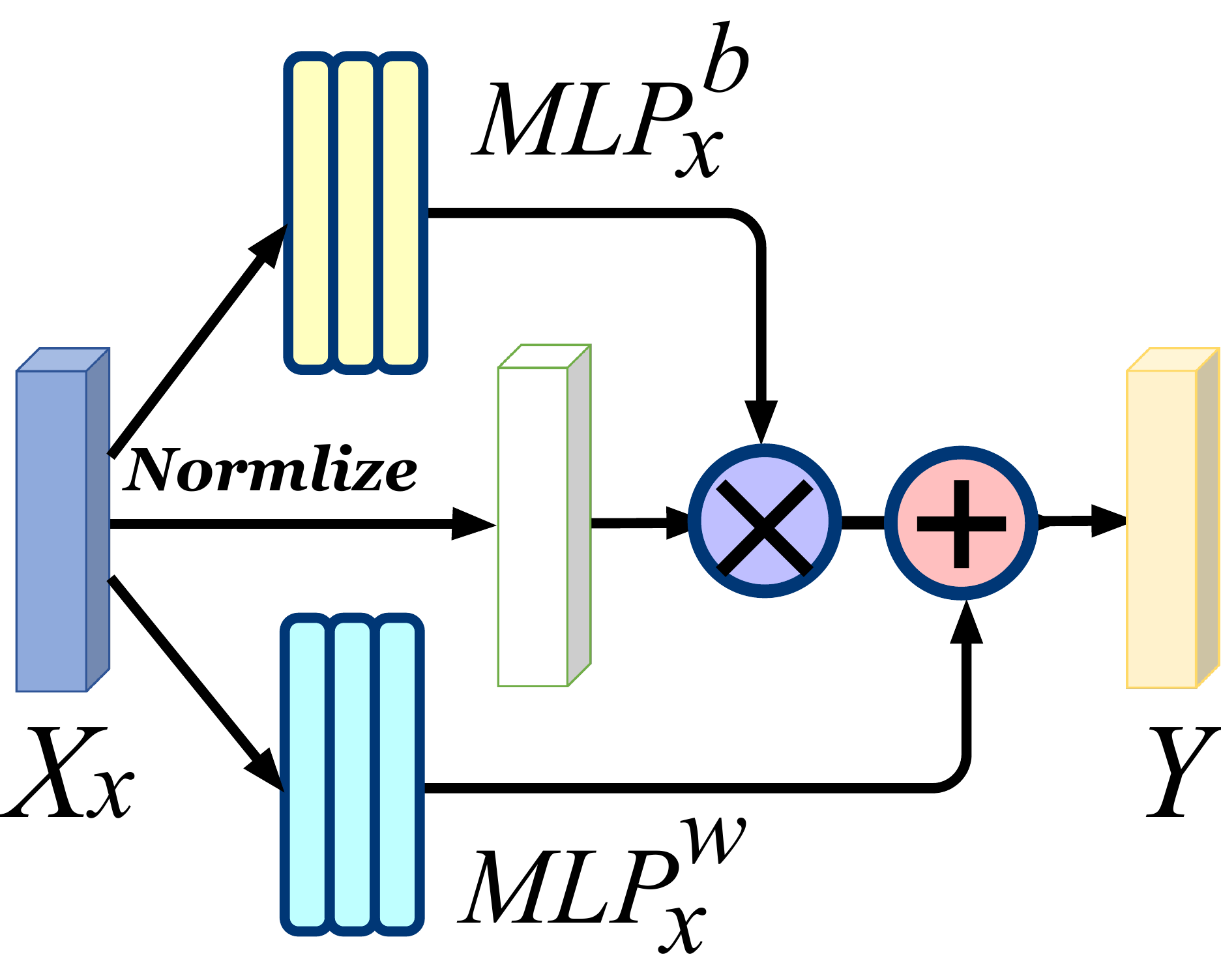}
    \vspace{-4mm}
    \caption{Structure of SBMM.}
    \label{fig:sbmm}
    \vspace{-4mm}
\end{wrapfigure}

To obtain subject-invariant features, we reduce subjective bias in \( V_{x} \) by introducing a Subject Bias Modulation Module (SBMM) within the fMRI encoder for each subject. As shown in Fig.~\ref{fig:sbmm}, the SBMM consists of multiple Multi-Layer Perceptrons (MLPs) that modulate the mean and variance of each subject’s fMRI features. Let \( X_{x} \) denotes the input feature of the SBMM. The SBMM uses two MLPs to predict the statistical characteristics of each subject, which are then used to modulate the normalized input. The operation in SBMM can be represented as:
\begin{equation}
Y = \mathcal{MLP}_x^W(X_x) \left( \frac{X_x - \mu(X_x)}{\sigma(X_x)} \right) + \mathcal{MLP}_x^{B}(X_x),
\end{equation}
where \( Y \) denotes the output of the SBMM, \( \mathcal{MLP}_x^W(\cdot) \) represents the MLP for modulating the variance of the input, and \( \mathcal{MLP}_x^B(\cdot) \) is the MLP for mean modulation. This module effectively reduces subject-specific bias. {With this module, the framework can be efficiently adapted to new subjects by training only the SBMM for the novel subject. More discussion about this module can be seen in Sec.~\ref{sec:ans}.}

The decoder \( \mathcal{D}_{V}(\cdot) \) has a mirrored structure to the encoder and is also integrated with SBMM, enabling it to decode subject-invariant features back into subject-specific fMRI voxels. We define an \( \mathcal{L}_2 \) distance between the input and reconstructed fMRI voxels for reconstruction:
\begin{equation}
\mathcal{L}^\text{Rec}_{V} = \|\tilde{V}_{x} - V_{x}\|_2.
\end{equation}

Similarly, the autoencoder for representations receives the three representations simultaneously. Each representation is processed with a specific encoding head, and the resulting features are concatenated. The concatenated feature is then decoded into reconstructed representations using distinct decoding heads. Note that SBMM is not applied in this autoencoder, as the representations are subject-invariant. The reconstruction pipeline for representations is represented as:
\begin{equation}
\tilde{S},\tilde{E},\tilde{C} = \mathcal{D}_R(\mathcal{E}_{R}(S,E,C)),
\label{eq:rec}
\end{equation}
where $\mathcal{E}_{R}(\cdot)$, $\mathcal{D}_{R}(\cdot)$ denote the encoder and decoder for representations respectively, and $\tilde{S}$,$\tilde{E}$,$\tilde{C}$ denote the reconstructed representations respectively. This autoencoder is trained with reconstruction loss of three representations:
\begin{equation}
\mathcal{L}^\text{Rec}_{E} = \texttt{BCE}(\tilde{E}, E), \quad \mathcal{L}^\text{Rec}_{C} = \|\tilde{C} - C \|_2,
\end{equation}
\begin{equation}
\mathcal{L}^\text{Rec}_{S} = 1 - \texttt{cos}(\tilde{S}, S) + \|\tilde{S} - S \|_2,
\end{equation}
where $\mathcal{L}^\text{Rec}_{E}$, $\mathcal{L}^\text{Rec}_{C}$, and $\mathcal{L}^\text{Rec}_{S}$ denote the reconstruction loss for edge map, color palette, and semantic embedding respectively. $\texttt{cos}(\cdot,\cdot)$ calculates the cosine similarity of two inputs, and $\texttt{BCE}(\cdot,\cdot)$ is the binary cross-entropy loss.

\textbf{Bidirectional Mapping.} The BAI framework supports flexible bidirectional translation by simply swapping the decoders. As illustrated by the red lines in Fig.~\ref{fig:overview}, the translation from fMRI voxels to representations is performed by learning an MLP, \( \mathcal{MLP}_{V{\Rightarrow}R}(\cdot) \), that maps the encoded fMRI features to representation features, which are then decoded using \( \mathcal{D}_{R} \):
\begin{equation}
\hat{S}, \hat{E}, \hat{C} = \mathcal{D}_R(\mathcal{MLP}_{V{\Rightarrow}R} (\mathcal{E}_{V}(V_{x}))).
\label{eq:r2v}
\end{equation}
On the other hand, we can also mimic the human visual system by mapping the representations to the corresponding fMRI voxels using another MLP, \( \mathcal{MLP}_{R{\Rightarrow}V}(\cdot) \), which maps representations features to the fMRI feature:
\begin{equation}
\hat{V}_{x} = \mathcal{D}_V(\mathcal{MLP}_{R{\Rightarrow}V} (\mathcal{E}_{R}(S,E,C))).
\label{eq:v2r}
\end{equation}
The translation pipeline is trained using fMRI-representation pairs. Similar to the reconstruction pipeline, the translation losses for fMRI voxels and the three representations are denoted as \( \mathcal{L}^\text{Tr}_{V} \), \( \mathcal{L}^\text{Tr}_{S} \), \( \mathcal{L}^\text{Tr}_{E} \), and \( \mathcal{L}^\text{Tr}_{C} \), respectively.

% \(\mathcal{L}^{\text{Cyc}}_X = \|X - \doublehat{X}\|_2^2, X \in \{S, E, C, {V_x}\} \)

Moreover, our bidirectional mapping framework introduces cycle consistency through cyclic mapping. Specifically, we map the real representations to fMRI voxels using Eq.~\ref{eq:v2r}, and then reconstruct the representations back from the mapped voxels using Eq.~\ref{eq:r2v}: $\{S, E, C\} \Rightarrow \hat{V}_{x}, \hat{V}_{x} \Rightarrow \{\doublehat{S}, \doublehat{E}, \doublehat{C}\},$, where \( \{\doublehat{\mathbf{\cdot}}\} \) denotes the cyclic reconstructed results. The cycle consistency loss for the three representations is computed by minimizing the distance between the input and cyclically reconstructed representations:
%\(\mathcal{L}^{\text{Cyc}}_X = \| X - \doublehat{X} \|_2^2, \quad X \in \{S, E, C, V_x\}\).
\begin{equation}
\mathcal{L}^{\text{Cyc}}_K = \| K - \doublehat{K} \|_2^2, \quad K \in \{S, E, C, V\}.
\end{equation}

Similarly, we can achieve cycle reconstruction for fMRI voxels with $V_{x} \Rightarrow \{\hat{S}, \hat{E}, \hat{C}\}, \{\hat{S}, \hat{E}, \hat{C}\} \Rightarrow \doublehat{V}_{x}$. The BAI framework is trained with the combined losses:
\begin{equation}
\begin{aligned}
\mathcal{L} = \lambda_1
&\underbrace{\mathcal{L}^\text{Rec}_{S} + \mathcal{L}^\text{Rec}_{E} + \mathcal{L}^\text{Rec}_{C} + \mathcal{L}^\text{Rec}_{V}}_{\text{Reconstruction}}
+ \lambda_2 \underbrace{\mathcal{L}^\text{Tr}_{S} + \mathcal{L}^\text{Tr}_{E} + \mathcal{L}^\text{Tr}_{C} + \mathcal{L}^\text{Tr}_{V}}_{\text{Translation}}   \\
+\lambda_3 &\underbrace{\mathcal{L}^\text{Cyc}_{S} + \mathcal{L}^\text{Cyc}_{E} + \mathcal{L}^\text{Cyc}_{C} + \mathcal{L}^\text{Cyc}_{V}}_{\text{Cycle-Consistency}},
\end{aligned}
\end{equation}
where $\lambda_s$ denote the balance factors, and we set $\lambda_1$ = 1, $\lambda_2$ = 1, and $\lambda_3$ = 0.5 empirically.

\subsection{Reconstruction from Inaccurate Representations}
\label{sec:rnp}

\begin{figure}[!t]
    \centering
    \captionsetup[subfloat]{justification=centering}
     \subfloat[\scriptsize{Prd. Edge}]{
     \begin{minipage}{0.19\linewidth}
     \includegraphics[width=\linewidth]{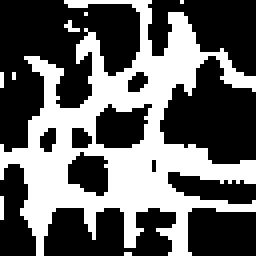}
     \includegraphics[width=\linewidth]{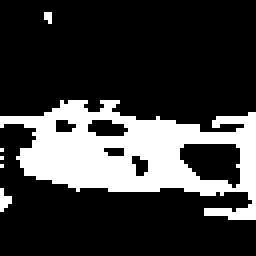}
     \end{minipage}
     }
     \hspace{-2.25mm}
     \subfloat[\scriptsize{Prd. Color}]{
     \begin{minipage}{0.19\linewidth}
     \includegraphics[width=\linewidth]{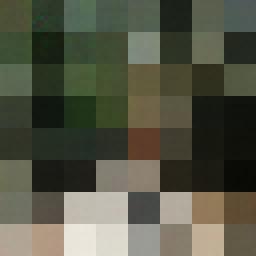}
     \includegraphics[width=\linewidth]{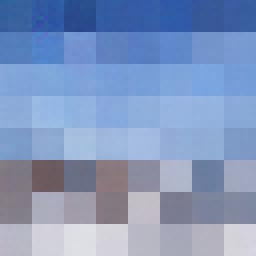}
     \end{minipage}
     }
     \hspace{-2.25mm}
     \subfloat[\scriptsize{Direct Decod.}]{\label{dd}
     \begin{minipage}{0.19\linewidth}
     \includegraphics[width=\linewidth]{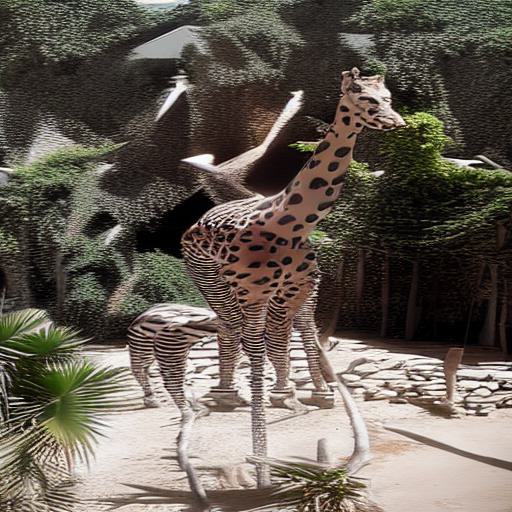}
     \includegraphics[width=\linewidth]{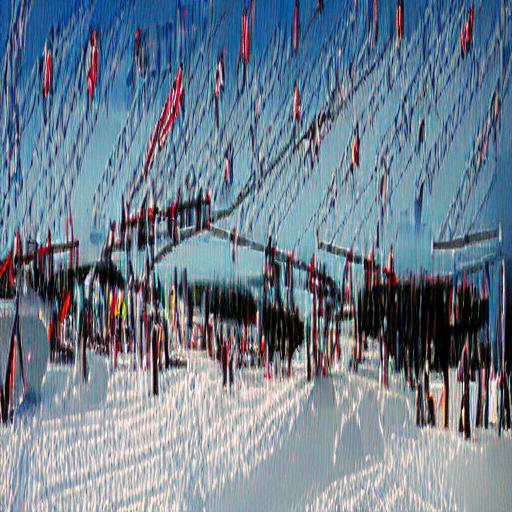}
     \end{minipage}
     }
     \hspace{-2.25mm}
     \subfloat[\scriptsize{Our Decod.}]{
     \begin{minipage}{0.19\linewidth}\label{bai}
     \includegraphics[width=\linewidth]{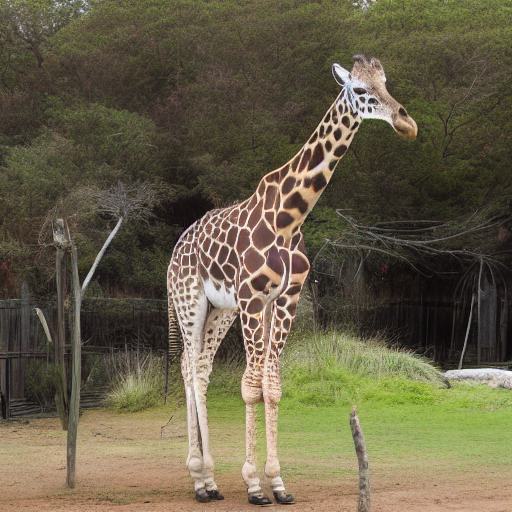}
     \includegraphics[width=\linewidth]{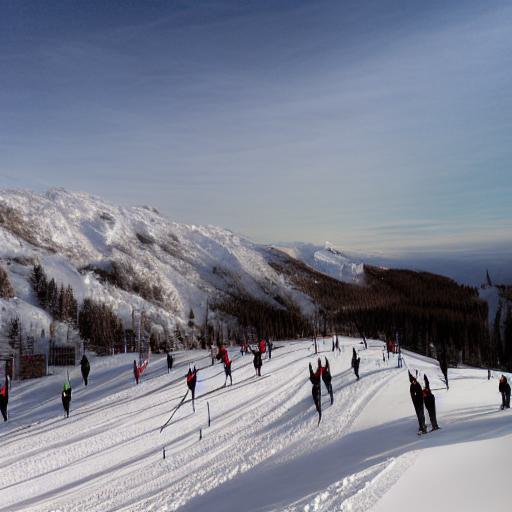}
     \end{minipage}
     }
     \hspace{-2.25mm}
     \subfloat[\scriptsize{Stimulus}]{
     \begin{minipage}{0.19\linewidth}
     \includegraphics[width=\linewidth]{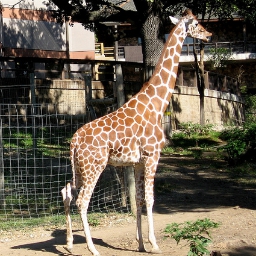}
     \includegraphics[width=\linewidth]{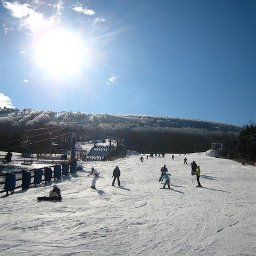}
     \end{minipage}
     }
\vspace{-2mm}
\caption{\rev{``Direct Decod.'' denotes the decodes the predicted representations using diffusion models directly, which loses realism and fidelity. While using our decoded results are more faithful with stimulus GT.}}
\vspace{-5mm}
\label{fig:figure_pre}
% \vspace{-3mm}
\end{figure}

The BAI maps fMRI voxels to image representations. A common solution for mind decoding involves using ControlNet~\cite{zhang2023adding} or T2I-adapter~\cite{mou2024t2i}, which controls the output using these representations~\cite{xia2024dream}. In this case, let \( F_i \) denote the feature of the \(i\)-th layer of the decoder \(\mathcal{D}^\text{UNet}\) in the diffusion model's UNet, and the control process is performed by adding the representation features directly to \( F_i \):
\begin{equation}
F_{i+1} = \mathcal{D}^\text{UNet}_{i} (F_i +  \hat{E}_i + \hat{C}_i, \hat{S}),
\label{eq:controlnet1}
\end{equation}
where \( \hat{E}_i \) and \( \hat{C}_i \) represent edge and color features obtained from \( \hat{E} \) and \( \hat{C} \), respectively.

However, this operation requires extremely precise representations for accurate reconstruction, and the predicted representations often do not meet the strict requirements. As shown in Fig.~\ref{dd}, \rev{decoding the predicted representations using diffusion models directly suffers from low fidelity due to inaccuracies.} To mitigate this, we introduce two modules: the Semantic Refinement Module (SRM) and the Visual Coherence Module (VCM), which respectively handle errors in semantic and visual representations. 

\begin{figure}[!t]
    \centering
    % \captionsetup[subfloat]{justification=centering}
    \captionsetup[subfloat]{justification=centering,skip=-5pt}  % 减少子图标题和图片的距离
     \subfloat[\scriptsize{Semantic Refinement Module}]{
     \begin{minipage}{0.475\linewidth}
     \includegraphics[width=0.99\textwidth]{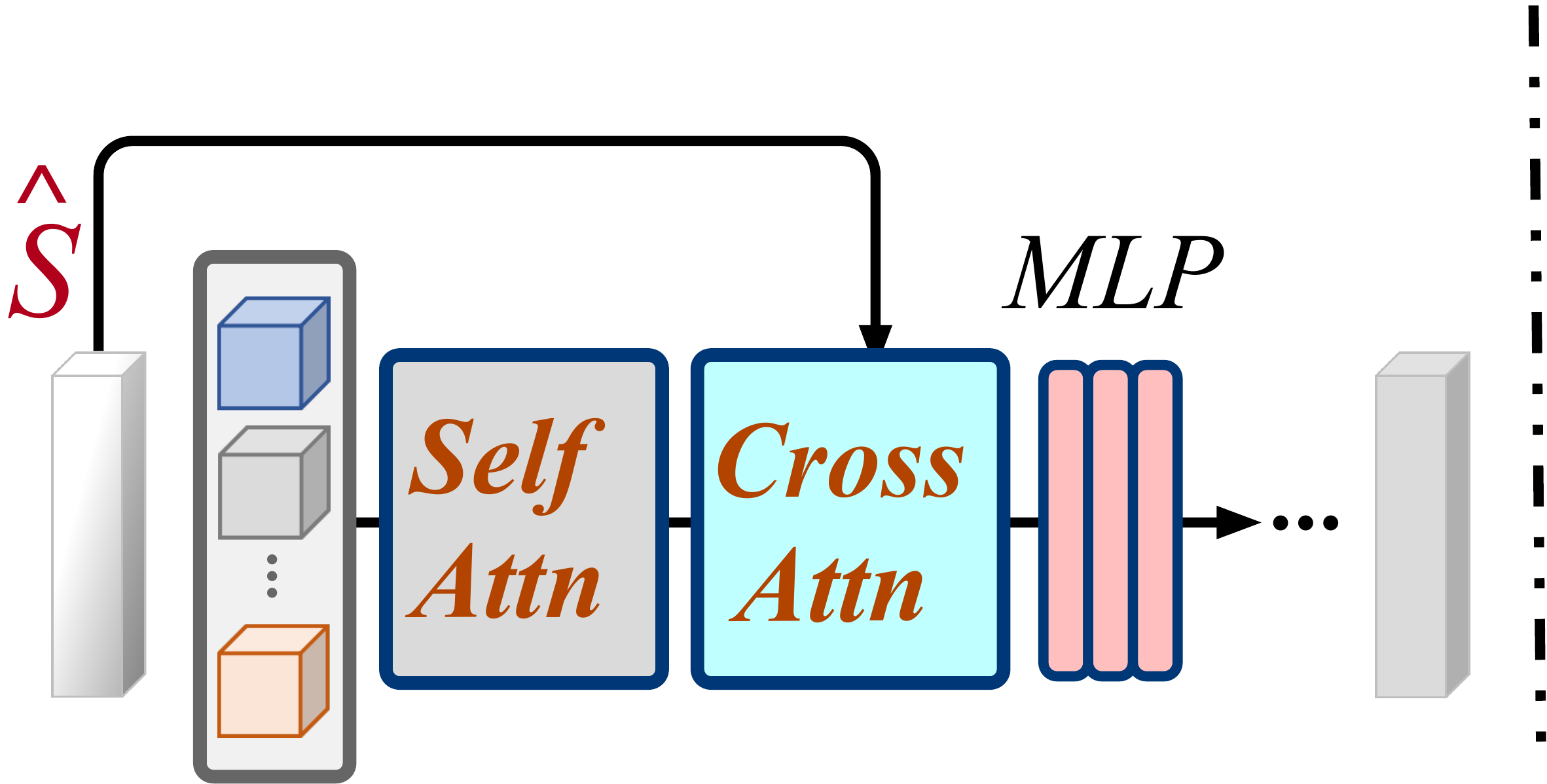}
     \label{fig:srm}
     \end{minipage}
     }
     \hspace{-1mm}
     \subfloat[\scriptsize{Visual Coherence Module}]{
     \begin{minipage}{0.475\linewidth}
     \includegraphics[width=0.99\textwidth]{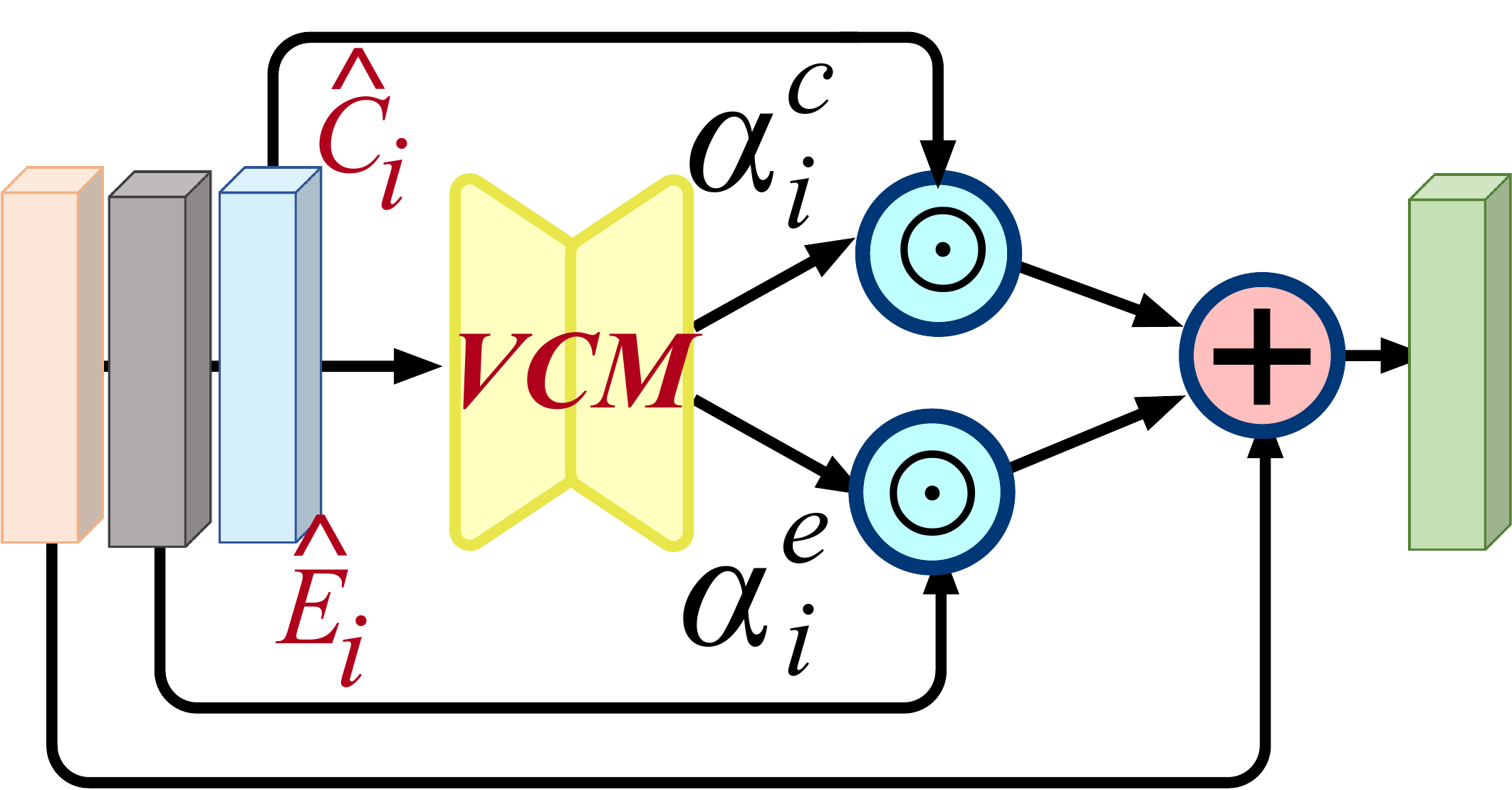}
     \label{fig:vcm}
     \end{minipage}
     }
\vspace{-2mm}
\caption{\rev{Detailed structure of our SRM and VCM.}}
\vspace{-5mm}
\end{figure}

The Semantic Refinement Module (SRM) refines imprecise semantic representations. It is a transformer-based structure (see in Fig.~\ref{fig:srm}), trained by minimizing the distance between its output and the ground truth semantic embeddings:
\begin{equation}
\mathcal{L}_{SRM} = 1 - \texttt{cos}(\text{SRM}(\tilde{S}), S) + \|\text{SRM}(\tilde{S}) - S \|_2,
\end{equation}
while this module use the similarly losses in training of BAI, this module tolerates imprecise semantic representations, thereby improving the fidelity during decoding stage.

We introduce VCM that harmonizes the imprecise visual representation features to original diffusion features. As indicated by~\cite{liu2024smartcontrol}, the weights of different features control the influence of visual conditions to the output image, our VCM is designed for predicting the weights that cohere different features. The structure of VCM is shown in Fig.~\ref{fig:vcm}, it takes the concatenation of three features as input and produces two weights $\alpha_{e}$ and $\alpha_{c}$ with the same spatial size of representations:

\begin{equation}
\alpha^{e}_i, \alpha^{c}_i = \text{VCM}(\texttt{concat}({F}_i, \hat{E}_i, \hat{C}_i)),
\end{equation}
where $\texttt{concat}(\cdot)$ is the concatenate operation. The controlling process can be rewritten as:

\begin{equation}
F_{i+1} = \mathcal{D}^\text{UNet}_{i} (F_i +  \alpha^{e}_i \odot \hat{E}_i + \alpha^{c}_i \odot \hat{C}_i, \hat{S}),
\end{equation}
and the VCM is trained with:
\begin{equation}
\mathcal{L}_{\mathrm{VCM}}=\| {\epsilon}-{\epsilon}_\theta ({z}_t,  t, \text{SRM}(\tilde{S}), \tilde{C}, \tilde{E}) \|^2_2.
\label{eq.vcm}
\end{equation}
Finally, the stimuli image $\tilde{I}$ can be reconstructed faithfully from imprecise representations with DDIM sampler~\cite{song2021denoising}.

\section{Experiments}
\subsection{Experimental Settings}

\label{sec:es}

\begin{table*}[h!]
\caption{{Qualitative comparisons with related works on NSD dataset. All metrics are calculated as the average across 4 subjects.}}
\vspace{-0.25cm}
% \footnotesize
\centering
\setlength{\tabcolsep}{0.8mm}{
\begin{threeparttable}
\begin{tabular}{c|c|cccc|cccc}
% \toprule
\hline
\multirow{2}{*}{\centering \textbf{Method}}   &  \multirow{2}{*}{\makecell{\textbf{Cross} \\ \textbf{Subject?}}}  & \multicolumn{4}{c|}{\textbf{Low-Level}} & \multicolumn{4}{c}{\textbf{High-Level}} \\
\cline{3-10}&  &{PixCorr} $\uparrow$ & {SSIM} $\uparrow$     & {AlexNet(2)} $\uparrow$  &  {AlexNet(5)} $\uparrow$    & {Incep} $\uparrow$    & {CLIP}  $\uparrow$  & {EffNet-B}  $\downarrow$     & {SwAV}  $\downarrow$    \\

\hline

% \Checkmark  \XSolidBrush
\rowcolor{gray!20}MindReader~\cite{lin2022mind} & ✗ & - &- &  - &  - &  78.2\% &  - & - &  -\\
Takagi~\etal~\cite{takagi2023high}  & ✗ &  -  &  -  &  83.0\%  &  83.0\%  &  76.0\%  &  77.0\%  &  --  &  --\\
\rowcolor{gray!20}BrainDiffuser~\cite{ozcelik2023natural}  & ✗ &  .254  &  .356  &  94.2\%  &  96.2\%  &  87.2\%  &  91.5\%  &  .775  &  .423\\
MindEye1~\cite{scotti2024reconstructing}  & ✗  &  .309  &  .323  &  94.7\%  &  97.8\%  &  93.8\%  &  94.1\%  &  .645  &  .367\\
\rowcolor{gray!20}Gu~\etal~\cite{gu2022decoding} &  ✗ & .150 &.325 &  - &  - &  - &  - & .862 &  .465 \\
MindVis~\cite{chen2023seeing} &  ✗ & .080 &.220 &  72.1\% &  83.2\% &  78.8\% &  76.2\% & .854 &  .491 \\
\rowcolor{gray!20}DREAM~\cite{xia2024dream}  &  ✗ & .288 &.338 &  95.0\% &  97.5\% &  94.8\% &  95.2\% & .638 &  .413 \\

\hline

MindBridge~\cite{wang2024mindbridge}  &   ✓  &  .151  &  .263  &  87.7\%  &  95.5\%  &  92.4\%  &  {94.7\%}  &  .712  &  .418\\
\rowcolor{gray!20}MindEye2~\cite{scotti2024mindeye2}\tnote{\dag}  &  ✓ &  .207  &  .350  &  91.6\%  &  96.4\%  &  89.4\%  &  83.6\%  &  .728  &  .423\\
NeuroPictor~\cite{huo2024neuropictor}&   ✓  &.229 &.375 &96\% &98.4\% &94.5\% &93.3\% &.639 &.350 \\
\rowcolor{gray!20}UMBRAE~\cite{xia2024umbrae}       &   ✓ &.283 &.341 &95.5\% &97.0\% &91.7\% &93.5\% &.700 &.393 \\
Psychometry~\cite{quan2024psychometry}  &  ✓  &  .297  &  .340  &  96.4\%  &  98.6\%  &  95.8\%  &  {96.8\%}  &  {.628}  &  {.345}\\
% Neuro-V~\cite{shen2024neuro}  &  ✓       & .265           & \textbf{.357}             & 93.1\%          & 97.1\%     & \textbf{96.8\%}           & \textbf{97.5\%}     & \textbf{.633} &\textbf{.321} \\
\hline
\rowcolor{gray!20}Ours  &  ✓   &\textbf{.318} &{.356} &\textbf{97.3\%}  &\textbf{98.8\%} &{96.7\%} &\textbf{97.5\%} &{.639} &{.345} \\

\hline
% \bottomrule
\end{tabular}

\begin{tablenotes}
   \footnotesize
   \item[\dag] {The result reported in original paper~\cite{scotti2024mindeye2} is trained on 8 subjects, we re-train their model on 4 subjects using the official code for fair comparison.}
\end{tablenotes}
\end{threeparttable}}
\label{tab:main}
\vspace{-5mm}
\end{table*}

\textbf{Dataset.} \rev{We follow previous works~\cite{wang2024mindbridge, quan2024psychometry, gu2022decoding, takagi2023high, xia2024dream} that conduct experiments on the largest mind decoding dataset, the Natural Scenes Dataset (NSD)\cite{allen2022massive}. NSD comprises 7-Tesla fMRI scans collected from eight subjects as they viewed thousands of stimulus images from the MS-COCO dataset\cite{lin2014microsoft}. Following prior studies~\cite{wang2024mindbridge, gu2022decoding, takagi2023high}, we use data from four subjects (Subj01, Subj02, Subj05, and Subj07) in our experiments. Each subject's training set consists of 8,859 fMRI-stimuli-caption pairs, while the test set includes 982 images viewed by all four subjects. The Regions of Interest (ROIs) in fMRI signals vary in size across subjects. To standardize these variations, we adopt the adaptive max pooling function used in MindBridge~\cite{wang2024mindbridge}, which resizes the ROI representations to a fixed dimension of 8,192. We obtain semantic embeddings by encoding the image captions with the CLIP text encoder~\cite{radford2021learning}. Additionally, edge maps are extracted from stimulus images using PidiNet~\cite{pdc-PAMI2023}. To obtain color palettes, we follow the approach in T2I-Adapter~\cite{mou2024t2i}, applying a $64\times$ downsampling and subsequent upsampling to the original resolution.}

\textbf{Evaluation Metrics.} Following existing works~\cite{wang2024mindbridge,ozcelik2023natural,scotti2024mindeye2,scotti2024reconstructing}, we use 8 evaluation metrics for the quantitative comparison from low and high levels. The low-level metrics include {PixCorr}, {SSIM}, {AlexNet(2)}, and {AlexNet(5)}, and the high-level metrics include {Inception}, {CLIP}, {EffNet-B}, and {SwAV}. For a detailed introduction to the metrics, please refer to the supplementary materials.

\subsection{Implementation Details}
\label{sec:imp}

We implement the proposed framework in PyTorch with Nvidia GeForce A100. The BAI framework is trained using the AdamW optimizer~\cite{loshchilov2017decoupled} with a learning rate of 1e-4. The batch size is set to 192 and we train for 1,000 epochs. The SRM and VCM require inaccurate representations for training, but the training set of NSD fits well with the trained BAI model, and their predicted representations contain fewer errors. To obtain the imprecise representations, we first generate 10,000 images using SD with random prompts generated using a Large Language Model (LLM), and then extract their representations as described in Sec.~\ref{sec:es}. We then reconstruct these representations with BAI's reconstruction pipelines, obtaining pseudo-imprecise representation-image pairs for training two modules. They are trained together with the AdamW optimizer~\cite{loshchilov2017decoupled} and a learning rate of 1e-4. We use Stable Diffusion V1.5 as our text-guided diffusion model, setting the inference steps in the DDIM sampler to 20. Two ControlNets are pre-trained on edge maps and color palettes respectively. For the structural details of the BAI, SRM, and VCM, please refer to the supplementary materials.

\subsection{Comparison on Mind decoding}

We present the quantitative comparisons with the state-of-the-art methods in Tab.~\ref{tab:main}. We can see that our method outperforms most of works on both low-level and high-level metrics. Especially, our method achieves higher values on PixCorr and SSIM metrics, which indicates that our decoded images are more similar to the stimulus images in structure and appearance. By predicting semantic knowledge through bidirectional mapping and refining it with SRM, our framework effectively decodes semantic information. As a result, our method receives the highest CLIP value among all methods. Moreover, our single cross-subject framework outperforms subject-specific frameworks on all metrics. By learning the bidirectional mapping with SBMM, our framework learns robust representation features that are agnostic to different subjects.

\begin{figure*}[!t]
    \centering
    \captionsetup[subfloat]{labelformat=empty,justification=centering}
     \subfloat[Takagi~\etal~\cite{takagi2023high}]{
     \begin{minipage}{0.12\linewidth}
     \includegraphics[width=\linewidth]{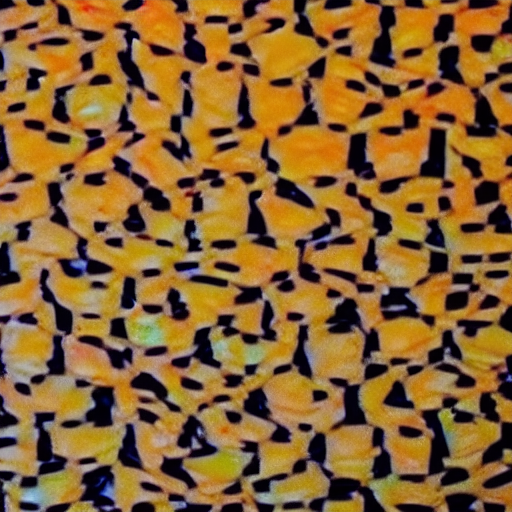}
     \includegraphics[width=\linewidth]{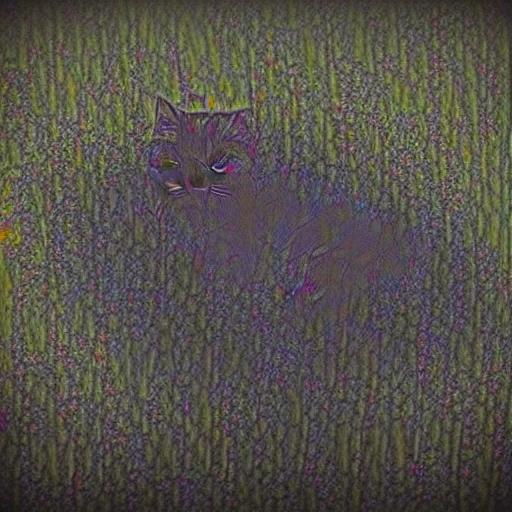}
     \includegraphics[width=\linewidth]{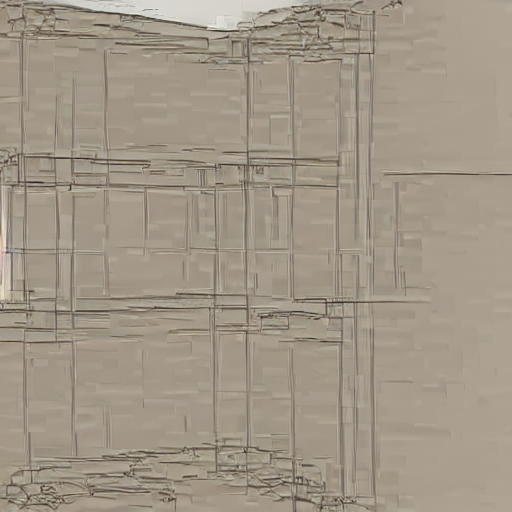}
     \includegraphics[width=\linewidth]{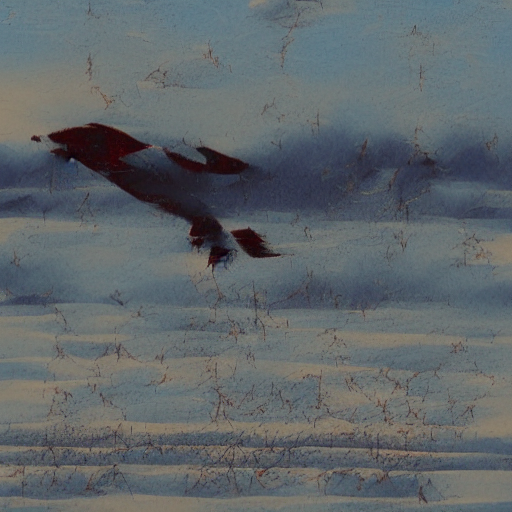}
     \includegraphics[width=\linewidth]{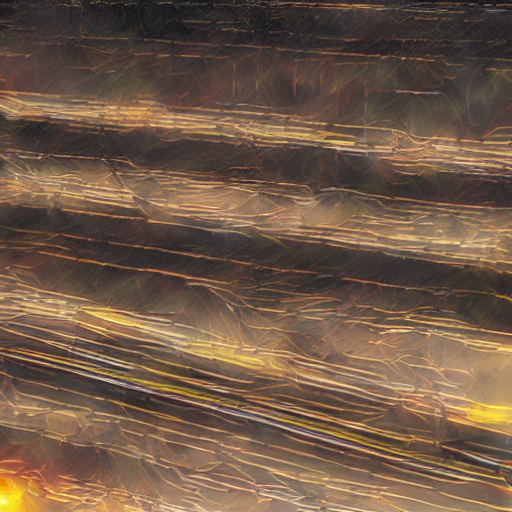}
     \includegraphics[width=\linewidth]{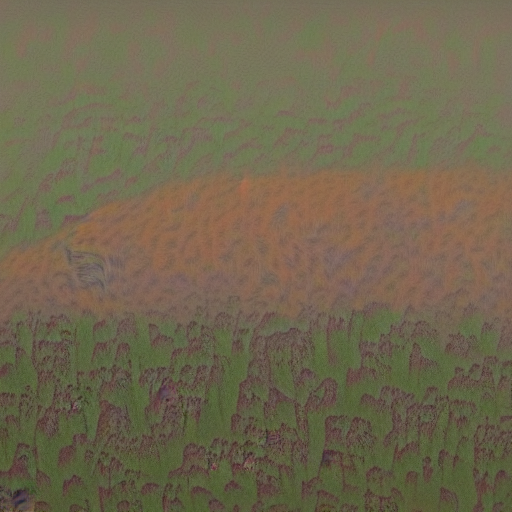}
     \end{minipage}
     }
     \hspace{-2.25mm}
     \subfloat[BrainDiffuser~\cite{ozcelik2023natural}]{
     \begin{minipage}{0.12\linewidth}
     \includegraphics[width=\linewidth]{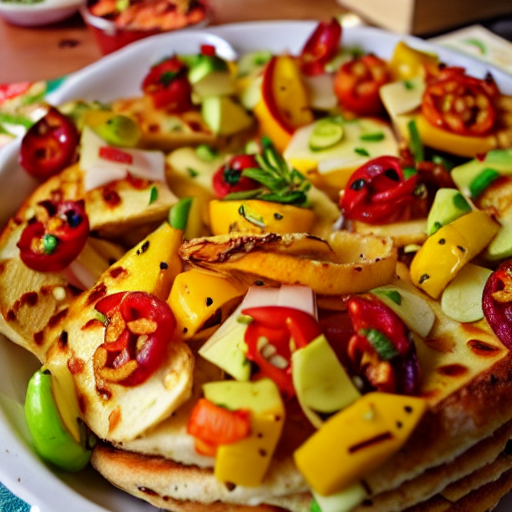}
     \includegraphics[width=\linewidth]{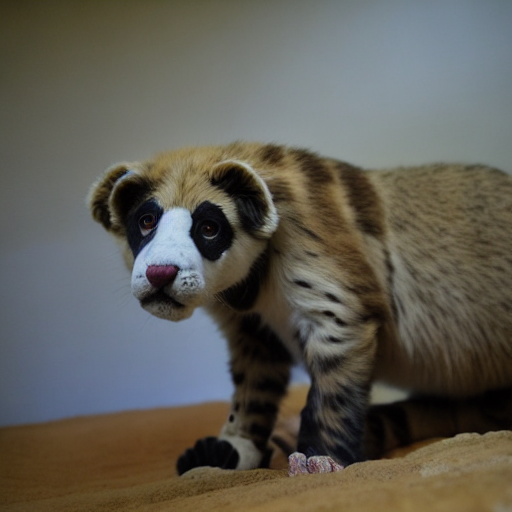}
     \includegraphics[width=\linewidth]{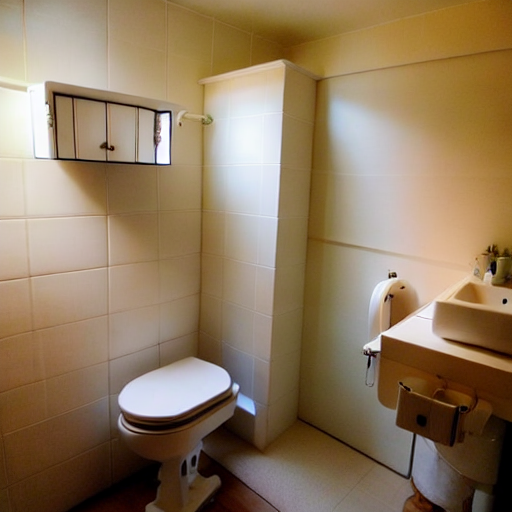}
     \includegraphics[width=\linewidth]{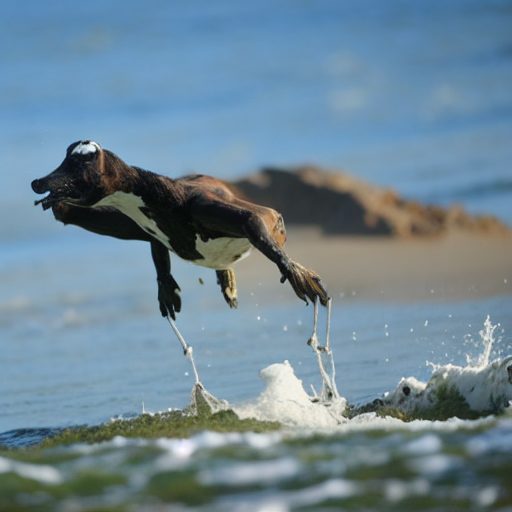}
     \includegraphics[width=\linewidth]{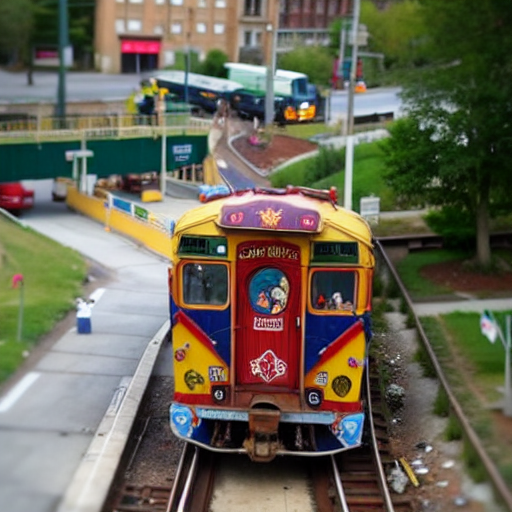}
     \includegraphics[width=\linewidth]{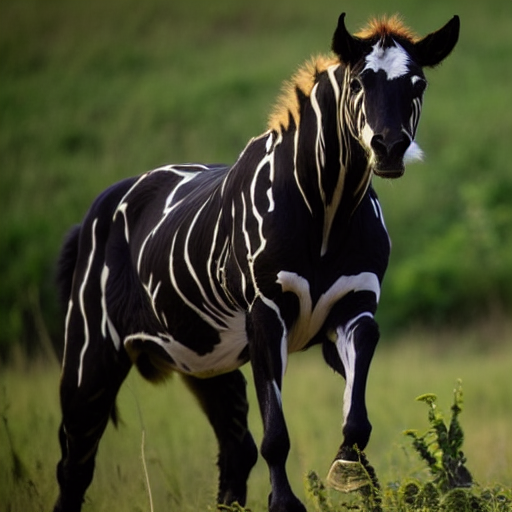}
     \end{minipage}
     }
     \hspace{-2.25mm}
     \subfloat[MindEye1~\cite{scotti2024reconstructing}]{
     \begin{minipage}{0.12\linewidth}
     \includegraphics[width=\linewidth]{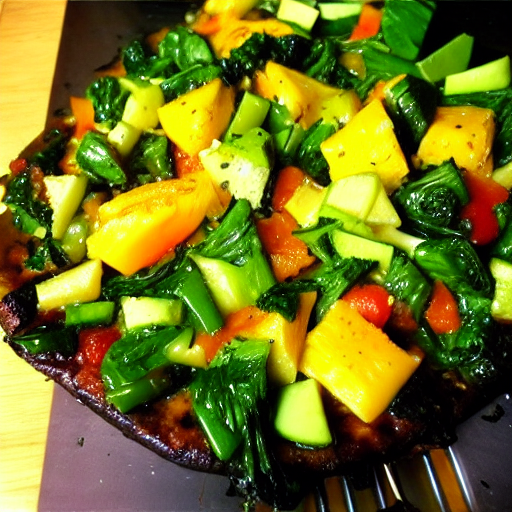}
     \includegraphics[width=\linewidth]{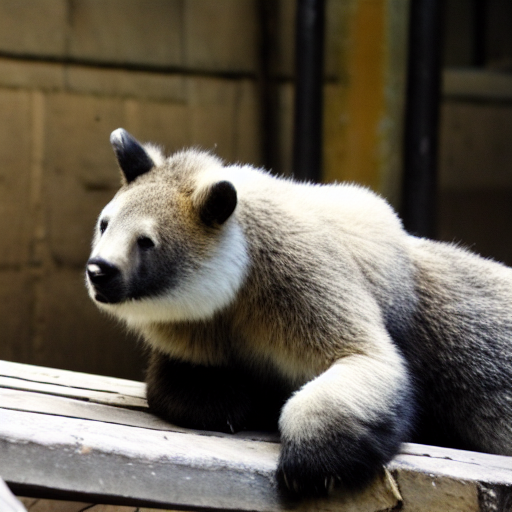}
     \includegraphics[width=\linewidth]{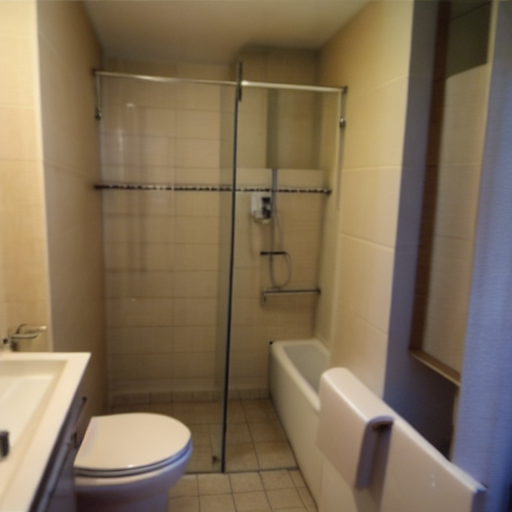}
     \includegraphics[width=\linewidth]{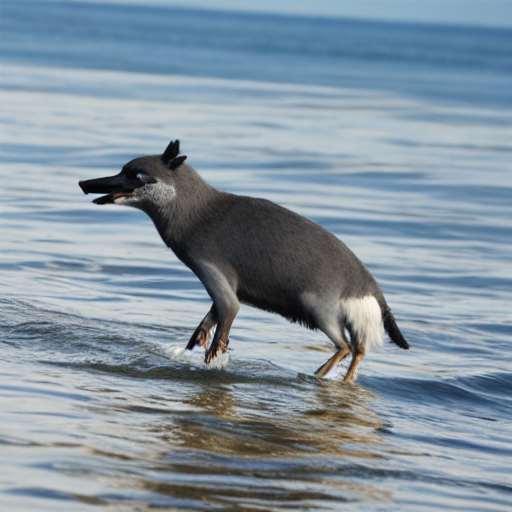}
     \includegraphics[width=\linewidth]{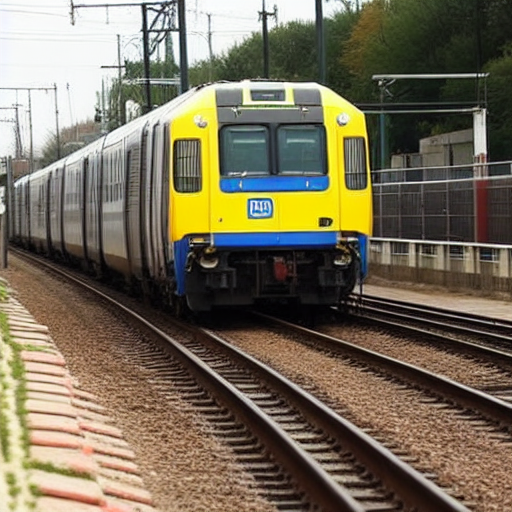}
     \includegraphics[width=\linewidth]{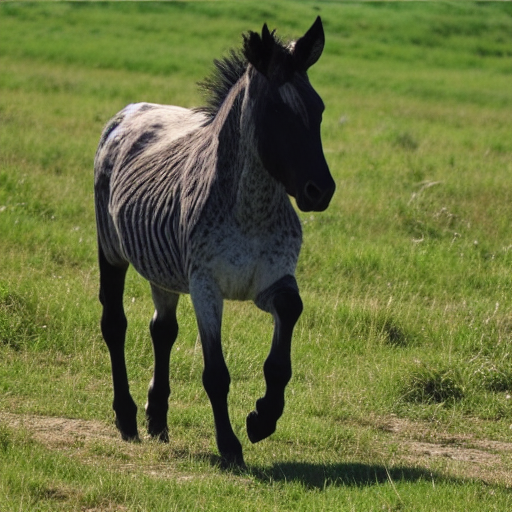}
     \end{minipage}
     }
     \hspace{-2.25mm}
     \subfloat[MindBridge~\cite{wang2024mindbridge}]{
     \begin{minipage}{0.12\linewidth}
     \includegraphics[width=\linewidth]{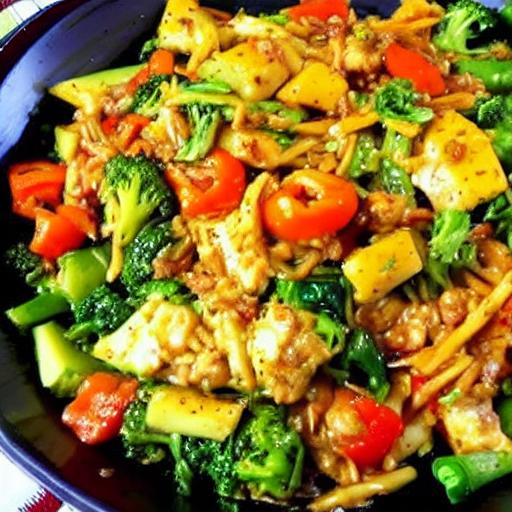}
     \includegraphics[width=\linewidth]{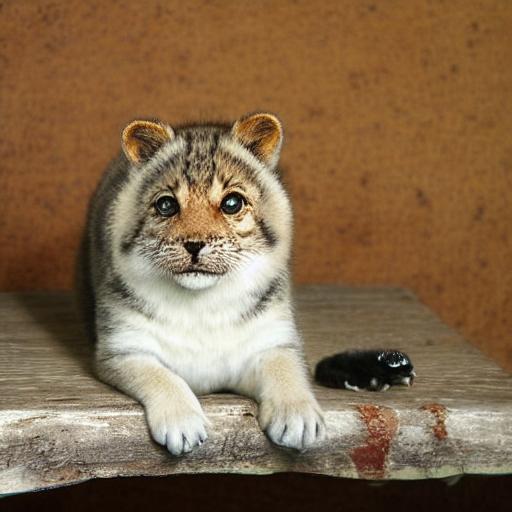}
     \includegraphics[width=\linewidth]{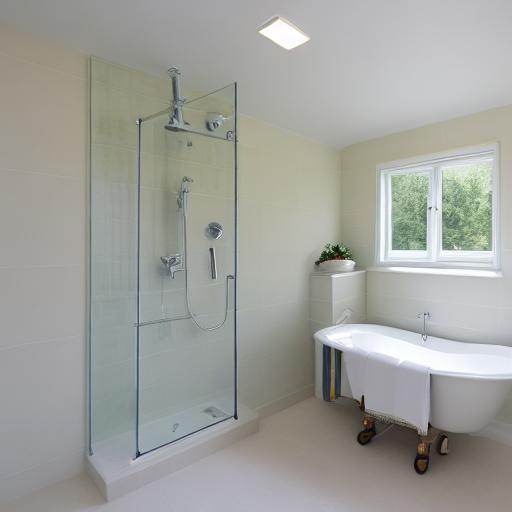}
     \includegraphics[width=\linewidth]{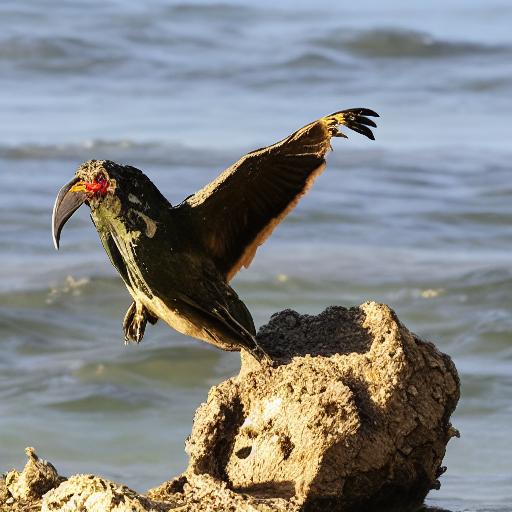}
     \includegraphics[width=\linewidth]{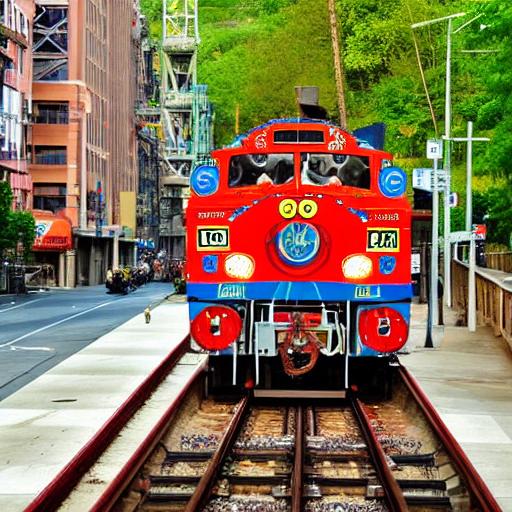}
     \includegraphics[width=\linewidth]{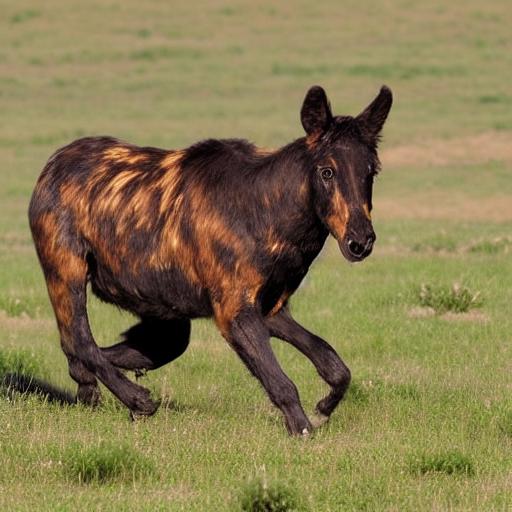}
     \end{minipage}
     }
     \hspace{-2.25mm}
     \subfloat[MindEye2~\cite{scotti2024mindeye2}]{
     \begin{minipage}{0.12\linewidth}
     \includegraphics[width=\linewidth]{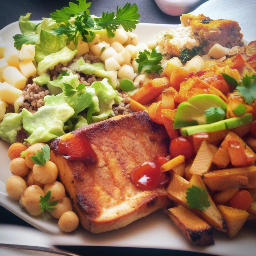}
     \includegraphics[width=\linewidth]{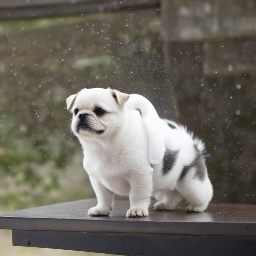}
     \includegraphics[width=\linewidth]{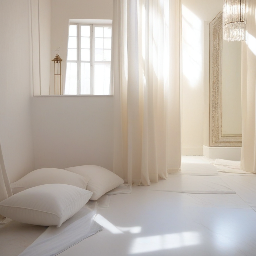}
     \includegraphics[width=\linewidth]{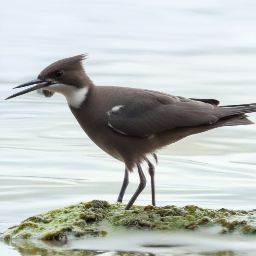}
     \includegraphics[width=\linewidth]{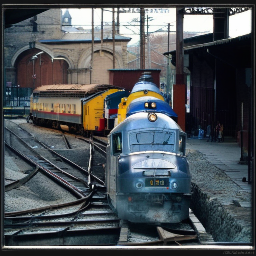}
     \includegraphics[width=\linewidth]{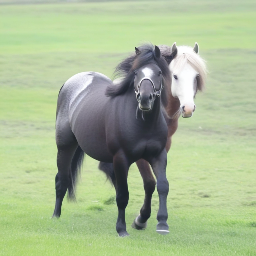}
     \end{minipage}
     }
     \hspace{-2.25mm}
     \subfloat[Neuropictor~\cite{huo2024neuropictor}]{
     \begin{minipage}{0.12\linewidth}
     \includegraphics[width=\linewidth]{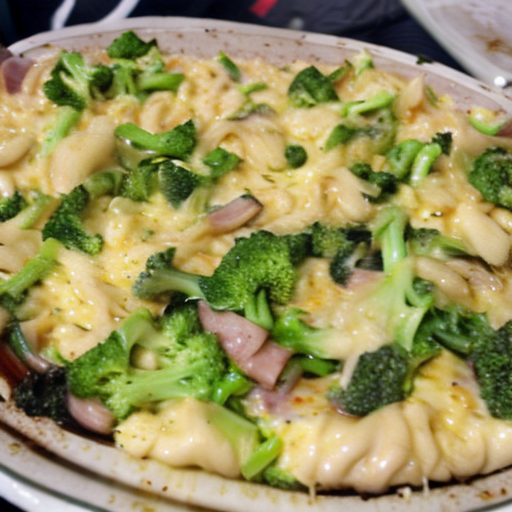}
     \includegraphics[width=\linewidth]{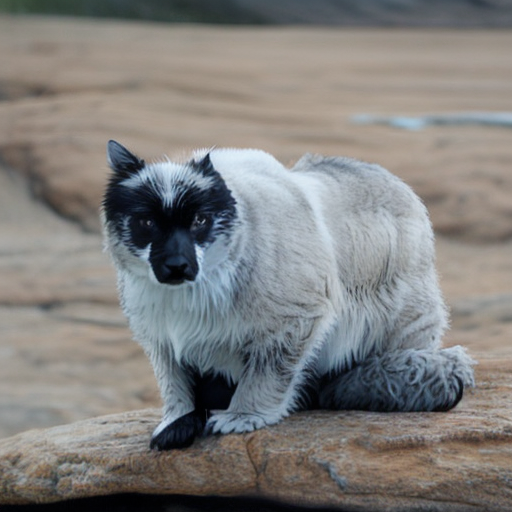}
     \includegraphics[width=\linewidth]{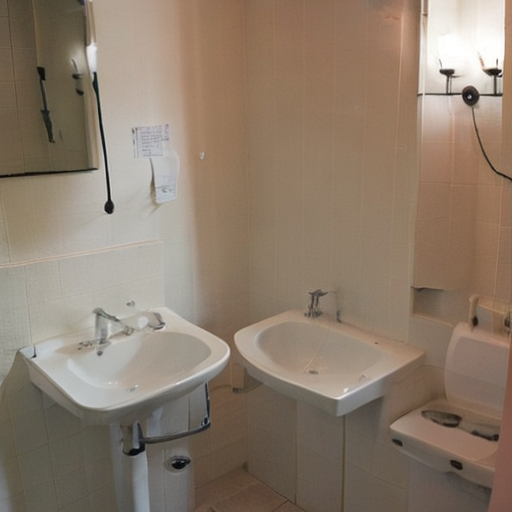}
     \includegraphics[width=\linewidth]{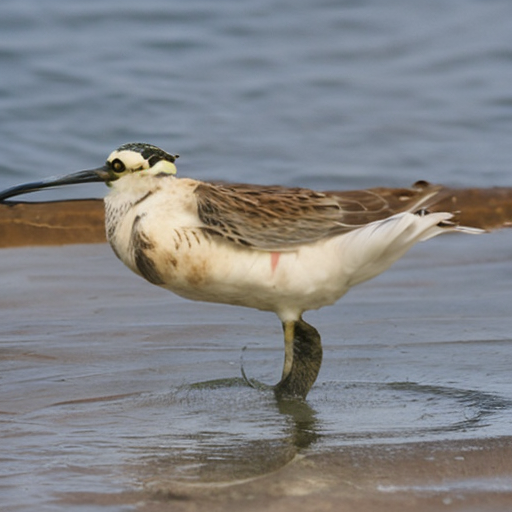}
     \includegraphics[width=\linewidth]{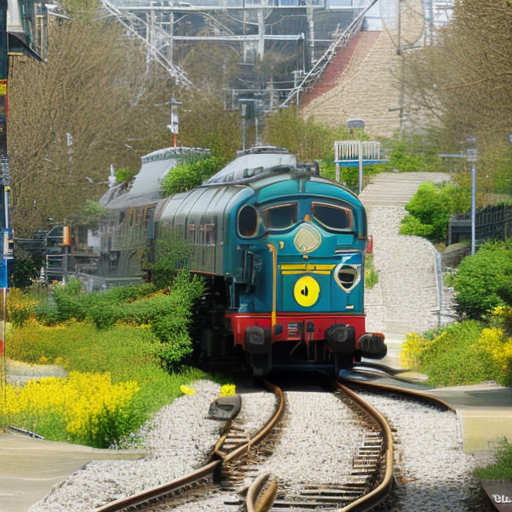}
     \includegraphics[width=\linewidth]{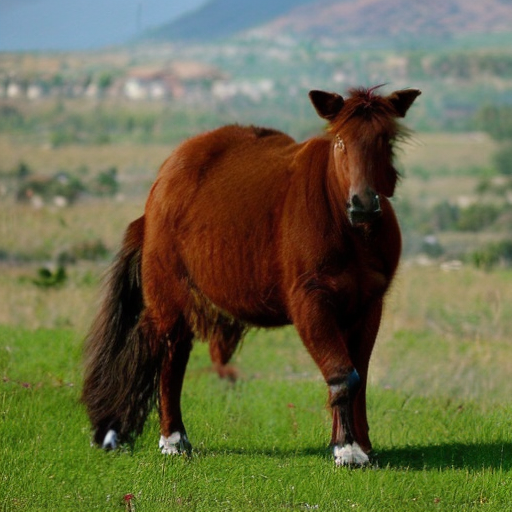}
     \end{minipage}
     }
     \hspace{-2.25mm}
     \subfloat[Ours]{
     \begin{minipage}{0.12\linewidth}
     \includegraphics[width=\linewidth]{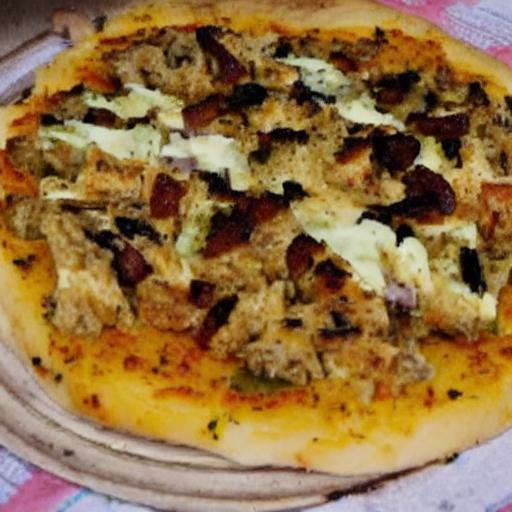}
     \includegraphics[width=\linewidth]{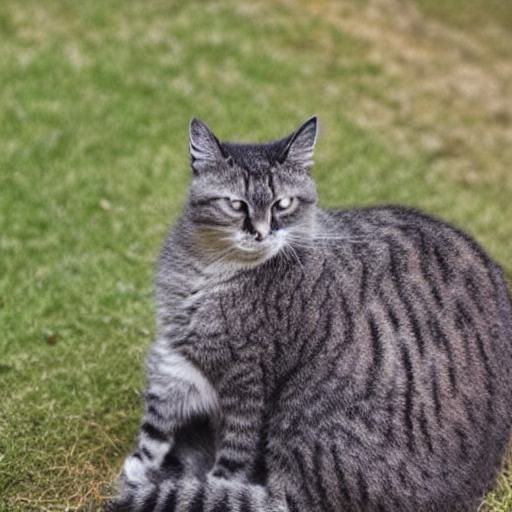}
     \includegraphics[width=\linewidth]{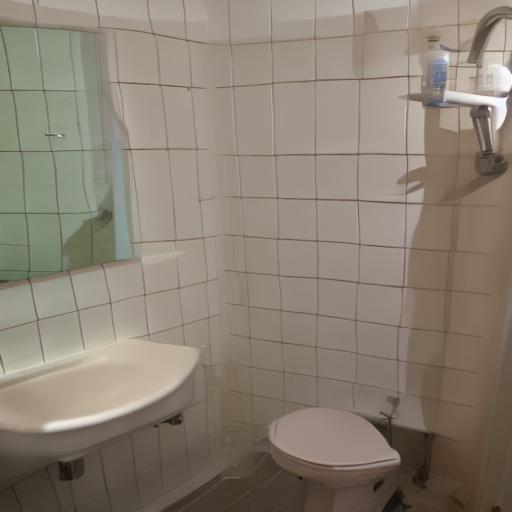}
     \includegraphics[width=\linewidth]{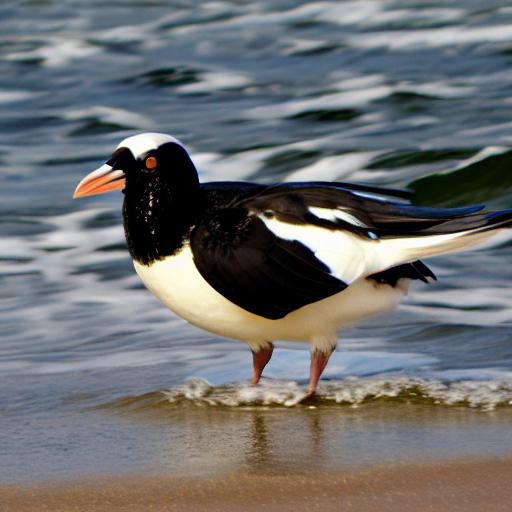}
     \includegraphics[width=\linewidth]{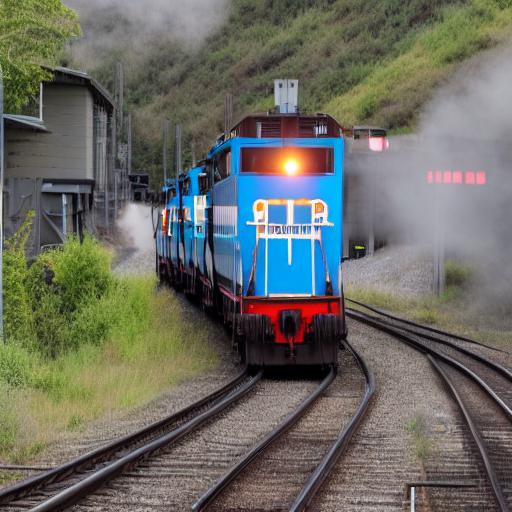}
     \includegraphics[width=\linewidth]{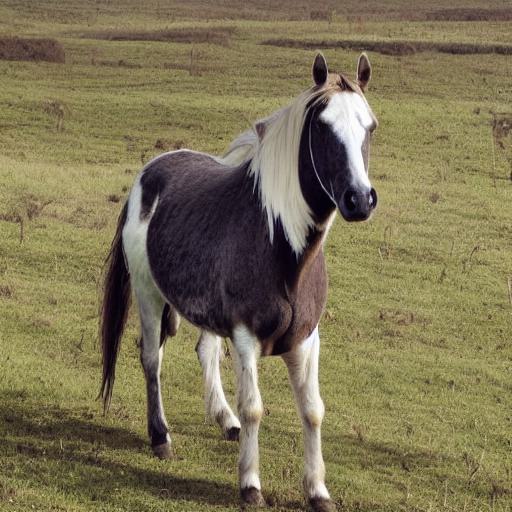}
     \end{minipage}
     }
     \hspace{-2.25mm}
     \subfloat[Stimulus]{
     \begin{minipage}{0.12\linewidth}
     \includegraphics[width=\linewidth]{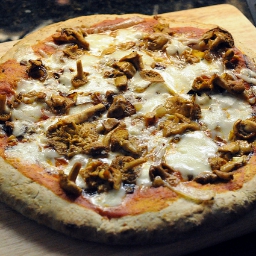}
     \includegraphics[width=\linewidth]{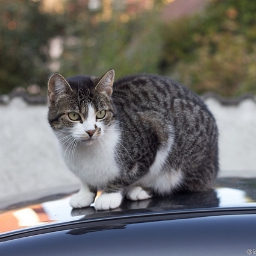}
     \includegraphics[width=\linewidth]{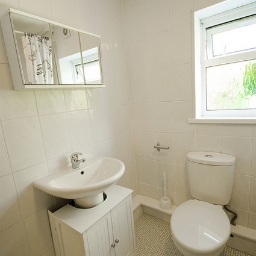}
     \includegraphics[width=\linewidth]{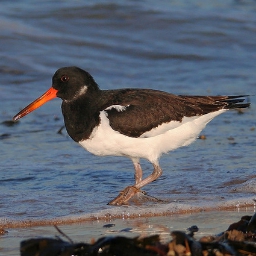}
     \includegraphics[width=\linewidth]{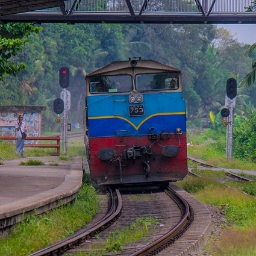}
     \includegraphics[width=\linewidth]{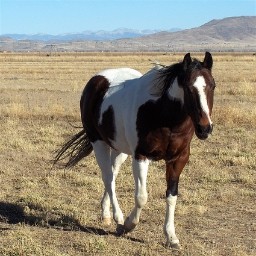}
     \end{minipage}
     }
\vspace{-2mm}
\caption{Qualitative comparison with competitors on mind decoding. Our reconstructed images are consistency with the stimulus images on semantic, structure, and appearance.}
\vspace{-5mm}
\label{fig:figure_main}
\end{figure*}

The qualitative comparison with other works can be seen in Fig.~\ref{fig:teaser} and Fig.~\ref{fig:figure_main}. We can observe that Takagi~\etal's method cannot reconstruct plausible images. By learning mappings to the versatile diffusion representation space~\cite{xu2023versatile}, other methods reconstruct realistic images but fail to achieve semantic and visual consistency with the stimulus images. For example, none of them successfully reconstruct \textit{``Pizza''} in the $1_{st}$ sample of Fig.~\ref{fig:figure_main}, and they also fail to reconstruct \textit{``Cat''} in the $2_{nd}$ sample. In contrast, our method captures the semantic information from fMRI voxels and reconstructs it correctly. Furthermore, our method also successfully captures the layouts of \textit{``Bathroom''}, which is contributed by our edge representation prediction, capturing the structural information hidden in the fMRI voxels. Finally, our method successfully reconstructs the stimulus image color patterns, such as the color of the \textit{``Bird''}, \textit{``Train''}, and \textit{``Horse''}, while none of the other methods recover the color of the \textit{``Train''} successfully. This demonstrates the effectiveness of our color palette prediction. Unlike those versatile diffusion-based methods that map the fMRI voxels to CLIP visual embeddings, we disentangle the structure and appearance into edge and color palettes, together with SRM and VCM, achieving a more faithful reconstruction.

\subsection{Adaptation on New Subject}
\label{sec:ans}
Our framework can be easily adapted to new subjects with few novel samples, which is valuable for real practice for reducing the number of fMRI-stimuli pairs. To simulate the scenarios of limited data, we follow MindBridge~\cite{wang2024mindbridge} that first trains the framework on three subjects (Subj01, 02, and 05), and adapts to the new subject (Subj07). Particularly, we only train the SBMM for the new subject while keeping the parameters in the other parameters fixed. We set 500, and 1,500 novel samples to adaptation. We also train a model from a sketch on a new subject with limited data. The adaptation results can be seen in Tab.~\ref{tab:adaptation}. We can see that with the increasing number of data samples, our method boosts its performance both for the adapted version and training from the sketch version. Our adapted results outperform the model trained from sketch, which demonstrates the effectiveness of our framework on new subject adaptation. Additionally, our framework outperforms MindBridge in most metrics. The framework learns subject-irrelevant mappings between fMRI and representation spaces with SBMM, making it robust and flexible to new subjects.

% \begin{table*}[h!]
% \caption{{Qualitative comparisons with variants on NSD dataset. All metrics are calculated as the average across 4 subjects.}}
% \vspace{-0.25cm}
% \footnotesize
% \centering
% \setlength{\tabcolsep}{2mm}{
% \begin{threeparttable}
% \begin{tabular}{c|c|cccc|cccc}
% % \toprule
% \toprule
% \multirow{2}{*}{{\textbf{Method}}}   &  \multirow{2}{*}{{\textbf{\# Models}}}  & \multicolumn{4}{c|}{\textbf{Low-Level}} & \multicolumn{4}{c}{\textbf{High-Level}} \\
% \cmidrule{3-10}&  &\textbf{PixCorr} $\uparrow$ & \textbf{SSIM} $\uparrow$     & \textbf{AlexNet(2)} $\uparrow$  &  \textbf{AlexNet(5)} $\uparrow$    & \textbf{Incep} $\uparrow$    & \textbf{CLIP}  $\uparrow$  & \textbf{EffNet-B}  $\downarrow$     & \textbf{SwAV}  $\downarrow$    \\

% \midrule

% Baseline &  4 & - &- &  - &  - &  78.2\% &  - & - &  -\\
% UM  &  4  &  -  &  -  &  83.0\%  &  83.0\%  &  76.0\%  &  77.0\%  &  --  &  --\\
% $w/o$ SBMM  &  1  &  .254  &  .356  &  94.2\%  &  96.2\%  &  87.2\%  &  91.5\%  &  .775  &  .423\\

% \midrule
% Direct Addition  &  1  &  .309  &  .323  &  94.7\%  &  97.8\%  &  93.8\%  &  94.1\%  &  .645  &  .367\\
% $w/o$ CSM &  1 & .150 &.325 &  - &  - &  - &  - & .862 &  .465 \\
% $w/o$ VFM &  1 & .080 &.220 &  72.1\% &  83.2\% &  78.8\% &  76.2\% & .854 &  .491 \\

% \midrule

% Ours  &  {1} & & &  & & & & & \\

% \bottomrule
% % \bottomrule
% \end{tabular}

% \end{threeparttable}}
% \label{tab:ablation}
% \vspace{-0.5cm}
% \end{table*} 

\begin{table*}[h!]
\caption{{Qualitative comparisons on new subject adaptation under various data limitation scenarios.}}
\vspace{-0.25cm}
% \footnotesize
\centering
\setlength{\tabcolsep}{0.5mm}{
\begin{threeparttable}
\begin{tabular}{c|c|c|cccc|cccc}
% \toprule
\hline
\multirow{2}{*}{{\textbf{Method}}}   & \multirow{2}{*}{{\textbf{\# Samples}}}  & \multirow{2}{*}{\makecell{\textbf{Adap-} \\ \textbf{tation?}}}  & \multicolumn{4}{c|}{\textbf{Low-Level}} & \multicolumn{4}{c}{\textbf{High-Level}} \\
\cline{4-11}&  & &{PixCorr} $\uparrow$ & {SSIM} $\uparrow$     & {AlexNet(2)} $\uparrow$  &  {AlexNet(5)} $\uparrow$    & {Incep} $\uparrow$    & {CLIP}  $\uparrow$  & {EffNet-B}  $\downarrow$     & {SwAV}  $\downarrow$    \\

\hline

MindBridge  &  500  &  ✗    &  .079  &  .171  &  73.5\%    &  83.3\%    &  74.4\%  &  80.1\%  &  .894  &  .587\\
\rowcolor{gray!20}MindBridge  &  500  &   ✓             &  .112  &  .229  &  {79.6\%}  &  {85.0\%}  &  {82.3\%}  &  {86.7\%}  &  {.840}  &  {.521}\\
Ours    &  500  &  ✗     &  .083  &.213    &  74.2\%    &  85.2\%    &  76.2\%  & 84.3\%    & .855  &  .572\\
\rowcolor{gray!20}Ours    &  500  &   ✓              &  \textbf{.145}     &  \textbf{.234 } &  \textbf{82.1\%}  &  \textbf{88.1\%} & \textbf{ 85.4\%}    &\textbf{89.1\%}   &  \textbf{.821}  &  \textbf{.503}\\
\hline

MindBridge  &  1,500  &  ✗  &  .107  &  .206  &  79.4\%  &  90.0\%  &  82.4\%  &  87.2\%  &  .844  &  .523\\
\rowcolor{gray!20}MindBridge  &  1,500  &   ✓           &  {.140}&  {.250}  &  {84.6\%}  &  \textbf{92.6\%}  &  {85.8\%}  &  {91.0\%}  &  \textbf{.796}  &  {.485}\\
Ours  &  1,500  &  ✗   &  .134  &  .242  &  82.1\%   &   91.2\%  & 83.4\%  & 88.9\%  &  .822  &  .513\\
\rowcolor{gray!20}Ours &  1,500  &   ✓            &  \textbf{.152}     &\textbf{.267}  &  \textbf{85.2\%}  &  92.3\% &  \textbf{86.1\%}   &\textbf{92.1\% }   &  {.814} &  \textbf{.491}\\
% \hline

% \midrule
% \midrule

% MindBridge  &  4,000  &  ✗  &  .114  &  .232  &  81.4\%  &  92.2\%  &  85.3\%  &  89.8\%  &  .815  &  .491\\
% MindBridge  &  4,000  &   ✓           &  {.156}  &  {.258}  &  {85.7\%}  &  {94.1\%}  &  \textbf{88.9\%}  &  {92.5\%}  &  {.765}  &  {.458}\\
% BAI(Ours)  &  4,000  &  ✗   &  .141  &.247  & 83.1\%  &  93.9\%  & 84.4\%  & 90.2\%  &  .809  &  .497\\
% BAI(Ours)  &  4,000  &   ✓            &  \textbf{.168}     &\textbf{.283}  &  \textbf{87.4\% } & \textbf{ 95.4\%} & {88.3\% }   &\textbf{93.2\%}    &  \textbf{.753}  &  \textbf{.431}\\

\hline
\end{tabular}
\end{threeparttable}}
\label{tab:adaptation}
\vspace{-2.5mm}
\end{table*} 

% \begin{table*}[!h]
% \caption{{Qualitative comparisons with variants on NSD dataset. All metrics are calculated as the average across 4 subjects.}}
% \vspace{-0.25cm}
% % \footnotesize
% \centering
% \setlength{\tabcolsep}{1mm}{
% \begin{threeparttable}
% \begin{tabular}{c|c|cccc|cccc}
% % \toprule
% \toprule
% \multirow{2}{*}{{\textbf{Method}}}   &  \multirow{2}{*}{\makecell{\textbf{Cross} \\ \textbf{Subject?}}}  & \multicolumn{4}{c|}{\textbf{Low-Level}} & \multicolumn{4}{c}{\textbf{High-Level}} \\
% \cline{3-10}&  &\textbf{PixCorr} $\uparrow$ & \textbf{SSIM} $\uparrow$     & \textbf{AlexNet(2)} $\uparrow$  &  \textbf{AlexNet(5)} $\uparrow$    & \textbf{Incep} $\uparrow$    & \textbf{CLIP}  $\uparrow$  & \textbf{EffNet-B}  $\downarrow$     & \textbf{SwAV}  $\downarrow$    \\

% \hline
% Baseline &  \Large{$\times$} &.194 &.298 & 84.2\% &  91.9\% &  89.5\% &  90.1\% & .737 &  .455\\
% UM  &  { ✓}  &  .200  &  .301  & 87.0\%  &  93.4\%  &  90.0\%  &  90.3\%  & .719  &  .436\\
% $w/o$ SBMM  &  ✓  &  .166  &  .274  &  83.7\%  &  90.9\%  &  86.6\%  &  87.2\%  &  .751  &  .460\\
% % \hline
% Direct Addition  &   ✓  &   .201  &  .231  &  79.0\%  &  85.8\%  & 80.2\%  & 82.8\%  & .866  &  .586\\

% $w/o$ SRM &   ✓ & .171 &.269 & 82.4\% &  89.0\% &  85.0\% &  86.7\% & .804 & .517 \\
% $w/o$ VCM &   ✓ & .223 & .259 & 85.7\% &  91.2\% &  88.2\% &  89.7\% & .777 &  .496 \\
% \hline
% BAI   &   ✓ &.192 &.297 &87.1\%  &93.1\% &89.6\% &90.0\% &.721 &.439 \\

% \bottomrule
% \end{tabular}

% \end{threeparttable}}
% \label{tab:ablation}
% \vspace{-0.5cm}
% \end{table*} 

\begin{table*}[!h]
\caption{{Qualitative comparisons with variants on NSD dataset. All metrics are calculated as the average across 4 subjects.}}
\vspace{-0.25cm}
% \footnotesize
\centering
\setlength{\tabcolsep}{0.325mm}{
\begin{threeparttable}
\begin{tabular}{c|c|cccc|cccc}
% \toprule
\hline
\multirow{2}{*}{{\textbf{Method}}}   &  \multirow{2}{*}{\makecell{\textbf{Cross} \\ \textbf{Subject?}}}  & \multicolumn{4}{c|}{\textbf{Low-Level}} & \multicolumn{4}{c}{\textbf{High-Level}} \\
\cline{3-10}&  &{PixCorr} $\uparrow$ & {SSIM} $\uparrow$     & {AlexNet(2)} $\uparrow$  &  {AlexNet(5)} $\uparrow$    & {Incep} $\uparrow$    & {CLIP}  $\uparrow$  & {EffNet-B}  $\downarrow$     & {SwAV}  $\downarrow$    \\

\hline
\rowcolor{gray!20}UM  &  { ✓}  &  .281  &  .301  & 92.0\%  &  93.4\%  &  93.0\%  &  90.3\%  & .719  &  .436\\

$w/o$ SBMM  &  ✓  &  .266  &  .274  &  83.7\%  &  91.9\%  &  89.6\%  &  87.2\%  &  .751  &  .460\\

\hline

\rowcolor{gray!20}\textit{Direct Addition}  &   ✓  &  .228  & .259   &  83.1\%  &  89.8\%  & 87.2\%  & 89.8\%  & .765  &  .477\\

\textit{Direct Addition} + SRM &       ✓ & .252 & .281 & 89.7\% &  94.2\% &  94.2\% &  95.7\% & .671 &  .374 \\
\rowcolor{gray!20}\textit{Direct Addition} + VCM &       ✓ & .298 &.287 & 94.4\% &  96.0\% &  92.1\% &  91.7\% & .714 & .391 \\

\hline

Our-SS &  ✗  &\textbf{.321} &.341 & 97.5\% &  98.9\% &  96.5\% &  97.1\% & \textbf{.637} &  \textbf{.355}\\
\rowcolor{gray!20}Our  &  ✓  &{.318} &\textbf{.356} &\textbf{97.3\%}  &\textbf{98.8\%} &\textbf{96.7\%} &\textbf{97.5\%} &{.639} &{.345} \\

\hline
\end{tabular}

\end{threeparttable}}
\label{tab:ablation}
\vspace{-5mm}
\end{table*}

\subsection{Ablation Studies}

In this section, we conduct ablation experiments to analyze our framework. We first analyze the effectiveness of our bidirectional mapping by proposing a Unidirectional Mapping (UM) from fMRI voxels to representations. In this variant, we remove the decoder of fMRI $\mathcal{E}_v$, the encoder of representations $\mathcal{D}_v$, and the MLP $\mathcal{MLP}_{R{\Rightarrow}V}(\cdot)$ that maps the representation features to the fMRI features. This variant is trained exclusively using translation losses $\mathcal{L}^\text{Tr}_{V}$, $\mathcal{L}^\text{Tr}_{S}$, and $\mathcal{L}^\text{Tr}_{E}$. To evaluate the effectiveness of SBMM on cross-subject decoding, we propose the variant $w/o$ SBMM, by removing this module. Moreover, to demonstrate the influence of inaccurate representation predictions, we introduce the variant \textit{Direct Addition}, by removing the SRM and VCM. In this variant, we feed the predicted representations to ControlNet directly and fuse the two conditional features by simple addition using Eq.~\ref{eq:controlnet1}. To demonstrate the effectiveness of each module, we also propose variants \textit{Direct Addition} + SRM and \textit{Direct Addition} + VCM, by adding a specific module on \textit{Direct Addition}. Finally, we train subject-specific models of our framework (Ours-SS) for each subject.

The quantitative comparison of various variants can be seen in Tab.~\ref{tab:ablation}. We can see that variant UM performs worse than the final framework, as UM learns the unidirectional mapping from fMRI voxels to representations, leading to inaccurate predicted representations. Although our SRM and VCM modules further reduce the need for accurate representations, images reconstructed by UM still suffer from low fidelity. The variant $w/o$ SBMM performs much worse than our final framework, which demonstrates its effectiveness on cross-subject mind decoding. Without this module, the framework cannot learn informative representations from multiple subjects due to individual differences. 
Variant \textit{Direct Addition} gets the worst on all metrics, which evidences the necessity of tolerating inaccurate representations in the second decoding stage. As discussed in Sec.~\ref{sec:rnp}, it requires accurate representations for fidelity reconstruction, and any inaccurate representations will mislead the reconstruction process. In contrast, our two proposed modules, SRM and VCM, improve the quality of the reconstruction from semantic and visual perspectives. Adding SRM to \textit{Direct Addition} increases semantic-related performances, such as the CLIP score. Compared to \textit{Direct Addition}, variant \textit{Direct Addition} + VCM improves the low-level metrics effectively. Finally, our cross-subject variant achieves similar performance to the Ours-SS, demonstrating the effectiveness of our framework in capturing subject-irrelevant representations. The qualitative comparison of various variants can be seen in the supplementary.

\section{Conclusion}
\label{sec:conclusion}

In this paper, we introduce a cross-subject mind decoding framework that reconstructs stimulus images from fMRI voxels. We leverage the bidirectional mappings between fMRI voxels and semantic/visual representations. Combined with the subject bias modulation module, our method effectively captures complex relationships between these two domains across different subjects. To further improve the fidelity of decoded images, we propose semantic refinement and visual coherence modules. These modules reduce the dependency on highly precise image representations in the decoding stage. Extensive evaluations of the NSD dataset demonstrate that our framework outperforms prior methods in mind decoding and can be easily adapted to new subjects with minimal additional data.

{\textbf{Acknowledgment:} This project is supported by National Natural Science Foundation of China (Grant 62102381), and the Guangdong Natural Science Funds for Distinguished Young Scholars (Grant 2023B1515020097), the National Research Foundation, Singapore under its AI Singapore Programme (AISG Award No.: AISG3-GV-2023-011), the Singapore Ministry of Education AcRF Tier 1 Grant (Grant No.: MSS25C004), and the Lee Kong Chian Fellowships. Tingting Zhu was supported by the Royal Academy of Engineering under Research Fellowship scheme.}

%%%%%%%%% REFERENCES
{\small
\bibliographystyle{ieeenat_fullname}
\bibliography{./egbib}
}

\clearpage
\section{Appendix} 

\textbf{Analysis of Shared Latent Space} 

\begin{figure}[t]
    \centering
    \captionsetup[subfloat]{justification=centering}
     \subfloat[Subject-Specific Representations]{
     \begin{minipage}{0.45\linewidth}
     \includegraphics[width=\linewidth]{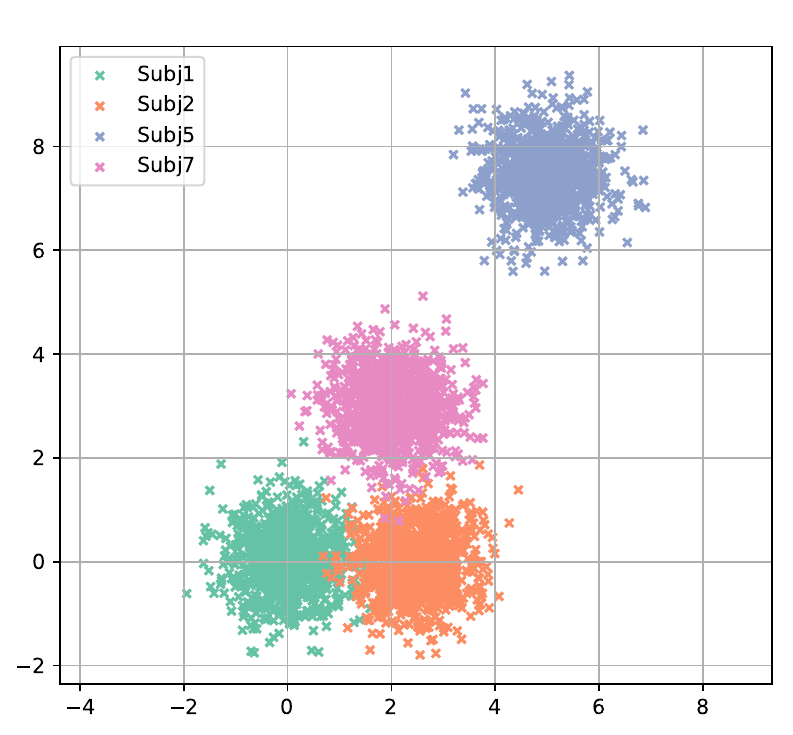}
     \end{minipage}
     }
     \hspace{-2.25mm}
     \subfloat[Cross Subject Representations]{
     \begin{minipage}{0.45\linewidth}
     \includegraphics[width=\linewidth]{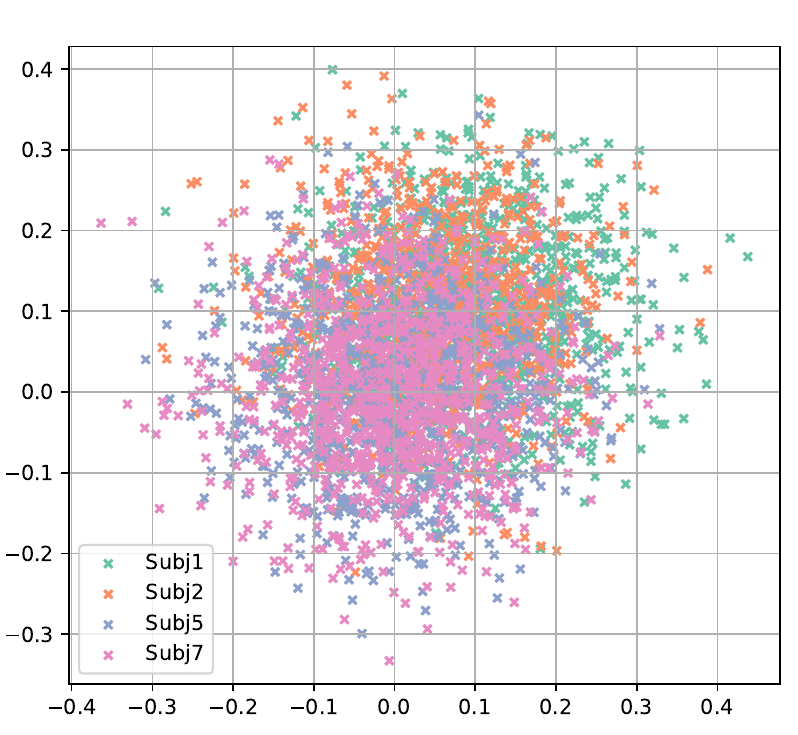}
     \end{minipage}
     }
\vspace{-2mm}
\caption{t-SNE visualizations of representations learned by subject-specific and cross subject decoding.}
\label{fig:tsne}
\end{figure}

Our framework learns a shared latent space that aligns fMRI and visual features across subjects, enabling generalization by capturing subject invariant patterns. We present the t-SNE visualization of subject-specific and cross-subject representations in Fig.~\ref{fig:tsne}, t-SNE visualizations reveal tighter clustering of cross-subject representations, indicating better alignment across different subjects.

\textbf{Neuroscience Interpretability} 

We further investigate the neuroscience interpretability of our model by analyzing voxel-level gradients derived from internal representations. As illustrated in Fig.~\ref{fig:gradient}, the results indicate that the Low-level Visual Cortex (LVC) predominantly supports edge decoding, while the High-level Visual Cortex (HVC) is more involved in semantic processing. Both regions contribute to color prediction. These findings suggest that our shared representational space captures and preserves the hierarchical structure of visual processing across subjects.

\textbf{Failure cases and Limitation}

\begin{figure}[!t]
    \centering
    \captionsetup[subfloat]{justification=centering}
     \subfloat[{Prediction}]{
     \begin{minipage}{0.4\linewidth}
     \includegraphics[width=\linewidth]{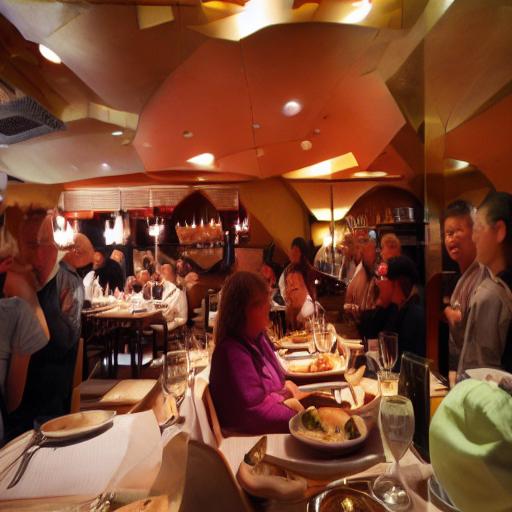}
     \includegraphics[width=\linewidth]{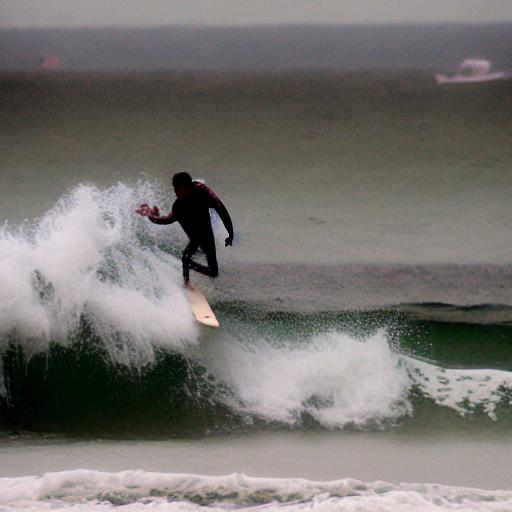}
     \end{minipage}
     }
     \subfloat[Stimulus]{
     \begin{minipage}{0.4\linewidth}
     \includegraphics[width=\linewidth]{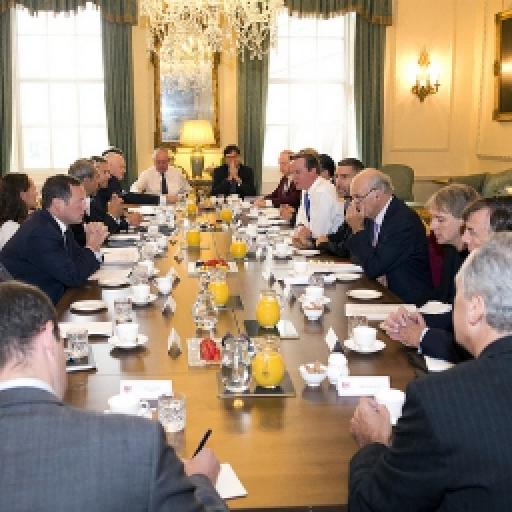}
     \includegraphics[width=\linewidth]{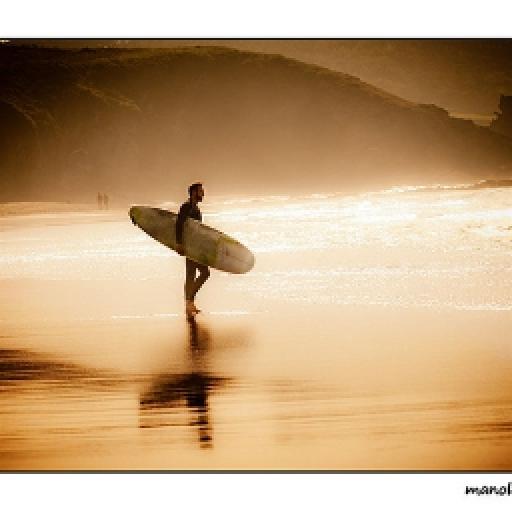}
     \end{minipage}
     }
\vspace{-2mm}
\caption{Failure cases of our method.}
% \vspace{-3mm}
\label{fig:fail}
\end{figure}

\begin{figure}[!h]
    \centering
    \includegraphics[width=.65\linewidth]{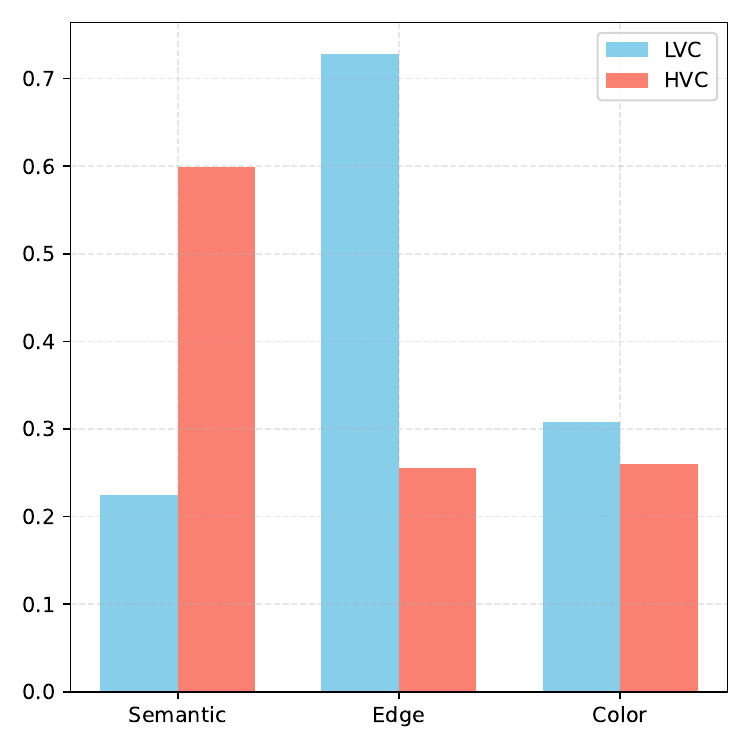}
\vspace{-3mm}
\caption{Voxel-level gradient analysis of visual features across brain regions.}
\label{fig:gradient}
\end{figure}

We present the failure cases of our method in Fig.~\ref{fig:fail}, our method inherits the limitation of SD, which cannot handle the complex scenes~\cite{yang2023law,liu2024draw}. Additionally, it also fails when encountering unnatural colors.

{One limitation of our work is that the SBMM is a subject-dependent component, which needs to be retrained for every new subject. This is due to the significant inter-subject variability and limited large-scale datasets, resulting in the requirement of subject-specific components, which is a shared limitation of current cross-subject studies~\cite{wang2024mindbridge,scotti2024mindeye2,quan2024psychometry,xia2024umbrae}.} 
Although not our purpose, our method could be misused for privacy invasion or other unethical purposes. Thus, strict and responsible data privacy protections must be established.

\begin{table*}[!h]
\caption{{Qualitative comparisons with other methods on three datasets.}}
\centering
\setlength{\tabcolsep}{7pt}
\begin{threeparttable}
\begin{tabular}{l|ccc|ccc|ccc}
\hline
\multirow{2}{*}{\textbf{Method}}& \multicolumn{3}{c|}{\textbf{NOD}}  & \multicolumn{3}{c|}{\textbf{GOD}} & \multicolumn{3}{c}{\textbf{BOLD5000}} \\
                               & Acc (\%) & PCC & SSIM & Acc (\%) & PCC & SSIM & Acc (\%) & PCC & SSIM \\
\hline
IC-GAN~\cite{ozcelik2022reconstruction}   &-  &-  &-  & 29.39  & 0.449 & 0.545 & -       & -       & - \\
\rowcolor{gray!20}MinD-Vis~\cite{chen2023seeing} &-  &-  &-  & 26.64  & 0.532 & 0.527 & 25.918  & 0.545 & 0.524 \\
CMVDM~\cite{zeng2024controllable}    &-  &-  &-  & {30.11}  & {0.768} & 0.632 & {27.791}  & {0.557} & {0.535} \\
\hline
\rowcolor{gray!20}Ours     &\textbf{35.12} &\textbf{0.734} & \textbf{0.745}  & \textbf{34.311}  & \textbf{0.794} & 0.704 & \textbf{29.088}  & \textbf{0.583} & \textbf{0.553} \\
\hline
\end{tabular}
\end{threeparttable}
\label{table3}
\end{table*}

\textbf{Details of Evaluation Metrics}

We use 8 evaluation metrics for the quantitative comparison from low and high levels. \textbf{PixCorr} measures the pixel-wise correlation of decoded and GT images, \textbf{SSIM} measures the structure similarity between two images~\cite{wang2004image}. \textbf{AlexNet(2)} is the two-way comparison of image features extracted from the second layer of AlexNet~\cite{krizhevsky2012imagenet}, and \textbf{AlexNet(5)} compares the features extracted from the fifth layer. The above four metrics evaluate the low-level similarity of reconstructed images. The high-level metrics including \textbf{Inception}, \textbf{CLIP}, \textbf{EffNet-B}, and \textbf{SwAV}. \textbf{Inception} is the two-way comparison of the features extracted from the last pooling layer of InceptionV3~\cite{szegedy2016rethinking}, CLIP compares the cosine similarity between the features extracted from the CLIP image encoder~\cite{radford2021learning}. \textbf{EffNet-B} and \textbf{SwAV} are distance metrics based on EfficientNet-B1~\cite{tan2019efficientnet} and SwAV-ResNet50~\cite{caron2020unsupervised}, respectively.

\textbf{Structure Details}

Our SRM adopts the Querying Transformer~\cite{li2023blip} architecture, comprising 12 hidden layers. As illustrated in the middle-right panel of Fig.~2 in the main paper, the predicted semantic representation $\tilde{S}$ is incorporated into the cross-attention layers of the Querying Transformer.

The VCM consists of 13 layers, each designed for different resolutions. Each layer includes three \texttt{Conv2D} layers with \texttt{SiLU} activation functions between them, and the final output is activated using a \texttt{Sigmoid} function.

The pseudo-code for the BAI framework is provided in Alg.~\ref{alg:bai}.

\begin{algorithm}[t]
\small
\caption{\small Structure details of BAI.
}
\label{alg:bai}

\definecolor{codeblue}{rgb}{0.25,0.5,0.5}
\definecolor{codegreen}{rgb}{0,0.6,0}
\definecolor{codekw}{RGB}{207,33,46}
\lstset{
  backgroundcolor=\color{white},
  basicstyle=\fontsize{7.5pt}{7.5pt}\ttfamily\selectfont,
  columns=fullflexible,
  breaklines=true,
  captionpos=b,
  commentstyle=\fontsize{7.5pt}{7.5pt}\color{codegreen},
  keywordstyle=\fontsize{7.5pt}{7.5pt}\color{codekw},
  escapechar={|}, 
}
\begin{lstlisting}[language=python]
class BAI:
    # Shared Encoder
    shared_encoder = Sequential([
        Linear(8192, 1024),
        LayerNorm((1024,))
    ])

    # SBMM in Encoders
    encoder_alpha_subj = ModuleDict({
        subj_id: Sequential([
            Linear(1024, 1024)
        ]) for subj_id in [1, 2, 5, 7]
    })

    encoder_beta_subj = ModuleDict({
        subj_id: Sequential([
            Linear(1024, 1024)
        ]) for subj_id in [1, 2, 5, 7]
    })

    # SBMM in Decoders
    decoder_alpha_subj = ModuleDict({
        subj_id: Sequential([
            Linear(1024, 1024)
        ]) for subj_id in [1, 2, 5, 7]
    })

    decoder_beta_subj = ModuleDict({
        subj_id: Sequential([
            Linear(1024, 1024)
        ]) for subj_id in [1, 2, 5, 7]
    })

    # Shared Decoder
    shared_decoder = Sequential([
        Linear(1024, 1024),
        LayerNorm((1024,)),
        Linear(1024, 8192)
    ])

    # Edge Prediction from Voxel Features
    vox2edge = Sequential([
        FC2Img(ConvTransposeAndResNet()),  # Custom architecture combining ConvTranspose and ResNet blocks
        Sigmoid()
    ])
    # Color Prediction from Voxel Features
    vox2color = Sequential([
        FC2Img(ConvTransposeAndResNet()),
        Tanh()
    ])
    # Text Prediction from Voxel Features
    vox2text = Sequential([
        Linear(1024, 1024),
        ResMLP([MLPBlock(1024) for _ in range(2)]),
        Linear(1024, vocab_size)  # E.g., 59136
    ])


    # Voxel Prediction from Edge
    edge2vox = Img2FC(ConvAndPoolingBlocks())

    # Voxel Prediction from Color
    color2vox = Img2FC(ConvAndPoolingBlocks())

    # Voxel Prediction from Semantic
    text2vox = Sequential([
        Linear(vocab_size, 1024),
        LayerNorm((1024,)),
        Linear(1024, 1024)
    ])


    # Translator MLP from Voxels to Representation
    translator2Rep = Sequential([
        ResMLP([MLPBlock(1024) for _ in range(4)])
    ])

    # Translator MLP from Representation to Voxels
    translator2Vox = Sequential([
        ResMLP([MLPBlock(1024) for _ in range(4)])
    ])

\end{lstlisting}

\end{algorithm}

\textbf{Evaluation on More Datasets}

We also extend our framework on other mind decoding benchmark, including NOD~\cite{gong2023large}, GOD~\cite{horikawa2017generic}, and BOLD5000~\cite{chang2019bold5000} datasets. We evaluate our method on three datasets using the same metrics as CMVDM~\cite{zeng2024controllable}. As shown in Tab.~\ref{table3}, our method consistently outperforms existing baselines across all datasets and metrics.

\textbf{More Qualitative Comparisons with Competitors}

\begin{figure*}[!t]
    \centering
    \captionsetup[subfloat]{labelformat=empty,justification=centering}
     \subfloat[Takagi~\etal~\cite{takagi2023high}]{
     \begin{minipage}{0.12\linewidth}
     \includegraphics[width=\linewidth]{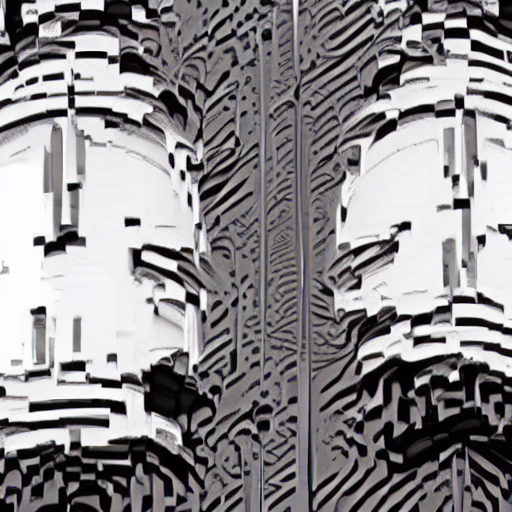}
     \includegraphics[width=\linewidth]{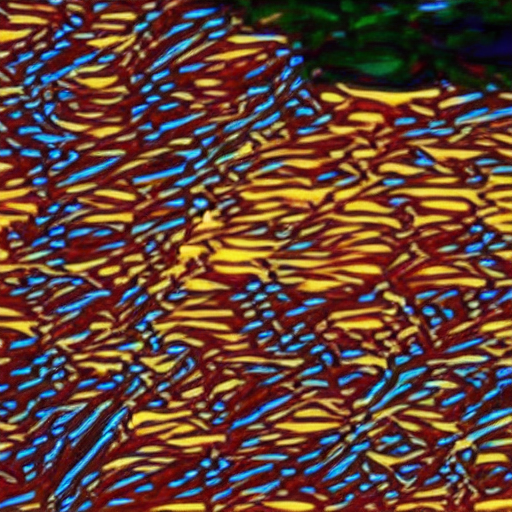}
     \includegraphics[width=\linewidth]{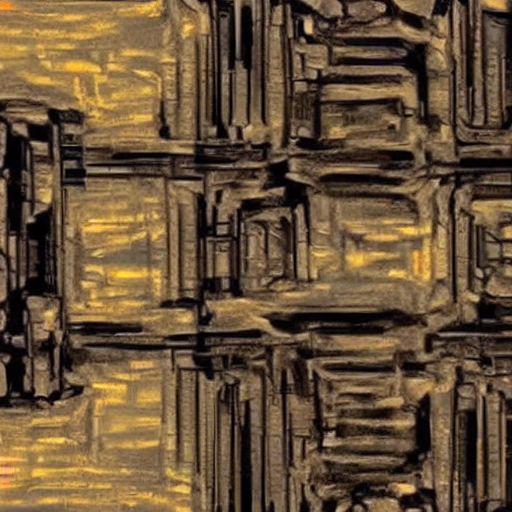}
     \includegraphics[width=\linewidth]{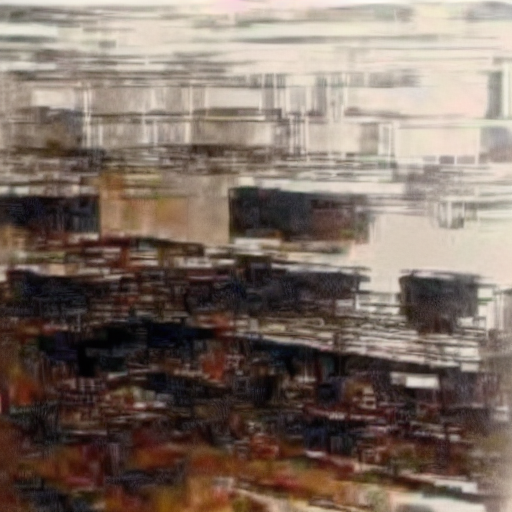}
     \includegraphics[width=\linewidth]{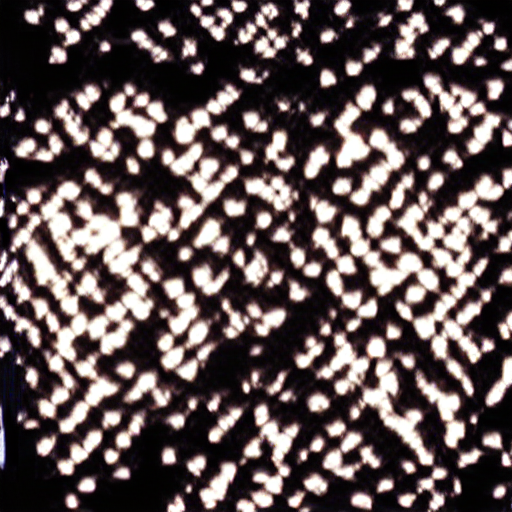}
     \includegraphics[width=\linewidth]{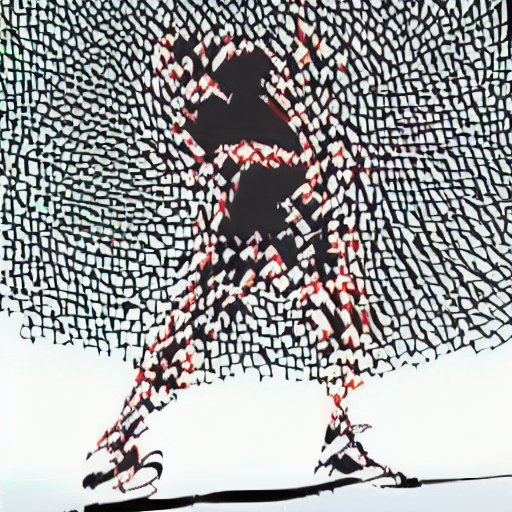}
     \includegraphics[width=\linewidth]{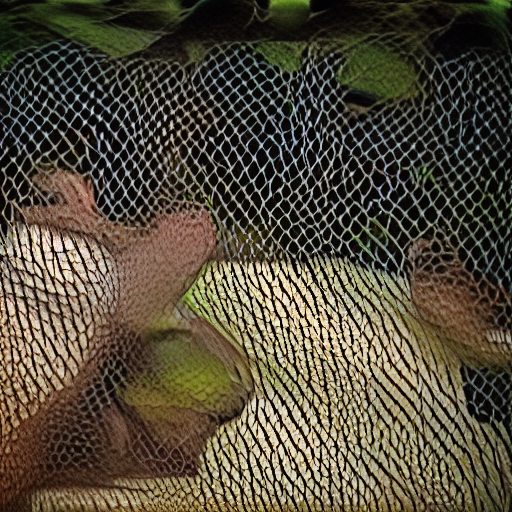}
     \includegraphics[width=\linewidth]{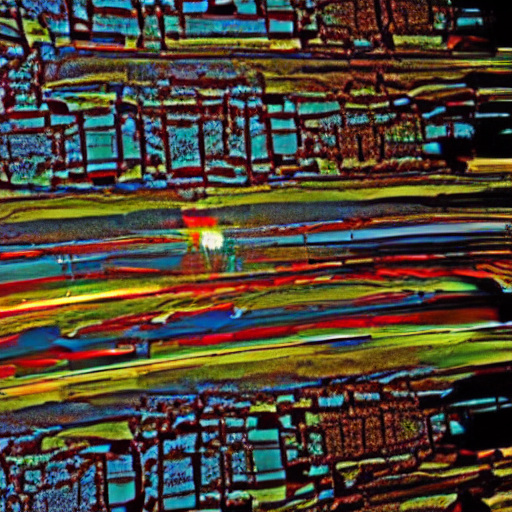}
     \includegraphics[width=\linewidth]{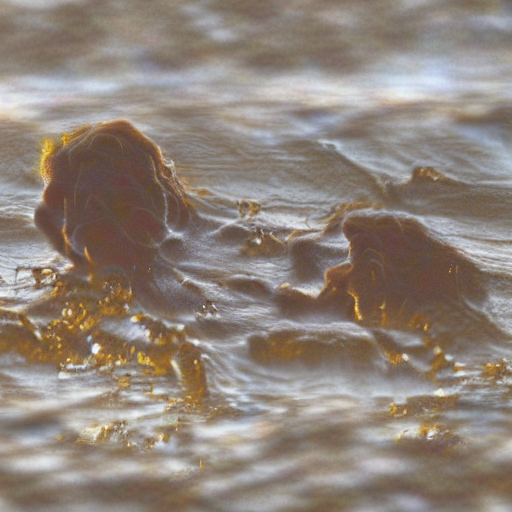}
     \end{minipage}
     }
     \hspace{-2.25mm}
     \subfloat[BrainDiffuser~\cite{ozcelik2023natural}]{
     \begin{minipage}{0.12\linewidth}
     \includegraphics[width=\linewidth]{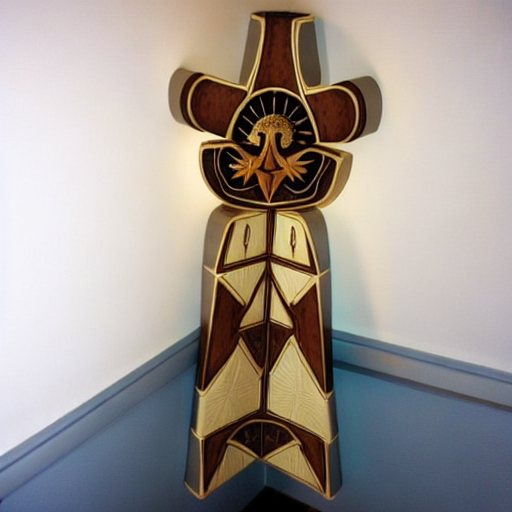}
     \includegraphics[width=\linewidth]{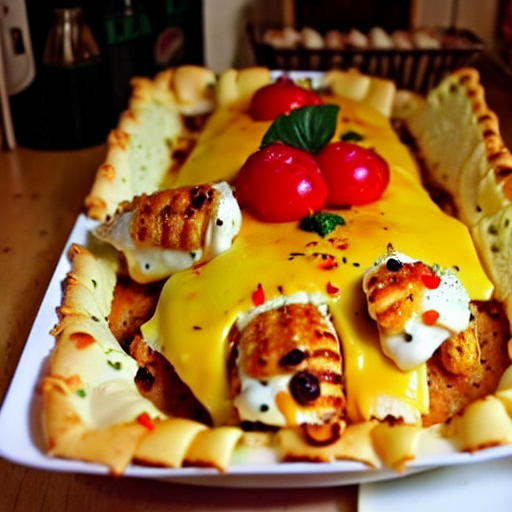}
     \includegraphics[width=\linewidth]{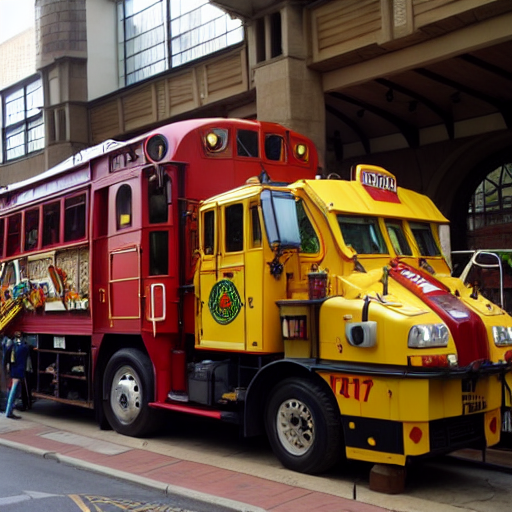}
     \includegraphics[width=\linewidth]{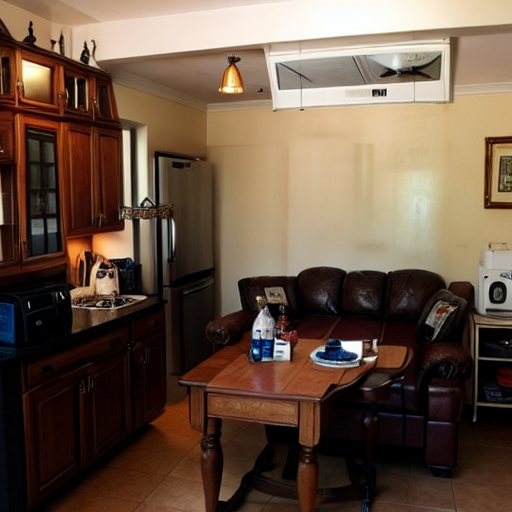}
     \includegraphics[width=\linewidth]{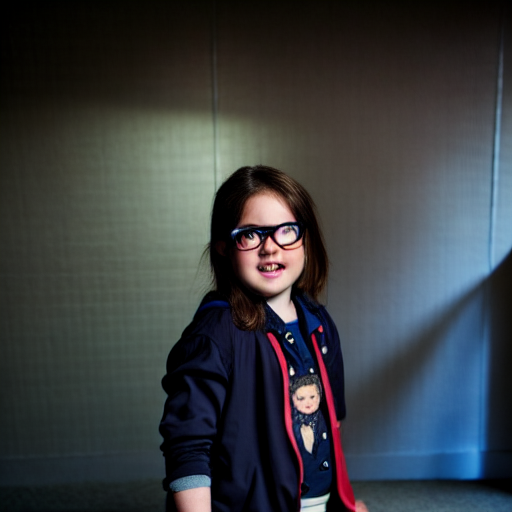}
     \includegraphics[width=\linewidth]{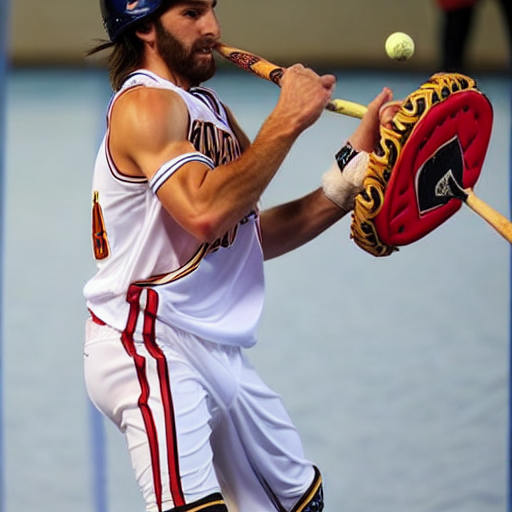}
     \includegraphics[width=\linewidth]{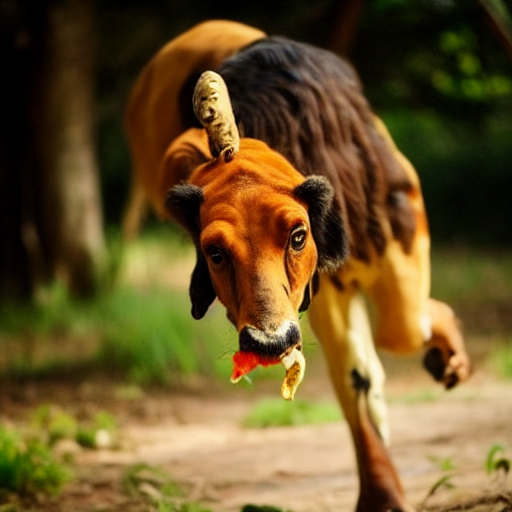}
     \includegraphics[width=\linewidth]{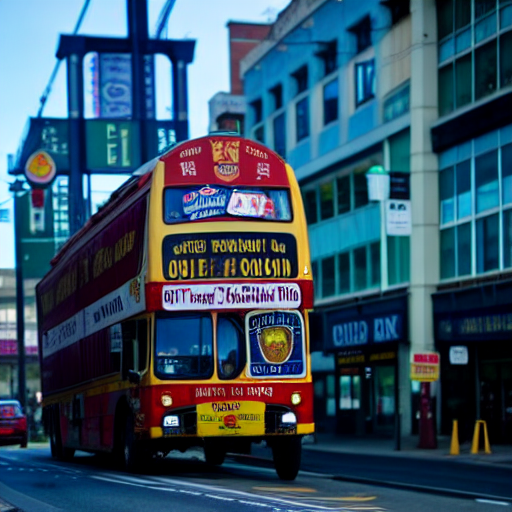}
     \includegraphics[width=\linewidth]{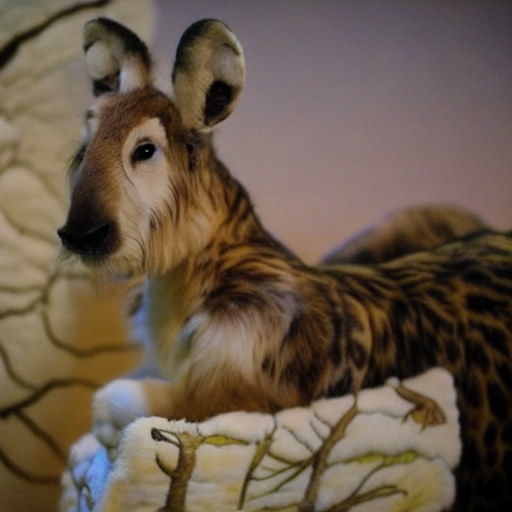}
     \end{minipage}
     }
     \hspace{-2.25mm}
     \subfloat[MindEye1~\cite{scotti2024reconstructing}]{
     \begin{minipage}{0.12\linewidth}
     \includegraphics[width=\linewidth]{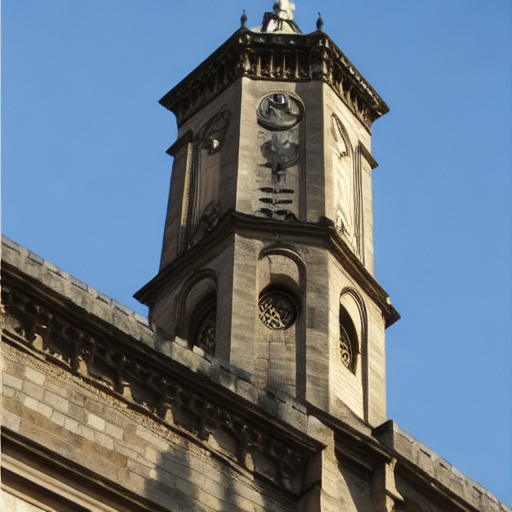}
     \includegraphics[width=\linewidth]{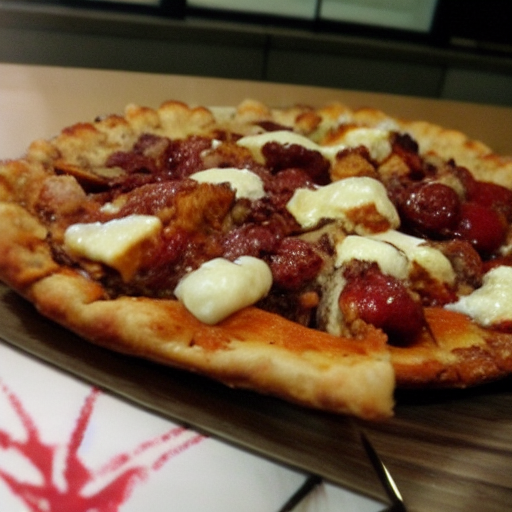}
     \includegraphics[width=\linewidth]{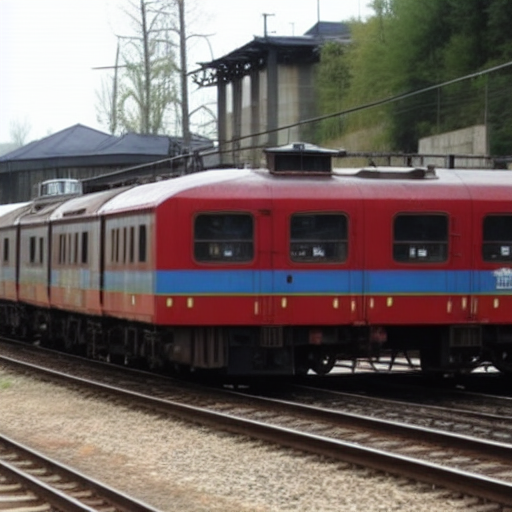}
     \includegraphics[width=\linewidth]{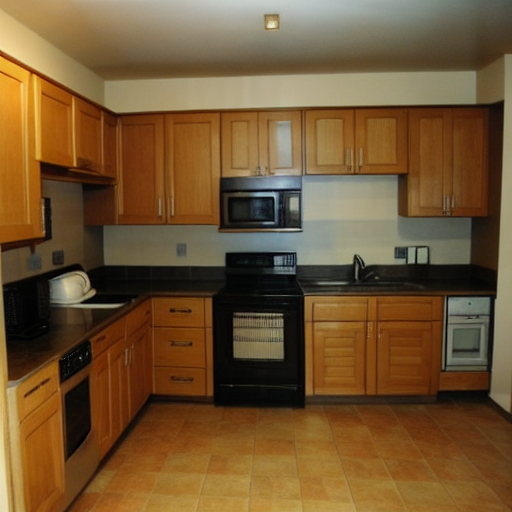}
     \includegraphics[width=\linewidth]{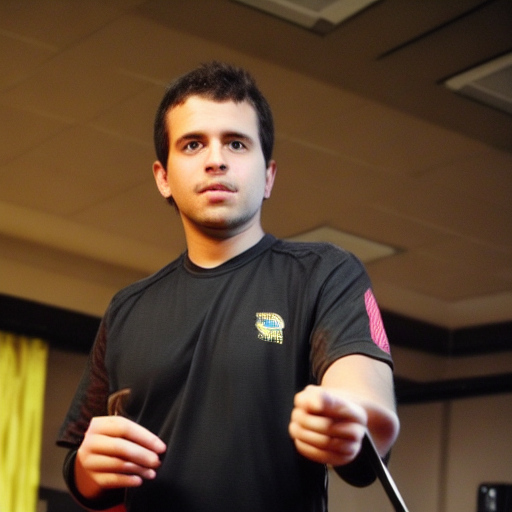}
     \includegraphics[width=\linewidth]{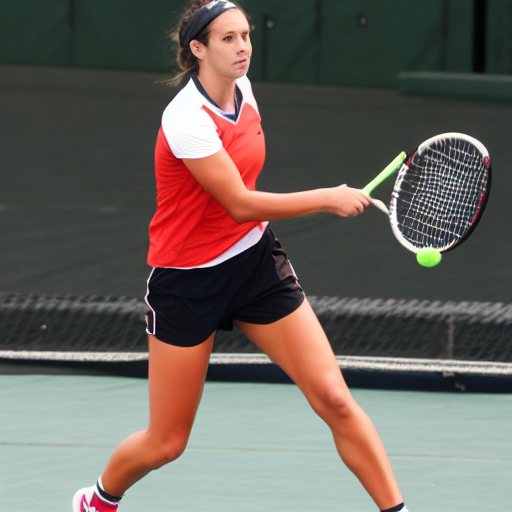}
     \includegraphics[width=\linewidth]{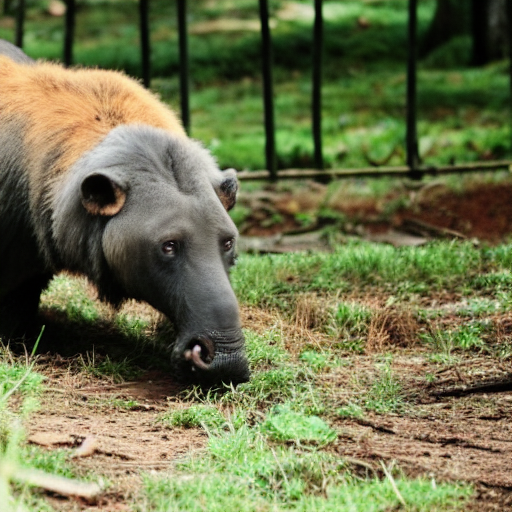}
     \includegraphics[width=\linewidth]{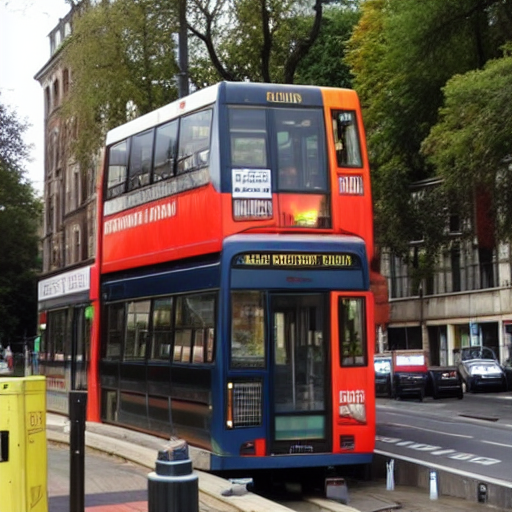}
     \includegraphics[width=\linewidth]{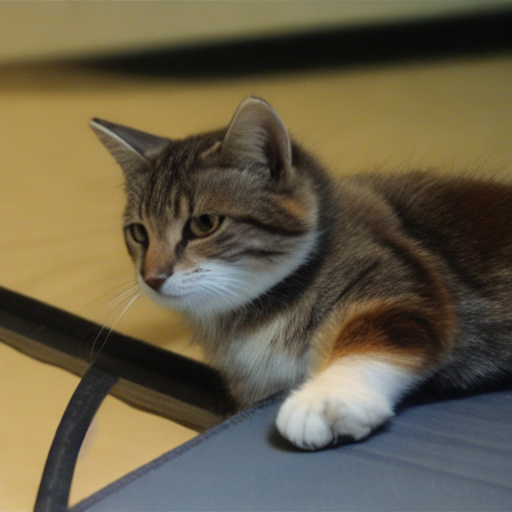}
     \end{minipage}
     }
     \hspace{-2.25mm}
     \subfloat[MindBridge~\cite{wang2024mindbridge}]{
     \begin{minipage}{0.12\linewidth}
     \includegraphics[width=\linewidth]{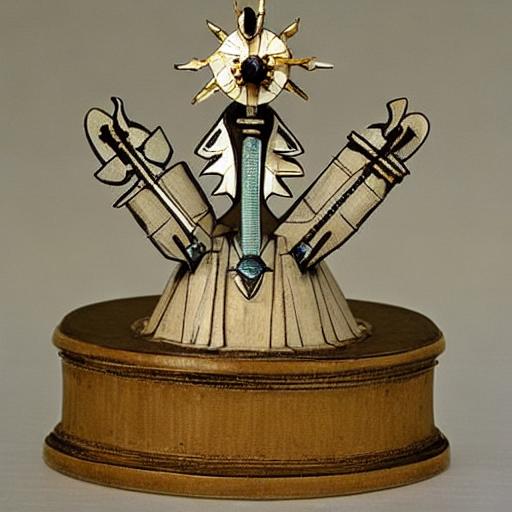}
     \includegraphics[width=\linewidth]{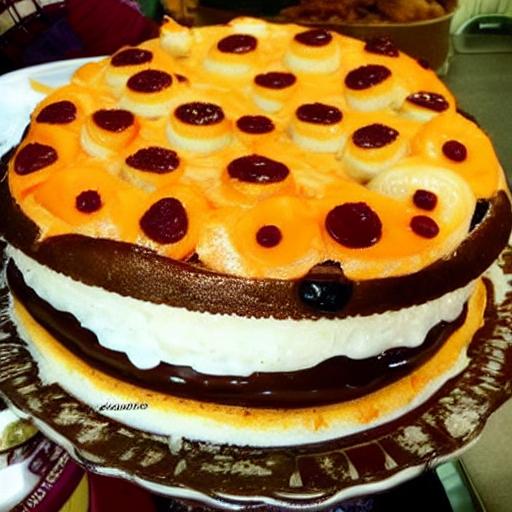}
     \includegraphics[width=\linewidth]{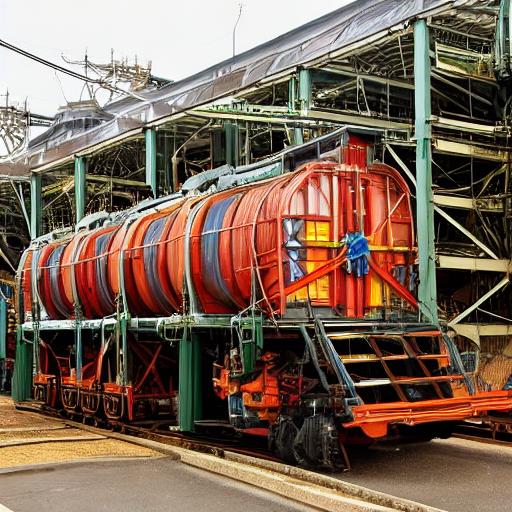}
     \includegraphics[width=\linewidth]{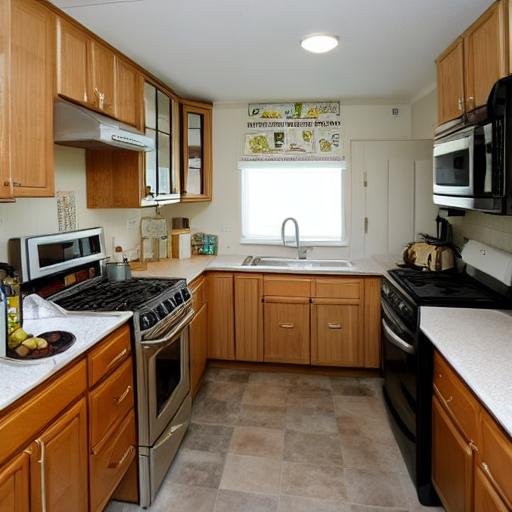}
     \includegraphics[width=\linewidth]{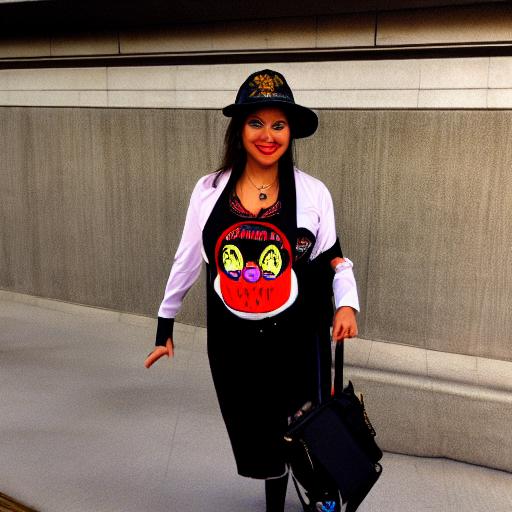}
     \includegraphics[width=\linewidth]{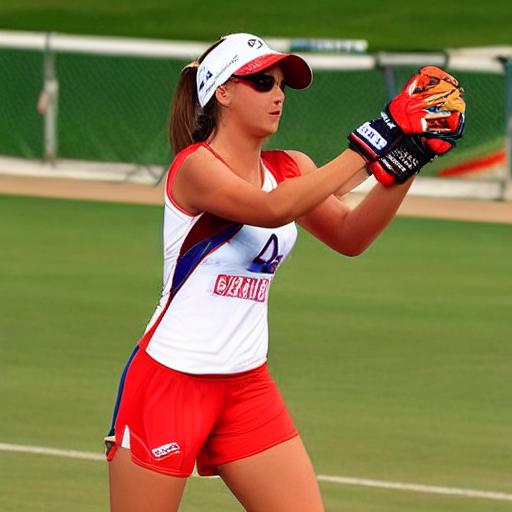}
     \includegraphics[width=\linewidth]{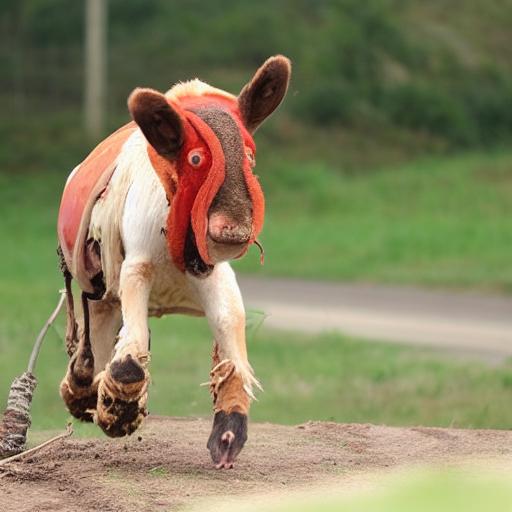}
     \includegraphics[width=\linewidth]{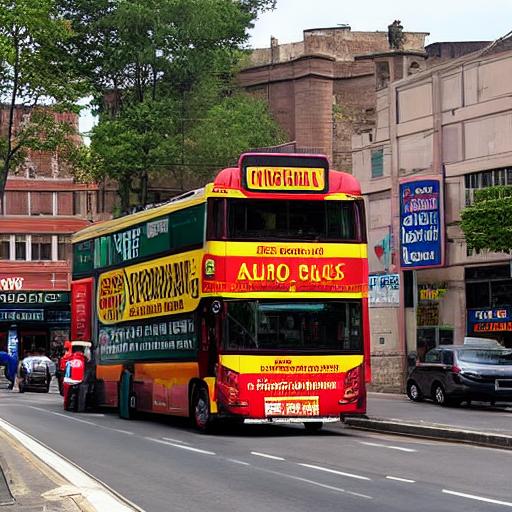}
     \includegraphics[width=\linewidth]{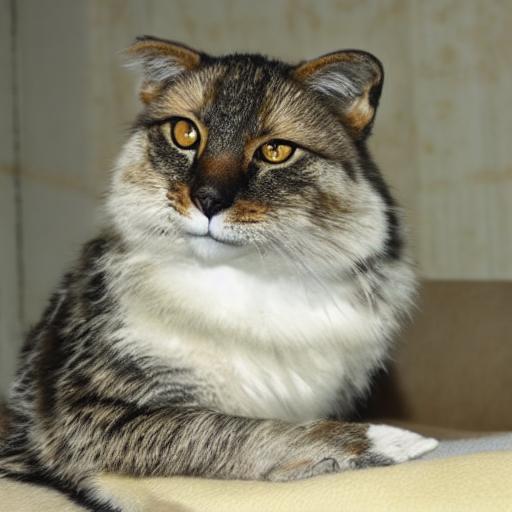}
     \end{minipage}
     }
     \hspace{-2.25mm}
     \subfloat[MindEye2~\cite{scotti2024mindeye2}]{
     \begin{minipage}{0.12\linewidth}
     \includegraphics[width=\linewidth]{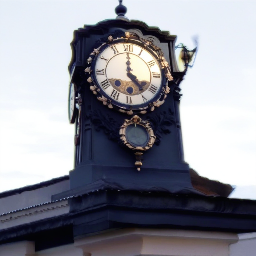}
     \includegraphics[width=\linewidth]{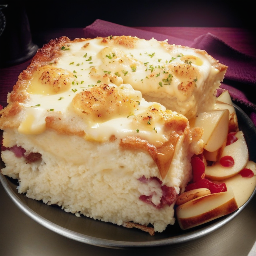}
     \includegraphics[width=\linewidth]{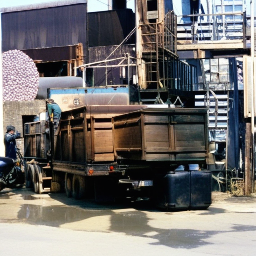}
     \includegraphics[width=\linewidth]{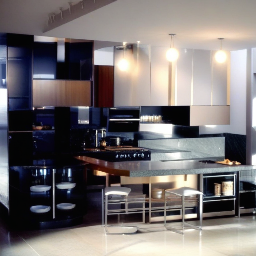}
     \includegraphics[width=\linewidth]{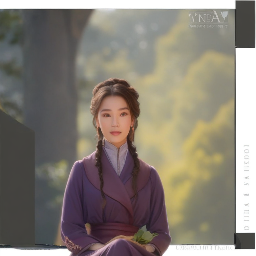}
     \includegraphics[width=\linewidth]{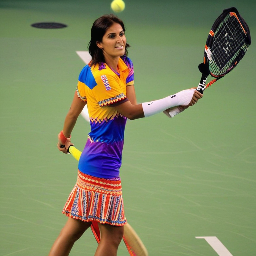}
     \includegraphics[width=\linewidth]{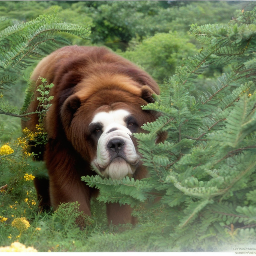}
     \includegraphics[width=\linewidth]{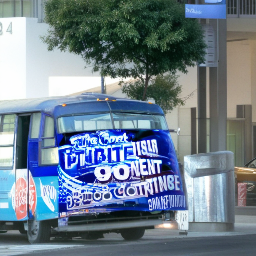}
     \includegraphics[width=\linewidth]{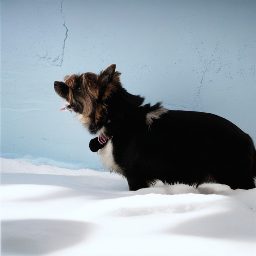}
     \end{minipage}
     }
     \hspace{-2.25mm}
     \subfloat[Neuropictor~\cite{huo2024neuropictor}]{
     \begin{minipage}{0.12\linewidth}
     \includegraphics[width=\linewidth]{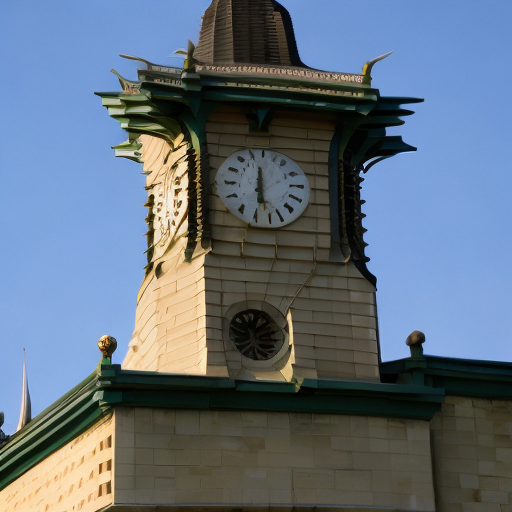}
     \includegraphics[width=\linewidth]{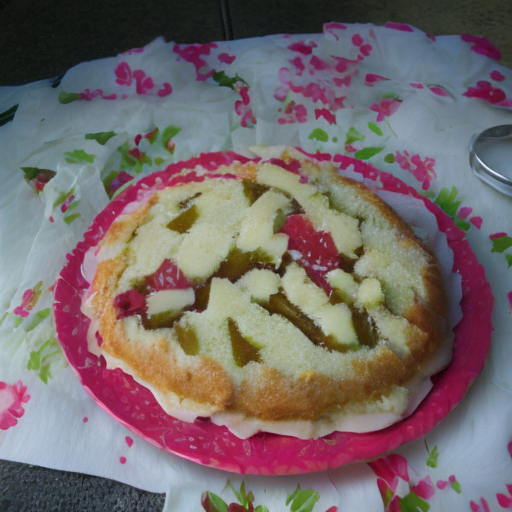}
     \includegraphics[width=\linewidth]{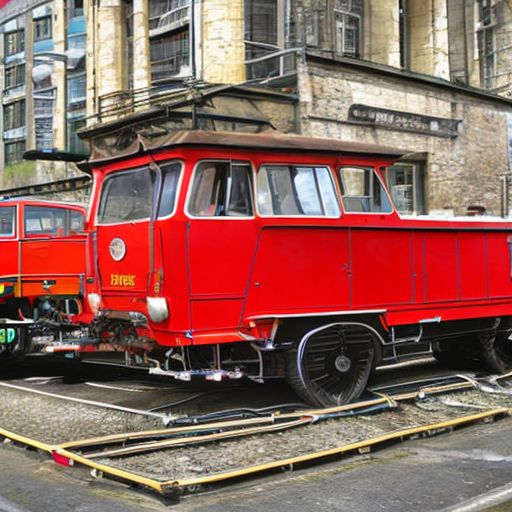}
     \includegraphics[width=\linewidth]{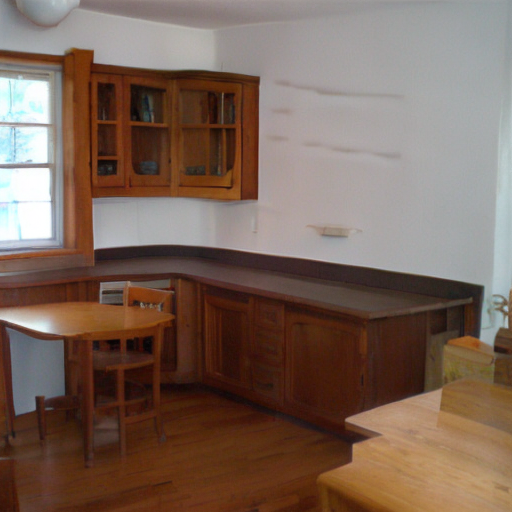}
     \includegraphics[width=\linewidth]{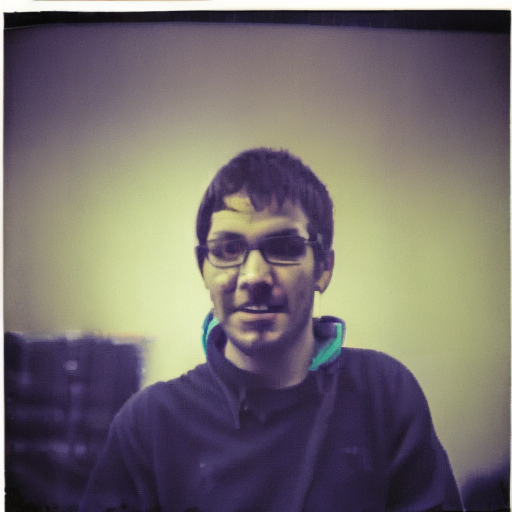}
     \includegraphics[width=\linewidth]{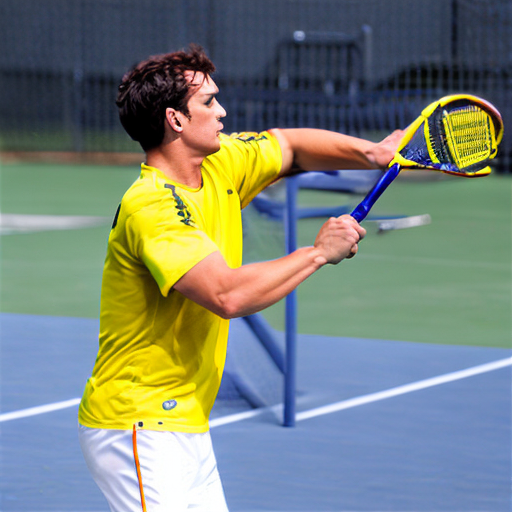}
     \includegraphics[width=\linewidth]{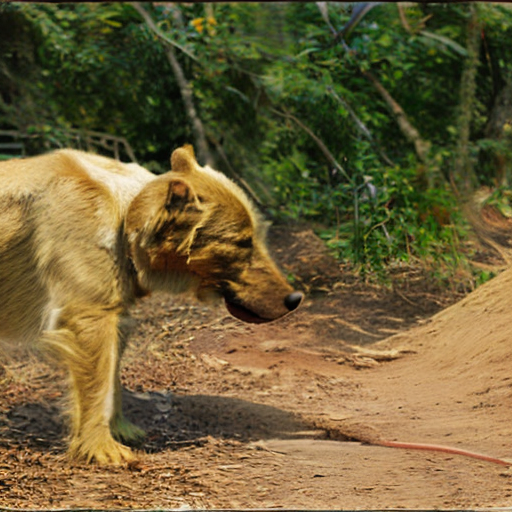}
     \includegraphics[width=\linewidth]{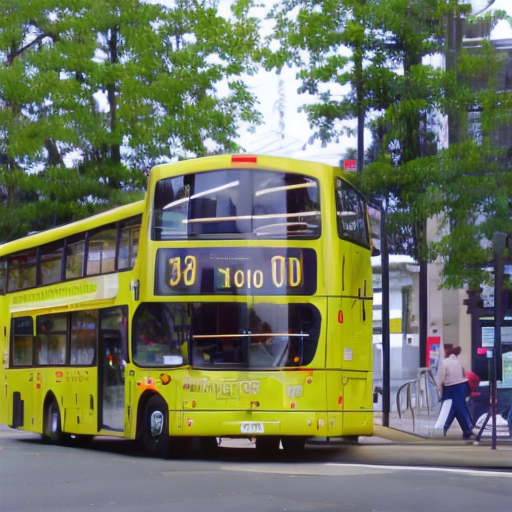}
     \includegraphics[width=\linewidth]{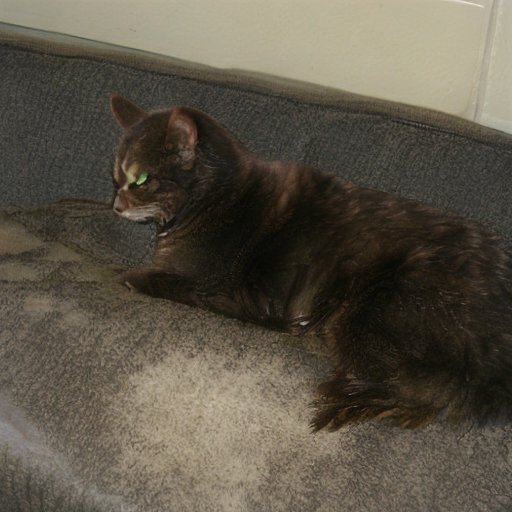}
     \end{minipage}
     }
     \hspace{-2.25mm}
     \subfloat[Ours]{
     \begin{minipage}{0.12\linewidth}
     \includegraphics[width=\linewidth]{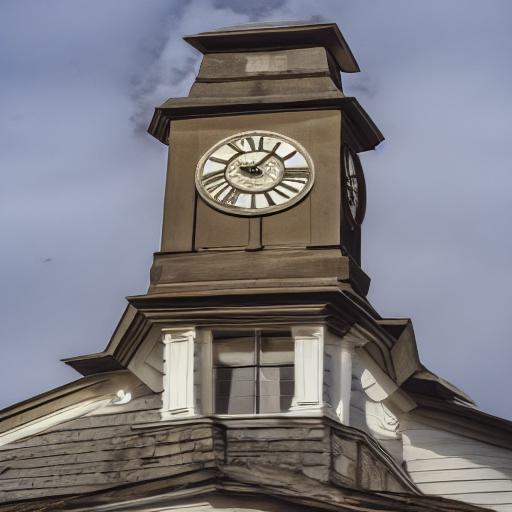}
     \includegraphics[width=\linewidth]{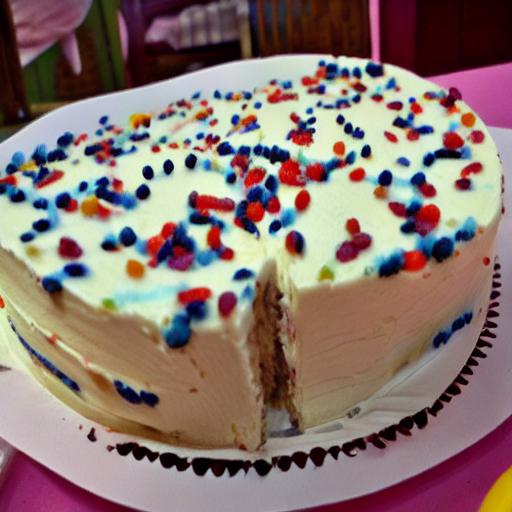}
     \includegraphics[width=\linewidth]{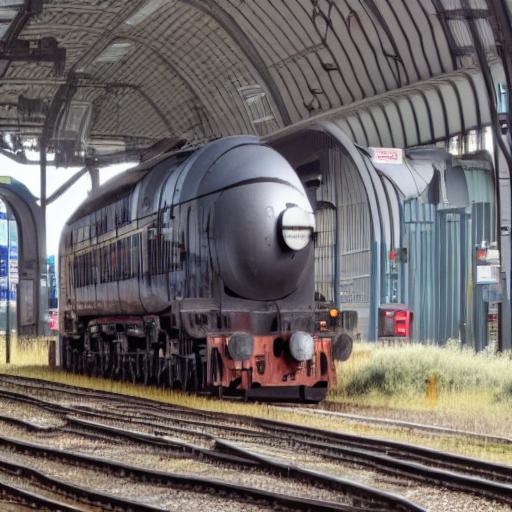}
     \includegraphics[width=\linewidth]{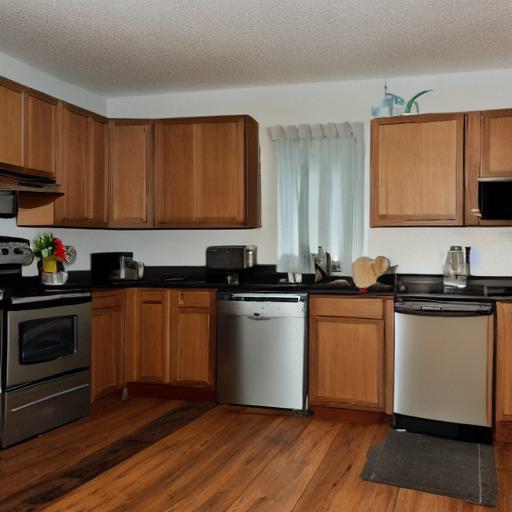}
     \includegraphics[width=\linewidth]{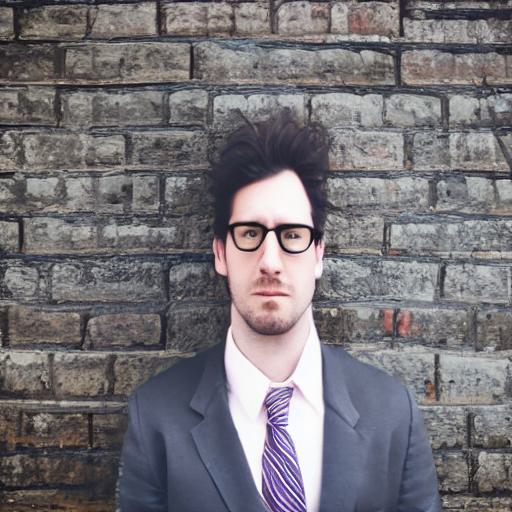}
     \includegraphics[width=\linewidth]{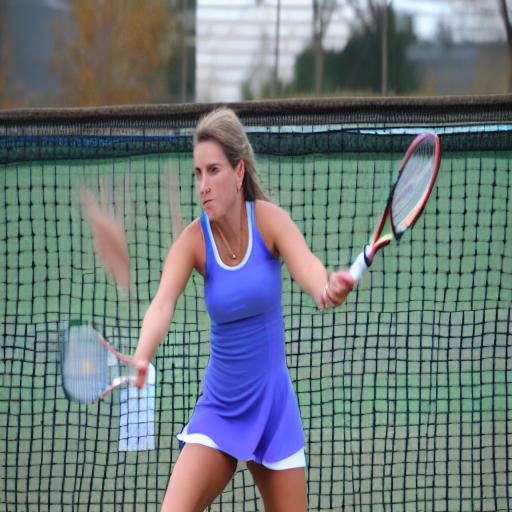}
     \includegraphics[width=\linewidth]{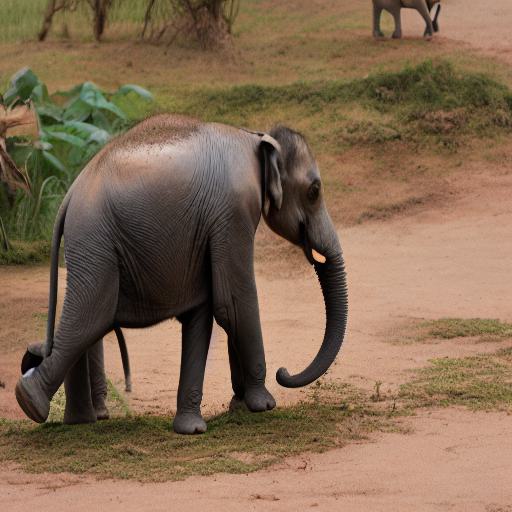}
     \includegraphics[width=\linewidth]{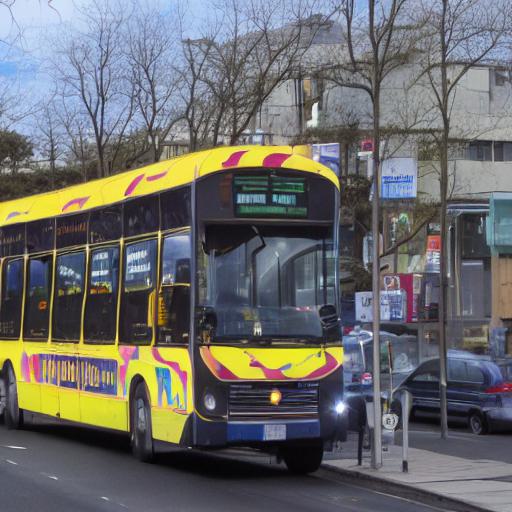}
     \includegraphics[width=\linewidth]{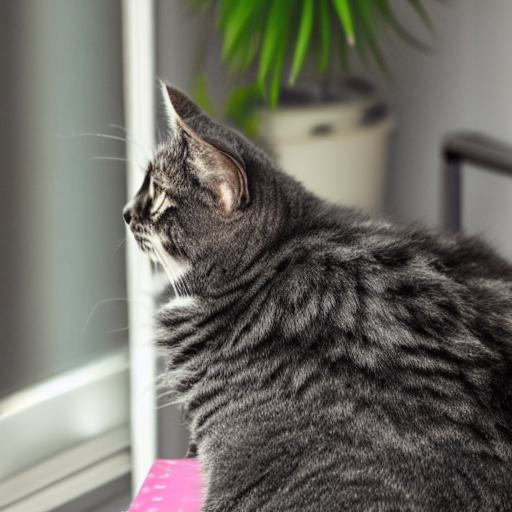}
     \end{minipage}
     }
     \hspace{-2.25mm}
     \subfloat[Stimulus]{
     \begin{minipage}{0.12\linewidth}
     \includegraphics[width=\linewidth]{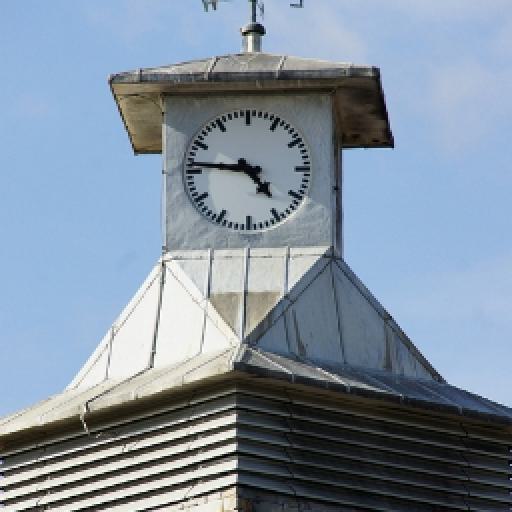}
     \includegraphics[width=\linewidth]{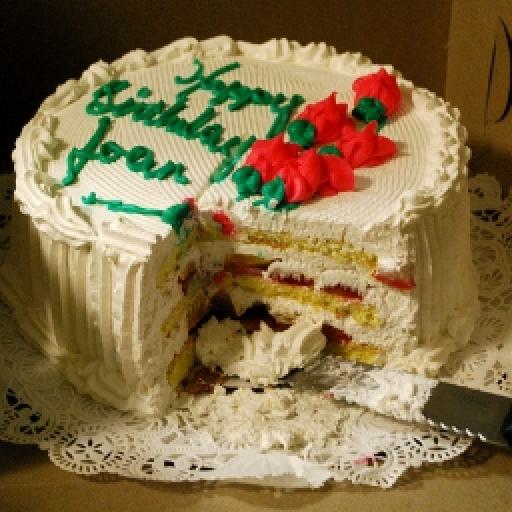}
     \includegraphics[width=\linewidth]{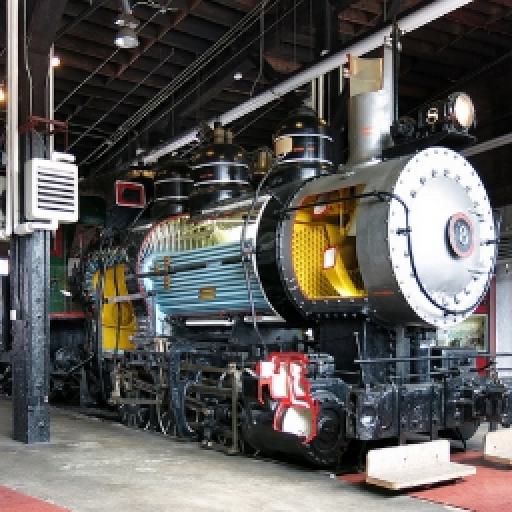}
     \includegraphics[width=\linewidth]{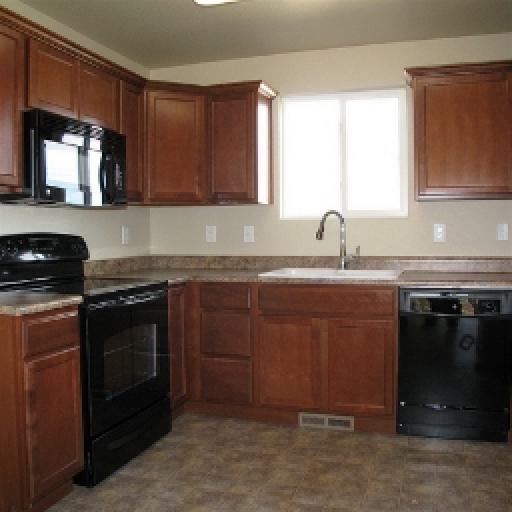}
     \includegraphics[width=\linewidth]{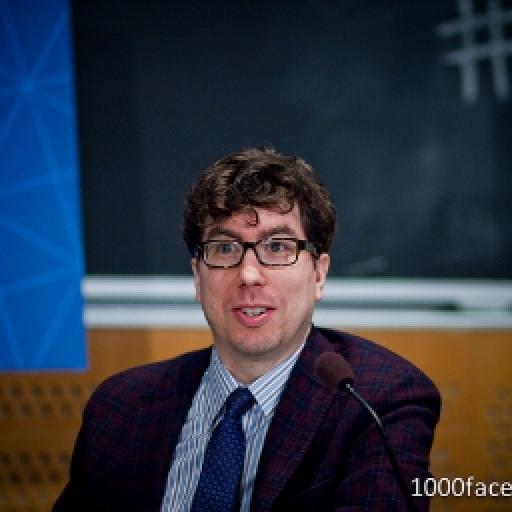}
     \includegraphics[width=\linewidth]{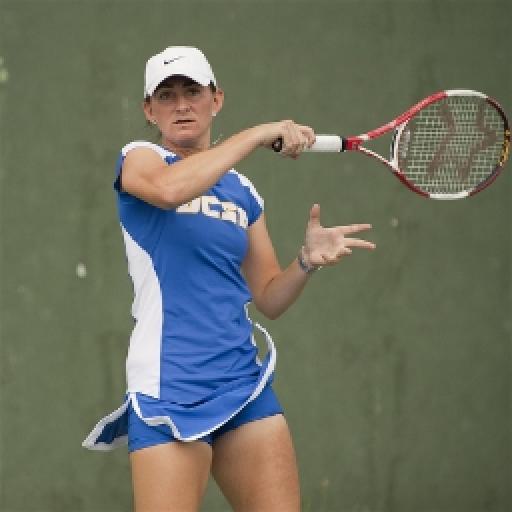}
     \includegraphics[width=\linewidth]{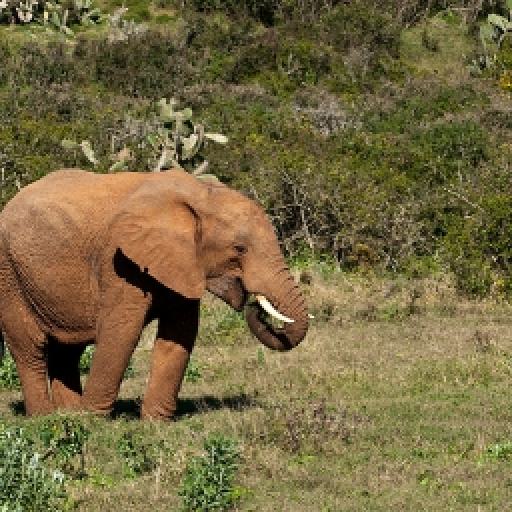}
     \includegraphics[width=\linewidth]{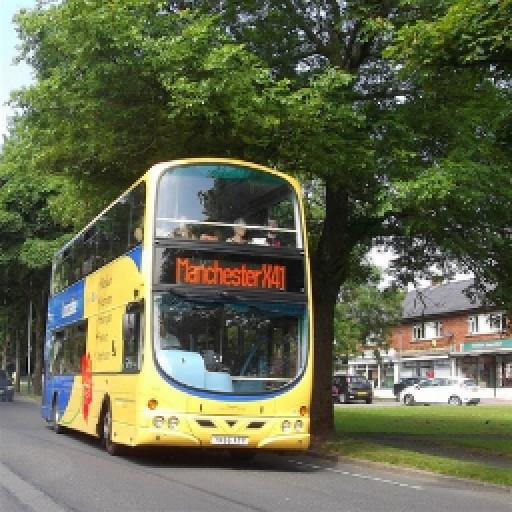}
     \includegraphics[width=\linewidth]{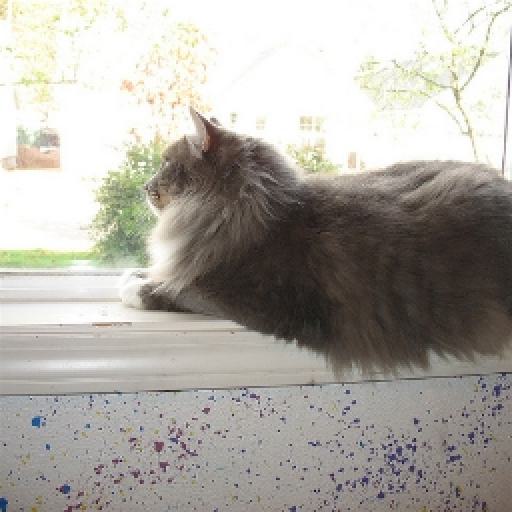}
     \end{minipage}
     }
\vspace{-2mm}
\caption{More qualitative comparison with competitors on mind decoding.}
\vspace{-5mm}
\label{fig:figure_supp}
\end{figure*}

Additional qualitative comparisons with competing methods are presented in Fig.~\ref{fig:figure_supp}. Our method demonstrates higher consistency with the GT stimulus images in terms of semantics, structure, and color.

\textbf{Qualitative Comparison of Variants}

\begin{figure*}[!t]
    \centering
    \captionsetup[subfloat]{labelformat=empty,justification=centering}
     \subfloat[UM]{
     \begin{minipage}{0.1\linewidth}
     \includegraphics[width=\linewidth]{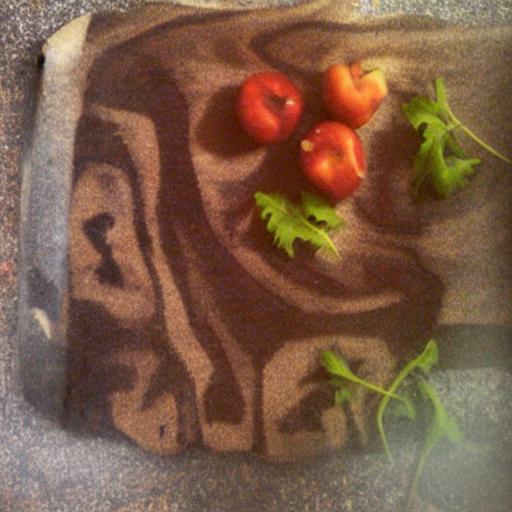}
     \includegraphics[width=\linewidth]{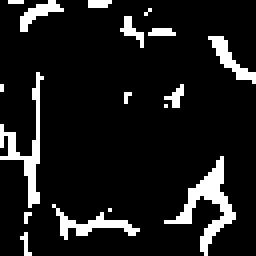}
     \includegraphics[width=\linewidth]{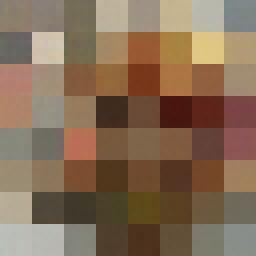}\vspace{2mm}
     \includegraphics[width=\linewidth]{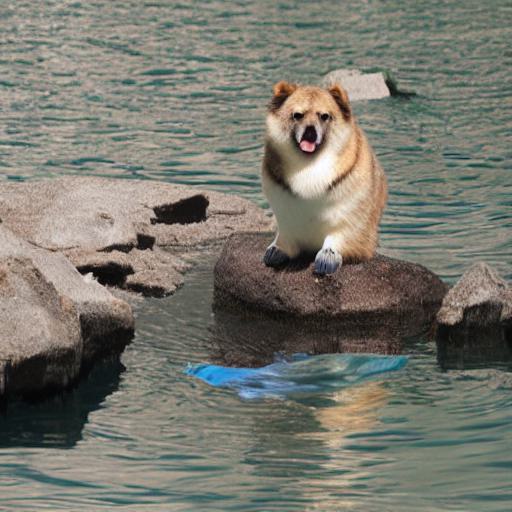}
     \includegraphics[width=\linewidth]{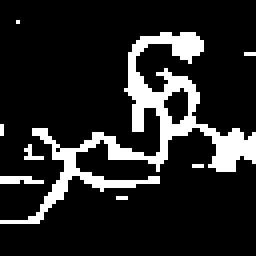}
     \includegraphics[width=\linewidth]{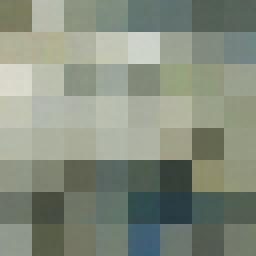}\vspace{2mm}
     \includegraphics[width=\linewidth]{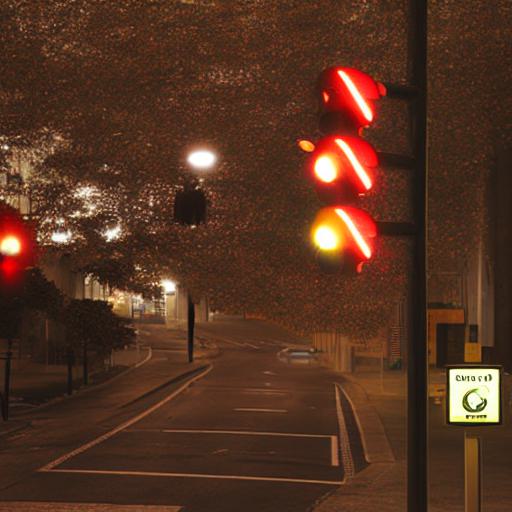}
     \includegraphics[width=\linewidth]{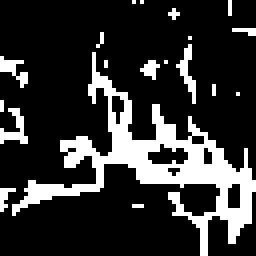}
     \includegraphics[width=\linewidth]{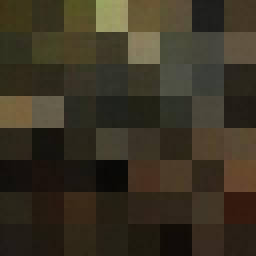}\vspace{2mm}
     \includegraphics[width=\linewidth]{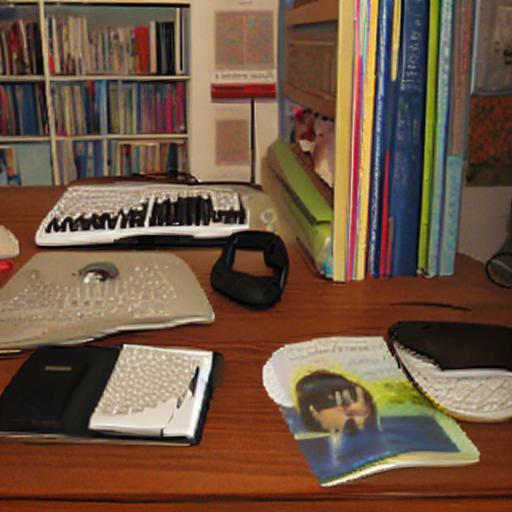}
     \includegraphics[width=\linewidth]{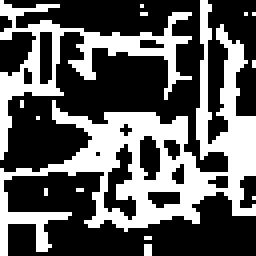}
     \includegraphics[width=\linewidth]{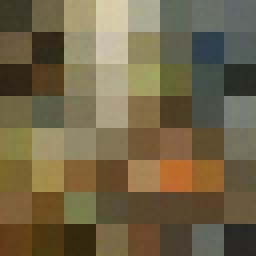}
     \end{minipage}
     }
     \hspace{-2.25mm}
     \subfloat[$w/o$ SBMM]{
     \begin{minipage}{0.1\linewidth}
     \includegraphics[width=\linewidth]{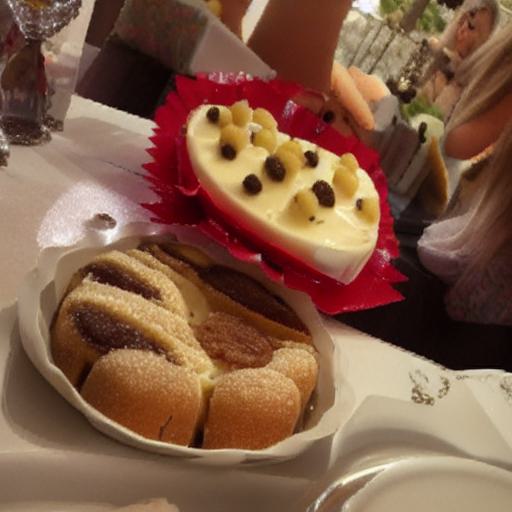}
     \includegraphics[width=\linewidth]{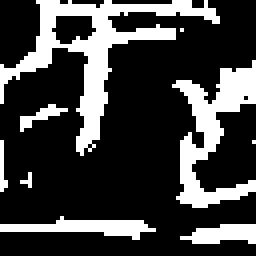}
     \includegraphics[width=\linewidth]{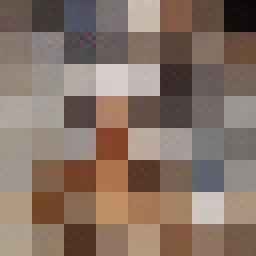}\vspace{2mm}
     \includegraphics[width=\linewidth]{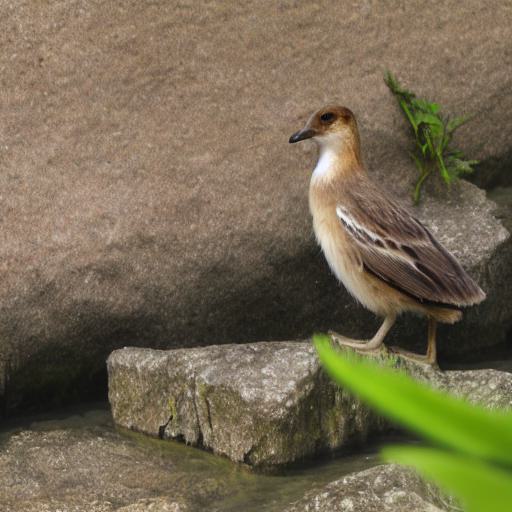}
     \includegraphics[width=\linewidth]{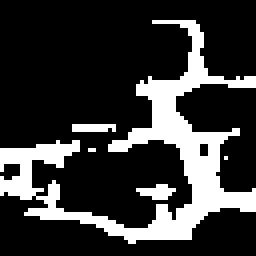}
     \includegraphics[width=\linewidth]{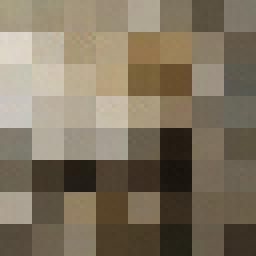}\vspace{2mm}
     \includegraphics[width=\linewidth]{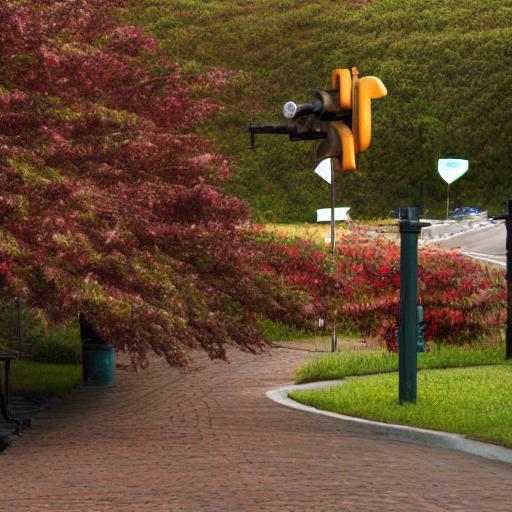}
     \includegraphics[width=\linewidth]{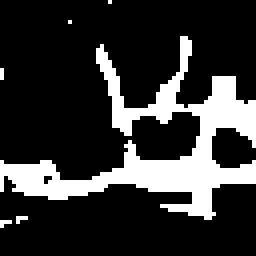}
     \includegraphics[width=\linewidth]{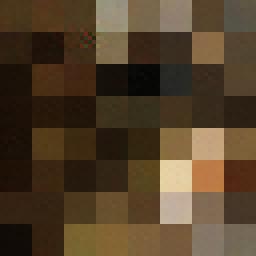}\vspace{2mm}
     \includegraphics[width=\linewidth]{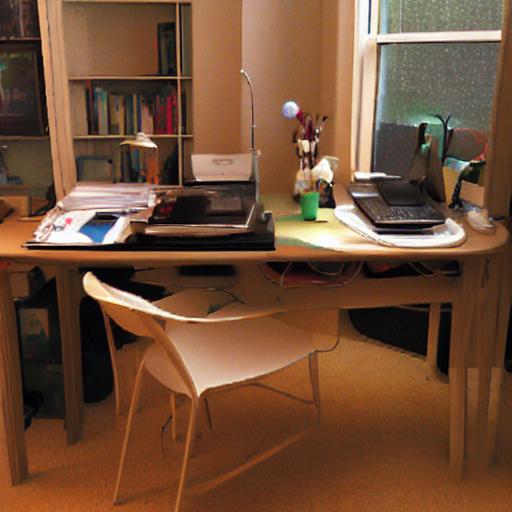}
     \includegraphics[width=\linewidth]{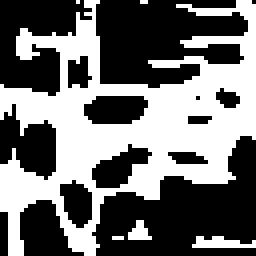}
     \includegraphics[width=\linewidth]{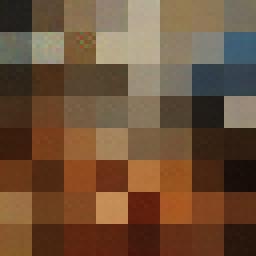}
     \end{minipage}
     }
     \hspace{-2.25mm}
     \subfloat[Our-SS]{
     \begin{minipage}{0.1\linewidth}
     \includegraphics[width=\linewidth]{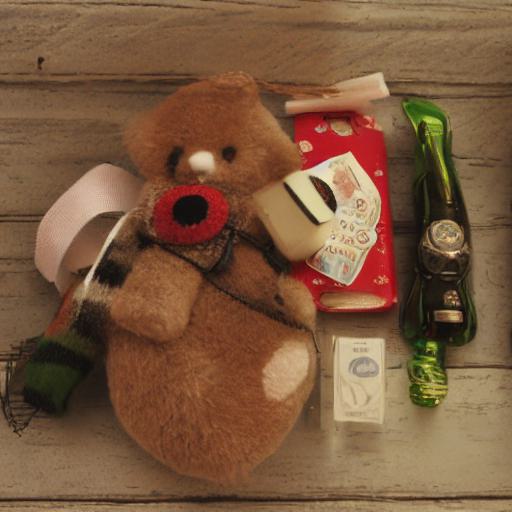}
     \includegraphics[width=\linewidth]{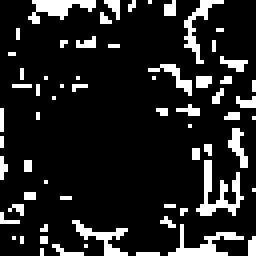}
     \includegraphics[width=\linewidth]{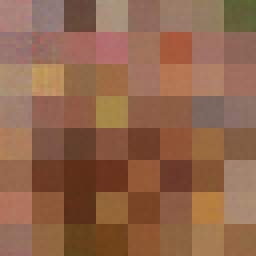}\vspace{2mm}
     \includegraphics[width=\linewidth]{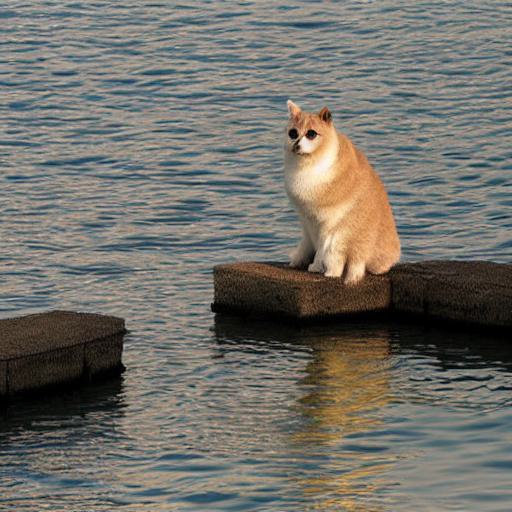}
     \includegraphics[width=\linewidth]{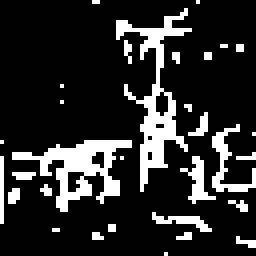}
     \includegraphics[width=\linewidth]{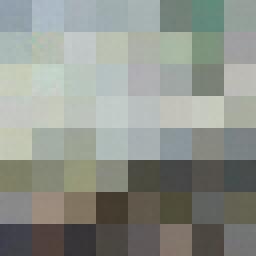}\vspace{2mm}
     \includegraphics[width=\linewidth]{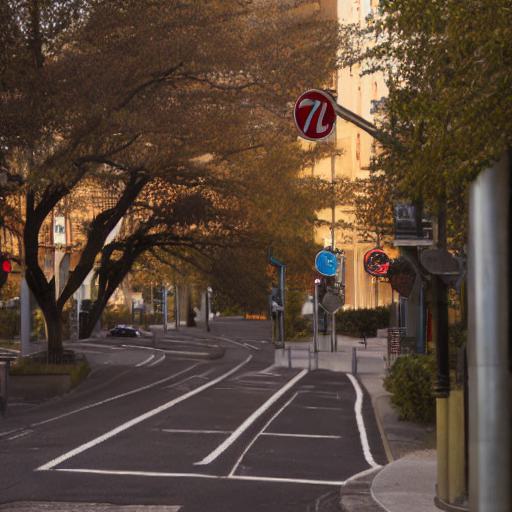}
     \includegraphics[width=\linewidth]{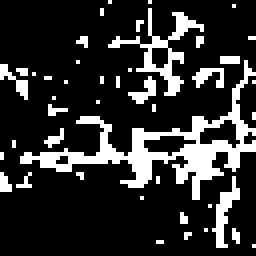}
     \includegraphics[width=\linewidth]{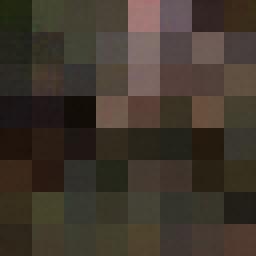}\vspace{2mm}
     \includegraphics[width=\linewidth]{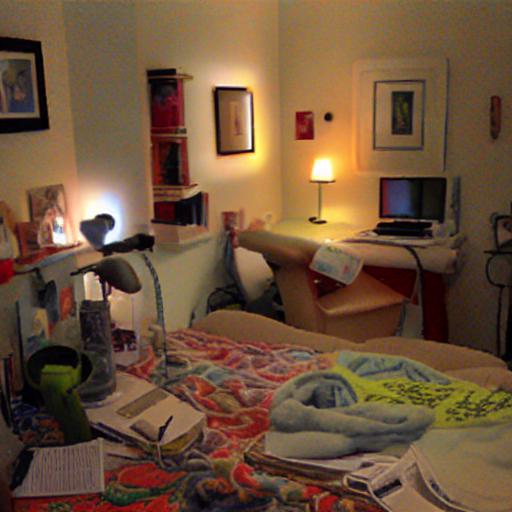}
     \includegraphics[width=\linewidth]{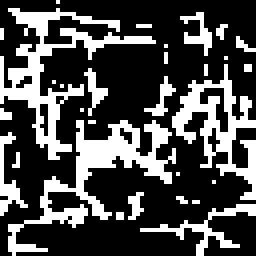}
     \includegraphics[width=\linewidth]{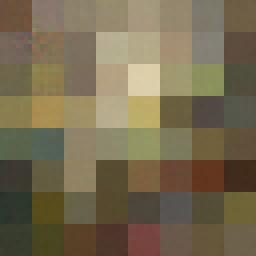}
     \end{minipage}
     }
     % \hspace{-2.25mm}
     \subfloat[\textit{Direct \\ Addition}]{
     \begin{minipage}{0.1\linewidth}
     \includegraphics[width=\linewidth]{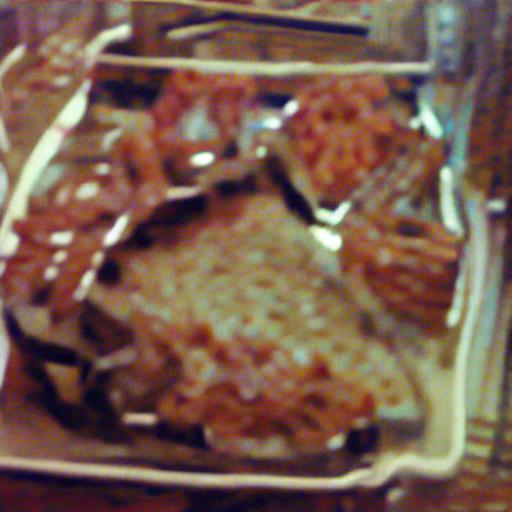}
     \includegraphics[width=\linewidth]{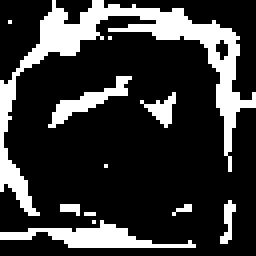}
     \includegraphics[width=\linewidth]{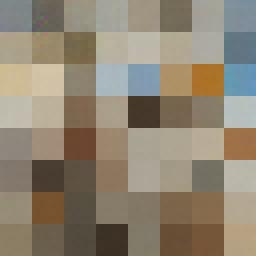}\vspace{2mm}
     \includegraphics[width=\linewidth]{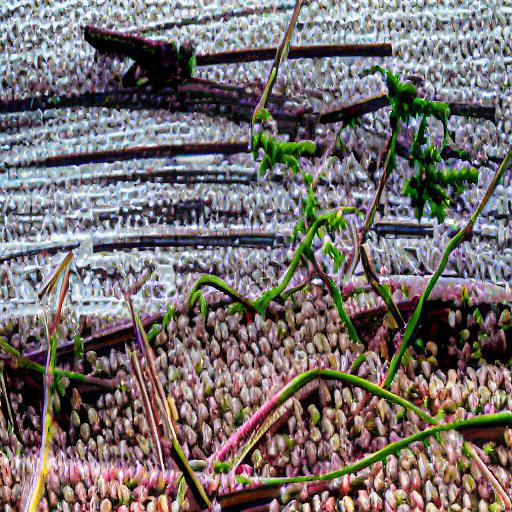}
     \includegraphics[width=\linewidth]{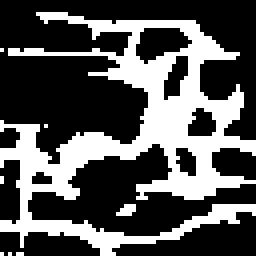}
     \includegraphics[width=\linewidth]{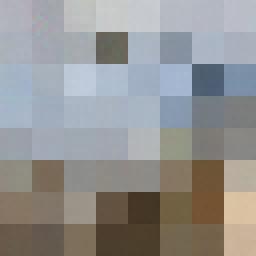}\vspace{2mm}
     \includegraphics[width=\linewidth]{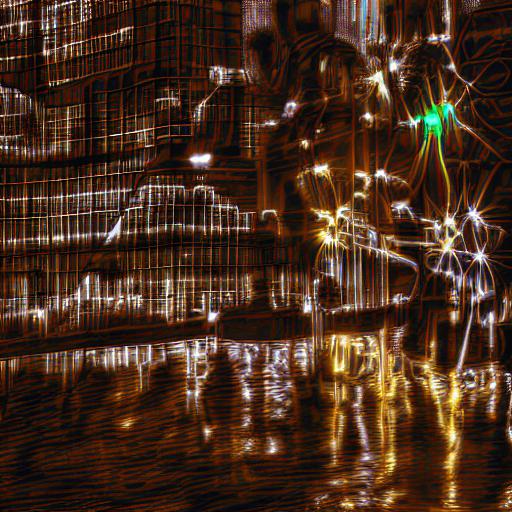}
     \includegraphics[width=\linewidth]{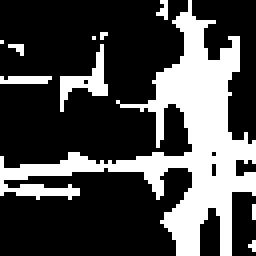}
     \includegraphics[width=\linewidth]{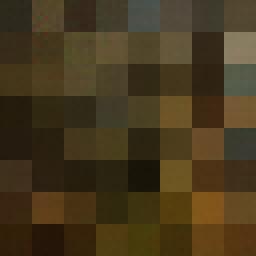}\vspace{2mm}
     \includegraphics[width=\linewidth]{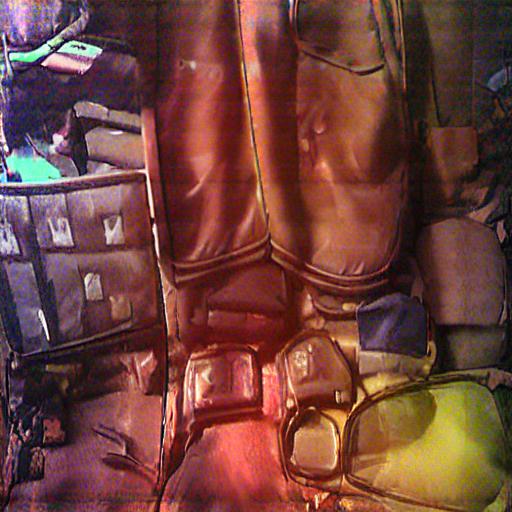}
     \includegraphics[width=\linewidth]{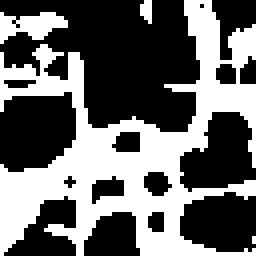}
     \includegraphics[width=\linewidth]{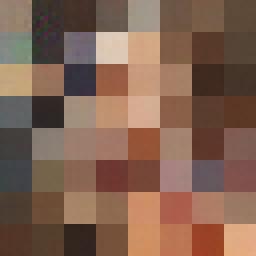}
     \end{minipage}
     }
     \hspace{-2.25mm}
     \subfloat[\textit{Direct \\ Addition} \\+ SRM]{
     \begin{minipage}{0.1\linewidth}
     \includegraphics[width=\linewidth]{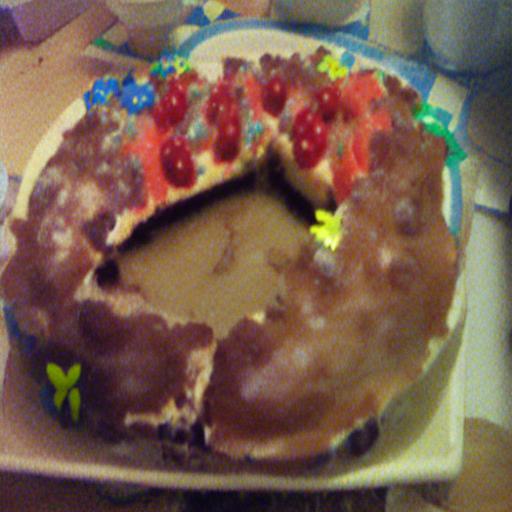}
     \includegraphics[width=\linewidth]{figures_ablation_supp_prd_src_main_1000_sample000001097_edge_1.jpg}
     \includegraphics[width=\linewidth]{figures_ablation_supp_prd_src_main_1000_sample000001097_color_1.jpg}\vspace{2mm}
     \includegraphics[width=\linewidth]{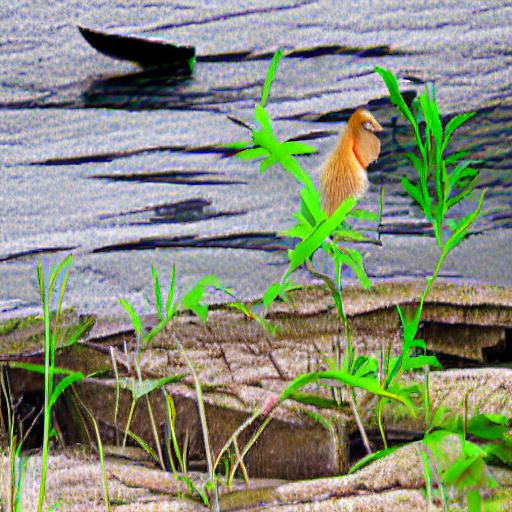}
     \includegraphics[width=\linewidth]{figures_ablation_supp_prd_src_main_1000_sample000003924_edge_1.jpg}
     \includegraphics[width=\linewidth]{figures_ablation_supp_prd_src_main_1000_sample000003924_color_1.jpg}\vspace{2mm}
     \includegraphics[width=\linewidth]{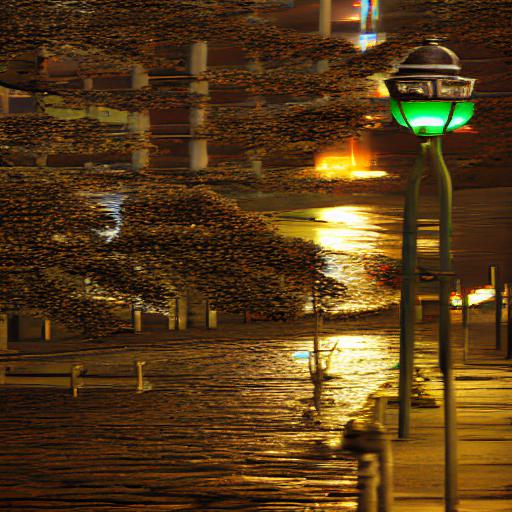}
     \includegraphics[width=\linewidth]{figures_ablation_supp_prd_src_main_1000_sample000003807_edge_1.jpg}
     \includegraphics[width=\linewidth]{figures_ablation_supp_prd_src_main_1000_sample000003807_color_1.jpg}\vspace{2mm}
     \includegraphics[width=\linewidth]{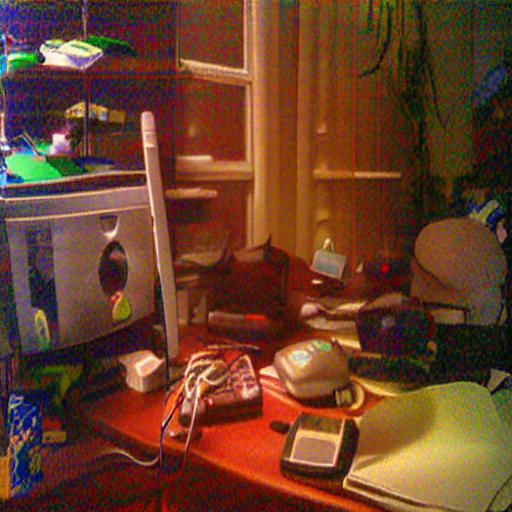}
     \includegraphics[width=\linewidth]{figures_ablation_supp_prd_src_main_1000_sample000006617_edge_1.jpg}
     \includegraphics[width=\linewidth]{figures_ablation_supp_prd_src_main_1000_sample000006617_color_1.jpg}
     \end{minipage}
     }
     \hspace{-2.25mm}
     \subfloat[\textit{Direct \\ Addition} \\ + VCM]{
     \begin{minipage}{0.1\linewidth}
     \includegraphics[width=\linewidth]{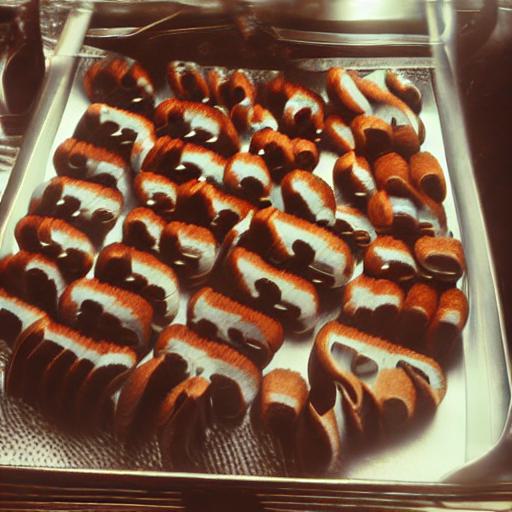}
     \includegraphics[width=\linewidth]{figures_ablation_supp_prd_src_main_1000_sample000001097_edge_1.jpg}
     \includegraphics[width=\linewidth]{figures_ablation_supp_prd_src_main_1000_sample000001097_color_1.jpg}\vspace{2mm}
     \includegraphics[width=\linewidth]{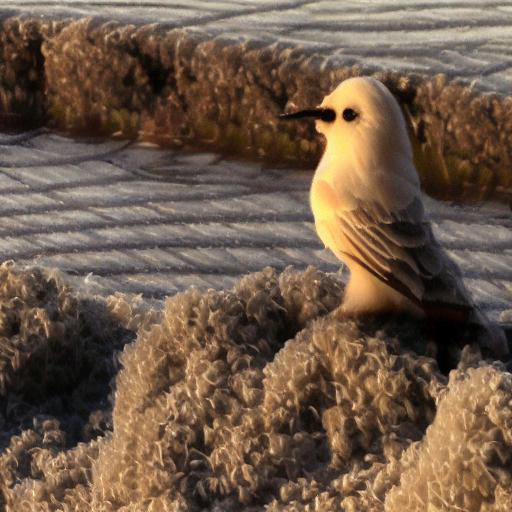}
     \includegraphics[width=\linewidth]{figures_ablation_supp_prd_src_main_1000_sample000003924_edge_1.jpg}
     \includegraphics[width=\linewidth]{figures_ablation_supp_prd_src_main_1000_sample000003924_color_1.jpg}\vspace{2mm}
     \includegraphics[width=\linewidth]{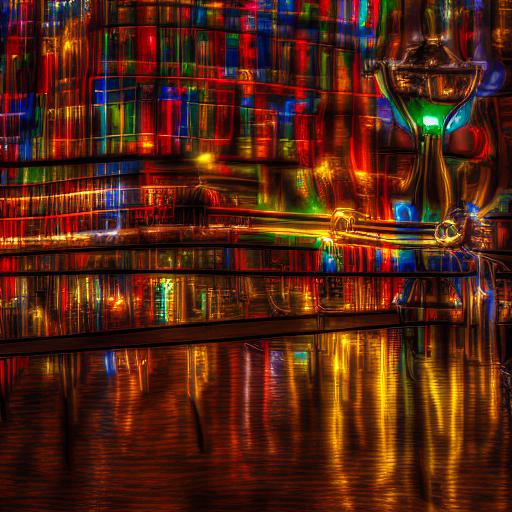}
     \includegraphics[width=\linewidth]{figures_ablation_supp_prd_src_main_1000_sample000003807_edge_1.jpg}
     \includegraphics[width=\linewidth]{figures_ablation_supp_prd_src_main_1000_sample000003807_color_1.jpg}\vspace{2mm}
     \includegraphics[width=\linewidth]{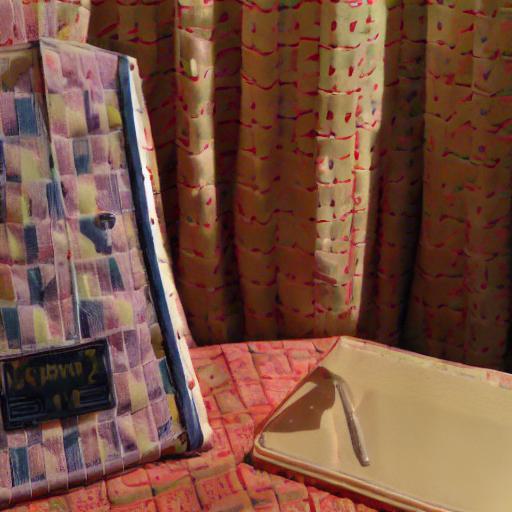}
     \includegraphics[width=\linewidth]{figures_ablation_supp_prd_src_main_1000_sample000006617_edge_1.jpg}
     \includegraphics[width=\linewidth]{figures_ablation_supp_prd_src_main_1000_sample000006617_color_1.jpg}
     \end{minipage}
     }
     % \hspace{-2.25mm}
     \subfloat[Our]{
     \begin{minipage}{0.1\linewidth}
     \includegraphics[width=\linewidth]{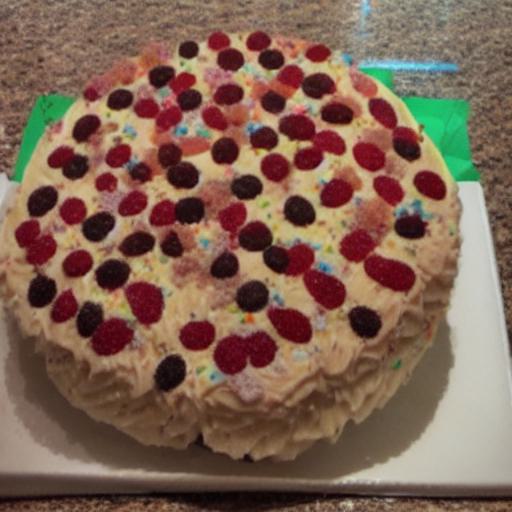}
     \includegraphics[width=\linewidth]{figures_ablation_supp_prd_src_main_1000_sample000001097_edge_1.jpg}
     \includegraphics[width=\linewidth]{figures_ablation_supp_prd_src_main_1000_sample000001097_color_1.jpg}\vspace{2mm}
     \includegraphics[width=\linewidth]{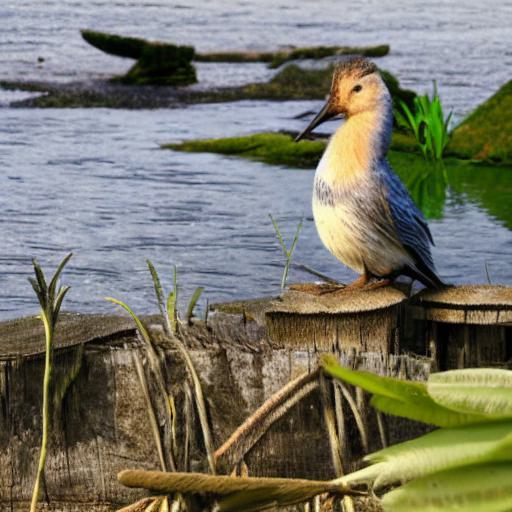}
     \includegraphics[width=\linewidth]{figures_ablation_supp_prd_src_main_1000_sample000003924_edge_1.jpg}
     \includegraphics[width=\linewidth]{figures_ablation_supp_prd_src_main_1000_sample000003924_color_1.jpg}\vspace{2mm}
     \includegraphics[width=\linewidth]{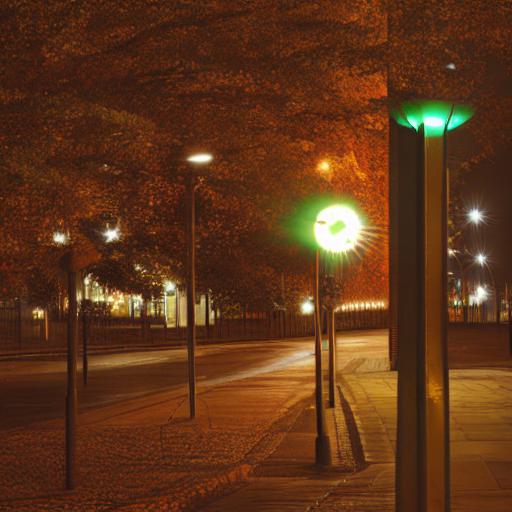}
     \includegraphics[width=\linewidth]{figures_ablation_supp_prd_src_main_1000_sample000003807_edge_1.jpg}
     \includegraphics[width=\linewidth]{figures_ablation_supp_prd_src_main_1000_sample000003807_color_1.jpg}\vspace{2mm}
     \includegraphics[width=\linewidth]{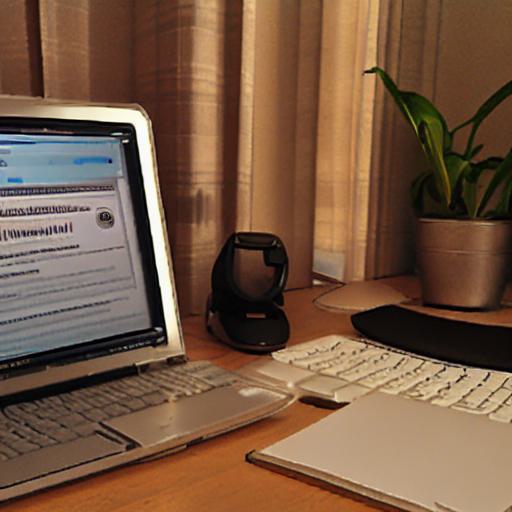}
     \includegraphics[width=\linewidth]{figures_ablation_supp_prd_src_main_1000_sample000006617_edge_1.jpg}
     \includegraphics[width=\linewidth]{figures_ablation_supp_prd_src_main_1000_sample000006617_color_1.jpg}
     \end{minipage}
     }
     \hspace{-2.25mm}
     \subfloat[Stimulus]{
     \begin{minipage}{0.1\linewidth}
     \includegraphics[width=\linewidth]{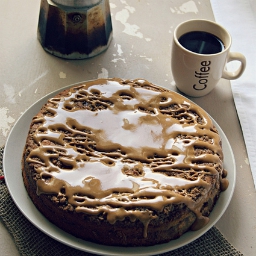}
     \includegraphics[width=\linewidth]{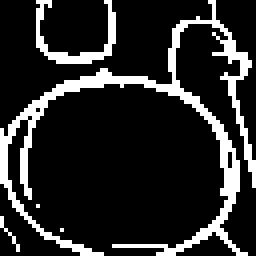}
     \includegraphics[width=\linewidth]{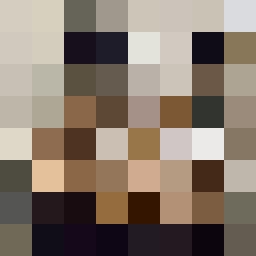}\vspace{2mm}
     \includegraphics[width=\linewidth]{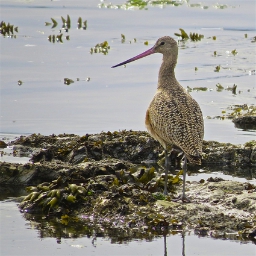}
     \includegraphics[width=\linewidth]{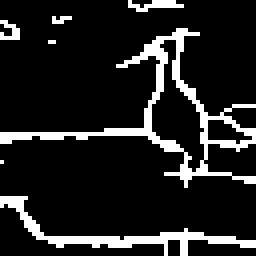}
     \includegraphics[width=\linewidth]{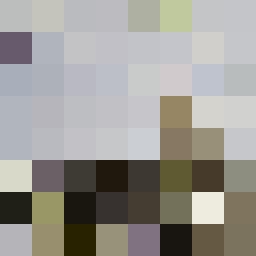}\vspace{2mm}
     \includegraphics[width=\linewidth]{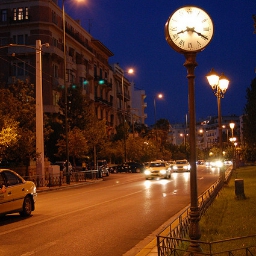}
     \includegraphics[width=\linewidth]{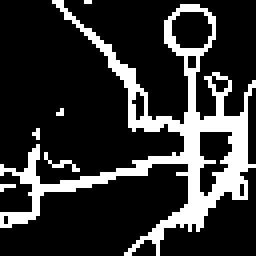}
     \includegraphics[width=\linewidth]{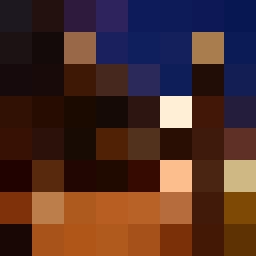}\vspace{2mm}
     \includegraphics[width=\linewidth]{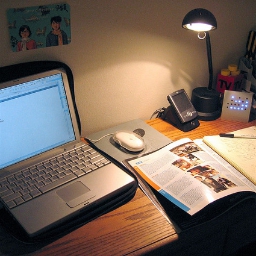}
     \includegraphics[width=\linewidth]{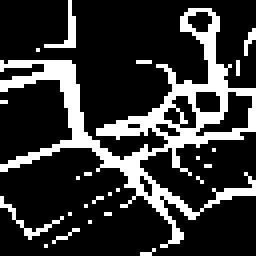}
     \includegraphics[width=\linewidth]{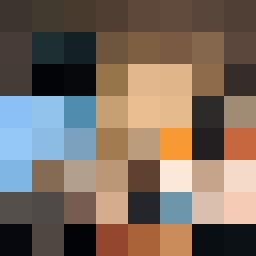}
     \end{minipage}
     }
    \rotatebox[origin=c]{270}{\hspace{-20cm}\small{\texttt{Image}} \hspace{5mm} \small{\texttt{Edge}} \hspace{5mm} \small{\texttt{Color}}  \hspace{8mm} \small{\texttt{Image}} \hspace{5mm} \small{\texttt{Edge}} \hspace{5mm} \small{\texttt{Color}}  \hspace{8mm} \small{\texttt{Image}} \hspace{5mm} \small{\texttt{Edge}} \hspace{5mm} \small{\texttt{Color}}  \hspace{8mm} \small{\texttt{Image}} \hspace{5mm} \small{\texttt{Edge}} \hspace{5mm} \small{\texttt{Color}}  }\\

\vspace{-2mm}
\caption{Qualitative comparison with various variants.}
\vspace{-5mm}
\label{fig:figure_ablation_supp}
\end{figure*}

We present a qualitative comparison of various model variants, including predicted edges and color representations, in Fig.~\ref{fig:figure_ablation_supp}. The variants UM, $w/o$ SBMM, and Ours-SS fail to predict the reasonable edges and color representations, whereas our model successfully predicts the rough edges and color of the image. Directly using the predicted representations does not yield plausible images, as the second-stage mind decoding requires accurate representations. Our proposed modules, SRM and VCM, enhance the quality of the reconstructed image by tolerating rough representations. Finally, our complete framework produces faithfully reconstructed results with plausible appearances.

\textbf{Analysis of VCM}

\begin{figure*}[!h]
    \vspace{-4mm}
    \centering
    \captionsetup[subfloat]{labelformat=empty,justification=centering}
    \subfloat[Prd. Edge]{
    \begin{minipage}{0.12\linewidth} % 每个子图的宽度
        \centering
        \includegraphics[width=\textwidth]{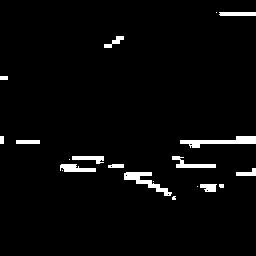}
        \includegraphics[width=\textwidth]{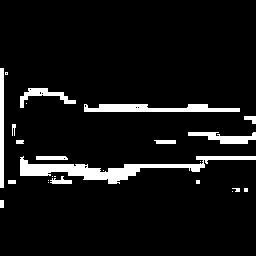}
    \end{minipage}}
    \subfloat[Prd. Color]{
    \begin{minipage}{0.12\linewidth}
        \centering
        \includegraphics[width=\textwidth]{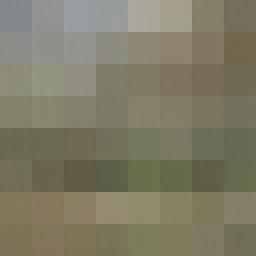}
        \includegraphics[width=\textwidth]{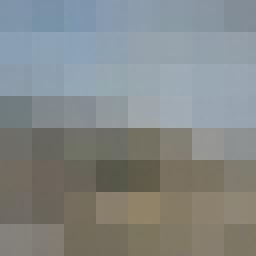}
    \end{minipage}}
    \subfloat[$\alpha_{e}$]{
    \begin{minipage}{0.12\linewidth}
        \centering
        \includegraphics[width=\textwidth]{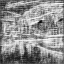}
        \includegraphics[width=\textwidth]{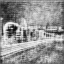}
    \end{minipage}}
    \subfloat[$\alpha_{c}$]{
    \begin{minipage}{0.12\linewidth}
        \centering
        \includegraphics[width=\textwidth]{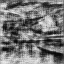}
        \includegraphics[width=\textwidth]{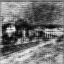}
    \end{minipage}}
    \subfloat[GT Edge]{
    \begin{minipage}{0.12\linewidth}
        \centering
        \includegraphics[width=\textwidth]{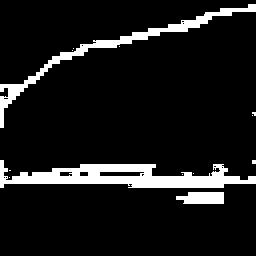}
        \includegraphics[width=\textwidth]{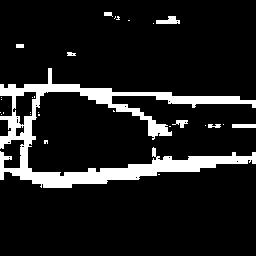}
    \end{minipage}}
    \subfloat[GT Color]{
    \begin{minipage}{0.12\linewidth}
        \centering
        \includegraphics[width=\textwidth]{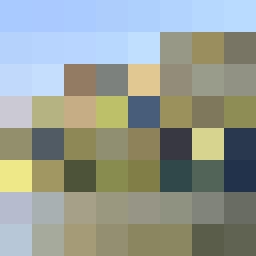}
        \includegraphics[width=\textwidth]{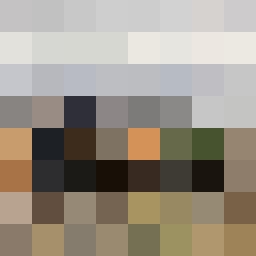}
    \end{minipage}}
    \subfloat[Prediction]{
    \begin{minipage}{0.12\linewidth}
        \centering
        \includegraphics[width=\textwidth]{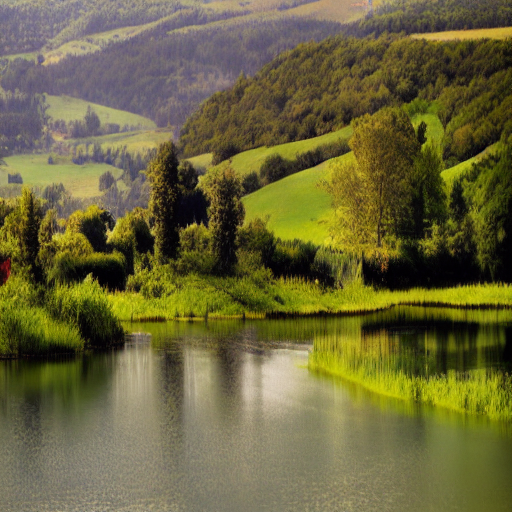}
        \includegraphics[width=\textwidth]{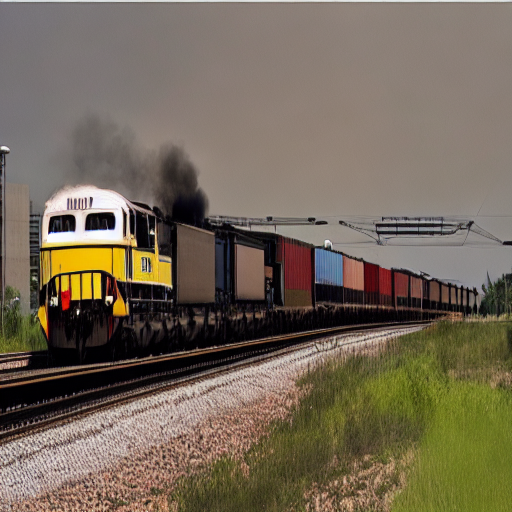}
    \end{minipage}}
    \subfloat[GT Stimulus]{
    \begin{minipage}{0.12\linewidth}
        \centering
        \includegraphics[width=\textwidth]{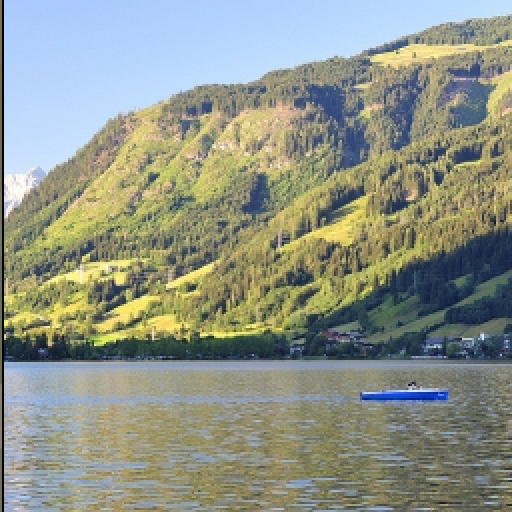}
        \includegraphics[width=\textwidth]{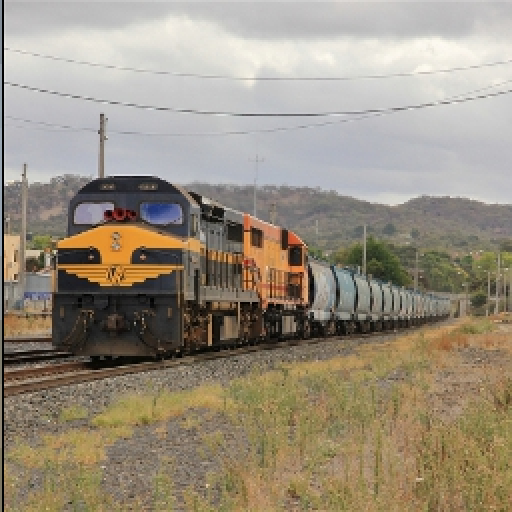}
    \end{minipage}}
    \caption{Visualization of VCM's output weights $\alpha_{e}$ and $\alpha_{c}$, they control the fusion weights to relax the influence of predicted edge and color conditions to output.}
    \label{vcm}
\end{figure*}

As shown in Fig.~\ref{fig:figure_ablation_supp}, the predicted edges and color palettes are dissimilar to the GT edges and colors, why the final reconstruction be faithful with the GT stimulus? Here we answer this question by visualizing the output of VCM. The visualization of VCM's output weights $\alpha_{e}$ and $\alpha_{c}$ are shown in Fig.~\ref{vcm} (the brighter indicates a higher value). The predicted $\alpha_{e}$ and $\alpha_{c}$ control fusion weights to relax the influence of predicted edge and color conditions to output, though predicted representations are inaccurate, the dissimilar representations can also lead to faithful mind decoding.

\textbf{Qualitative Comparison under Different Data Limitation Scenarios}

\begin{figure*}[t]
    \centering
    \captionsetup[subfloat]{labelformat=empty,justification=centering}
     \subfloat[Sketch \\ +\#500]{
     \begin{minipage}{0.1375\linewidth}
     \includegraphics[width=\linewidth]{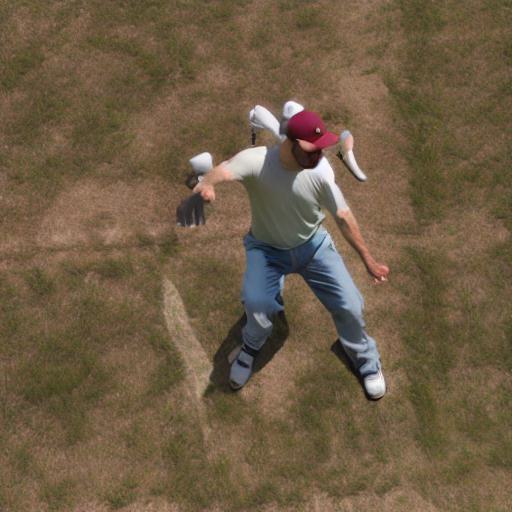}
     \includegraphics[width=\linewidth]{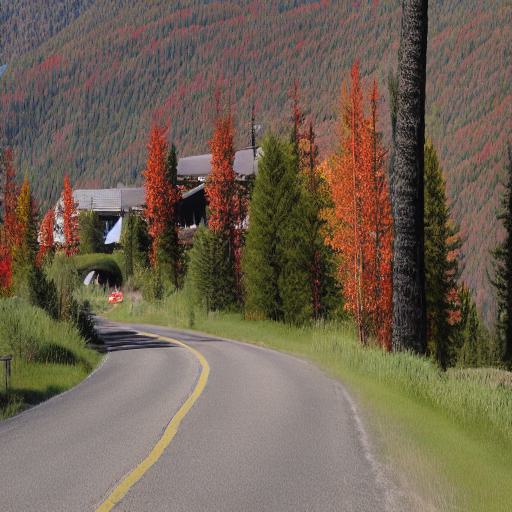}
     \includegraphics[width=\linewidth]{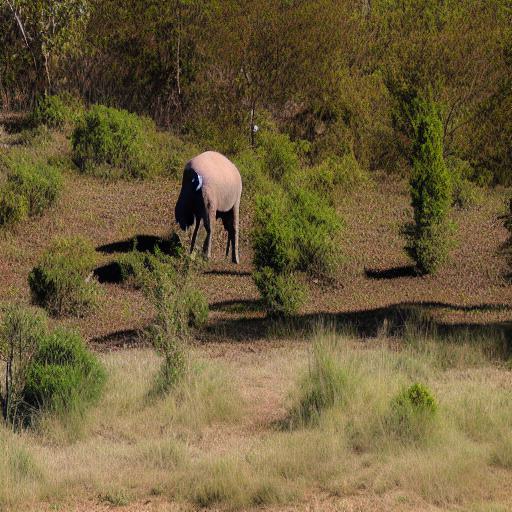}
     \includegraphics[width=\linewidth]{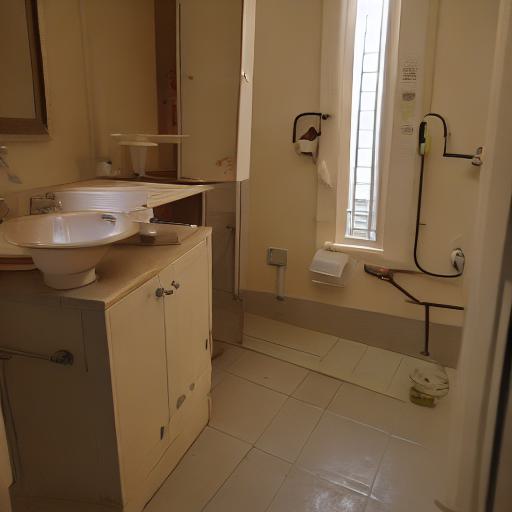}
     \end{minipage}
     }
     \hspace{-2.25mm}
     \subfloat[Sketch \\ +\#1,500]{
     \begin{minipage}{0.1375\linewidth}
     \includegraphics[width=\linewidth]{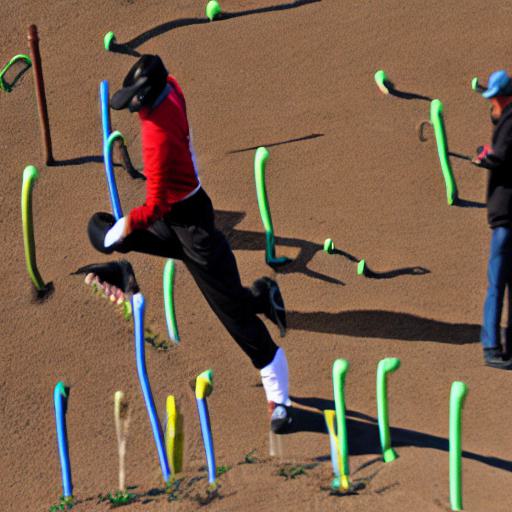}
     \includegraphics[width=\linewidth]{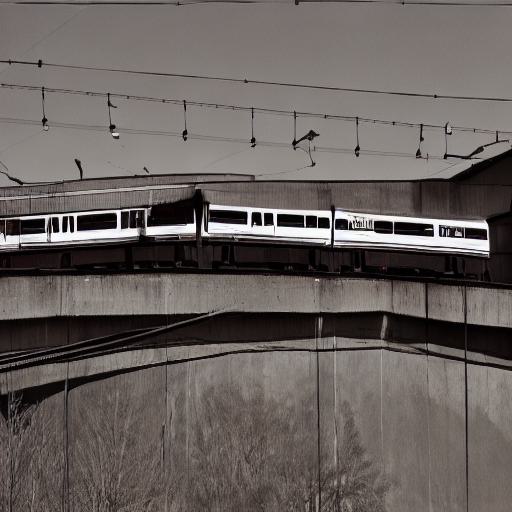}
     \includegraphics[width=\linewidth]{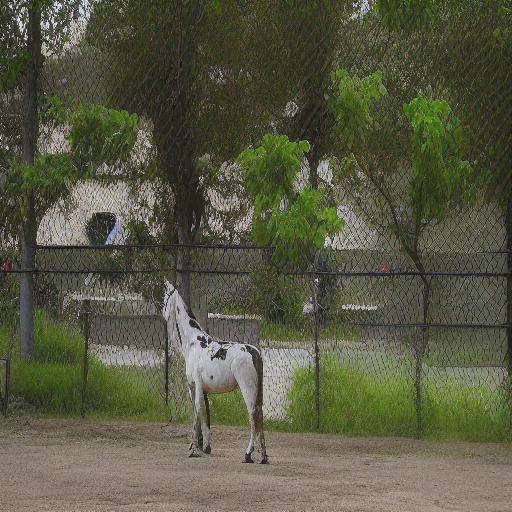}
     \includegraphics[width=\linewidth]{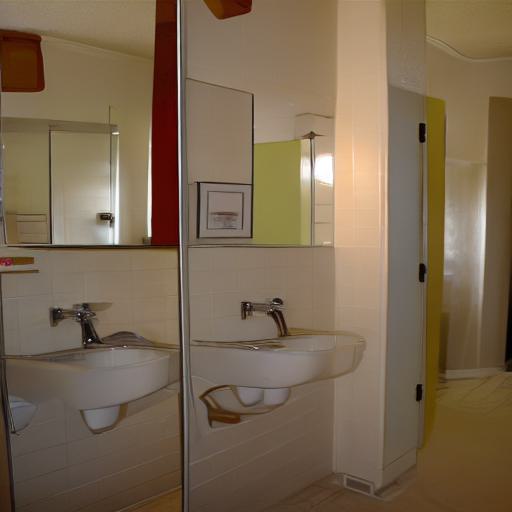}
     \end{minipage}
     }
     \hspace{-2.25mm}
     \subfloat[Adaptation \\ +\#500]{
     \begin{minipage}{0.1375\linewidth}
     \includegraphics[width=\linewidth]{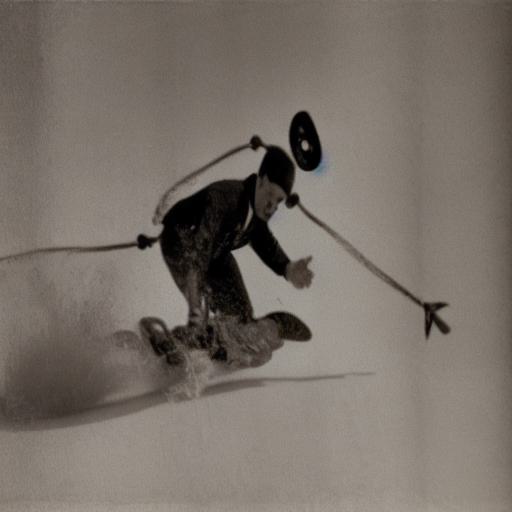}
     \includegraphics[width=\linewidth]{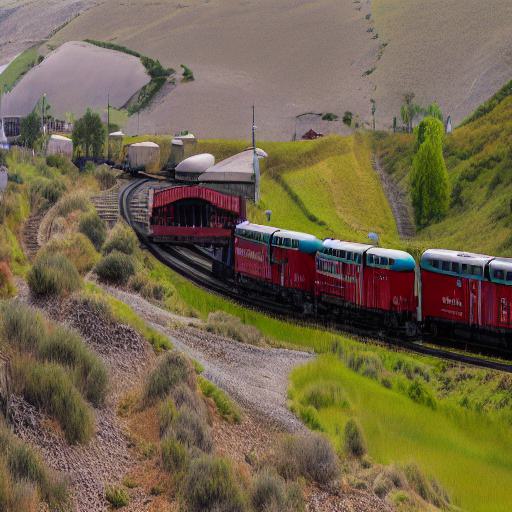}
     \includegraphics[width=\linewidth]{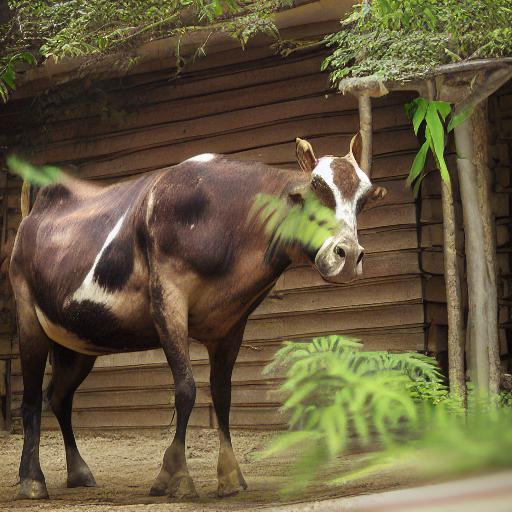}
     \includegraphics[width=\linewidth]{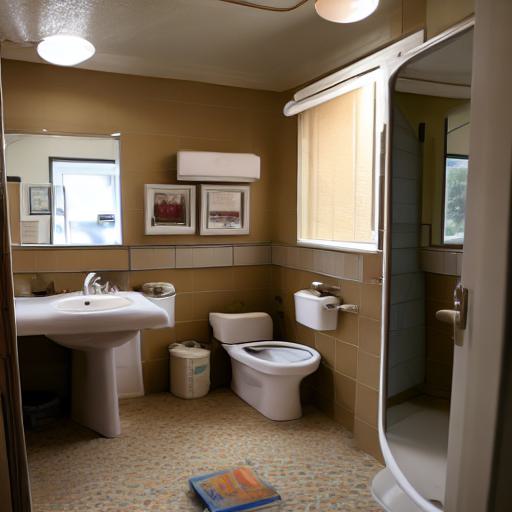}
     \end{minipage}
     }
     \hspace{-2.25mm}
     \subfloat[Adaptation\\ +\#1,500]{
     \begin{minipage}{0.1375\linewidth}
     \includegraphics[width=\linewidth]{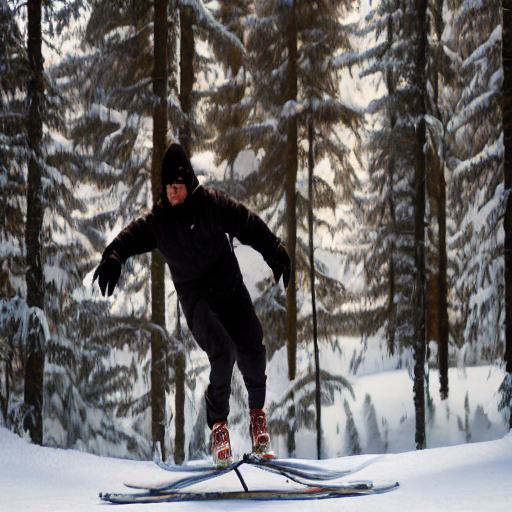}
     \includegraphics[width=\linewidth]{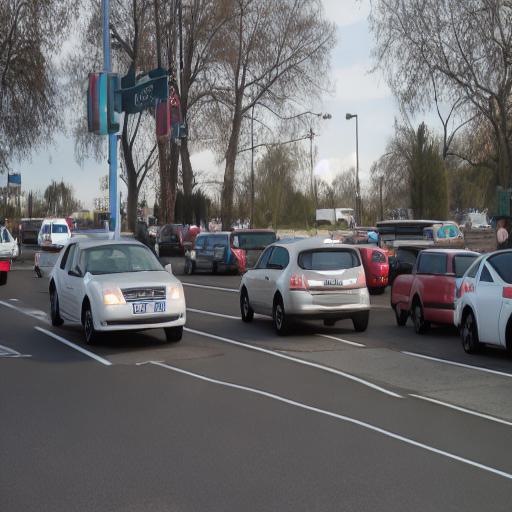}
     \includegraphics[width=\linewidth]{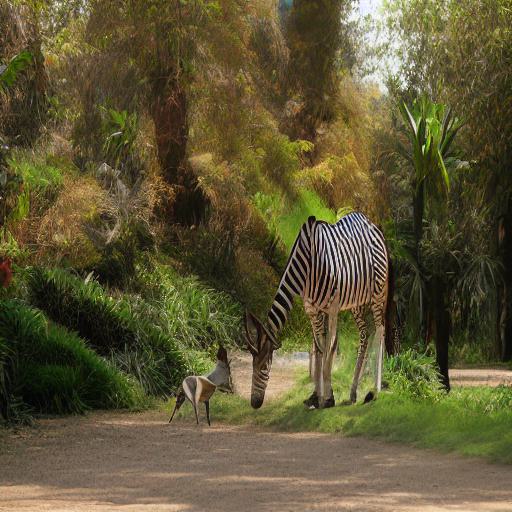}
     \includegraphics[width=\linewidth]{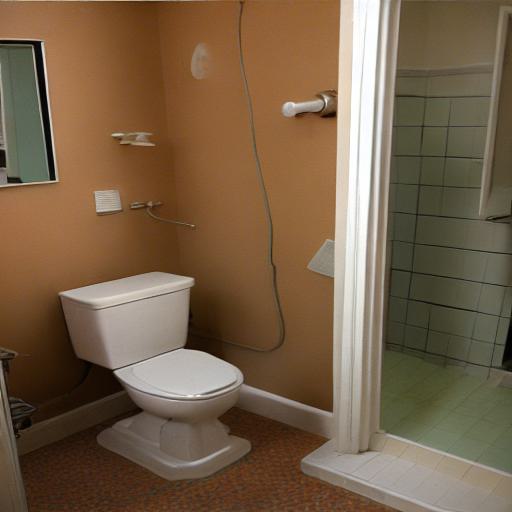}
     \end{minipage}
     }
     \hspace{-2.25mm}
     \subfloat[Stimulus]{
     \begin{minipage}{0.1375\linewidth}
     \includegraphics[width=\linewidth]{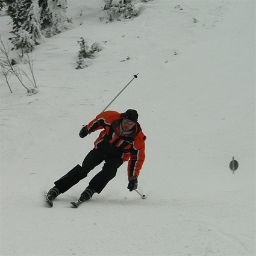}
     \includegraphics[width=\linewidth]{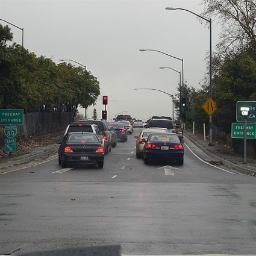}
     \includegraphics[width=\linewidth]{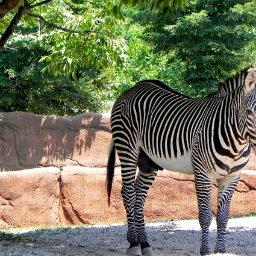}
     \includegraphics[width=\linewidth]{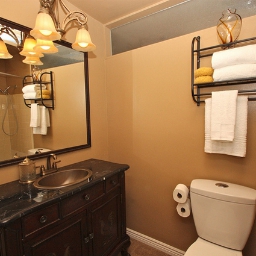}
     \end{minipage}
     }
\vspace{-2mm}
\caption{Qualitative comparison under different data limitation scenarios.}
\vspace{-5mm}
\label{fig:figure_adap_supp}
\end{figure*}

We present a qualitative comparison under different data limitation scenarios in Fig.~\ref{fig:figure_adap_supp}. As the number of training samples increases, the reconstruction quality also improves. Compared to training from sketches with limited data, our adapted method reconstructs the image more faithfully.

\textbf{Synthesis fMRI for Specific Subject}

\begin{figure*}[!t]
    \centering
    \captionsetup[subfloat]{labelformat=empty,justification=centering}
     \subfloat[Subj.1]{
     \begin{minipage}{0.1375\linewidth}
     \includegraphics[width=\linewidth]{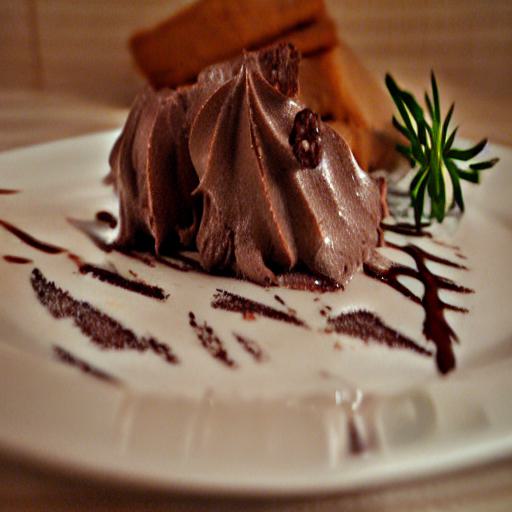}
     \includegraphics[width=\linewidth]{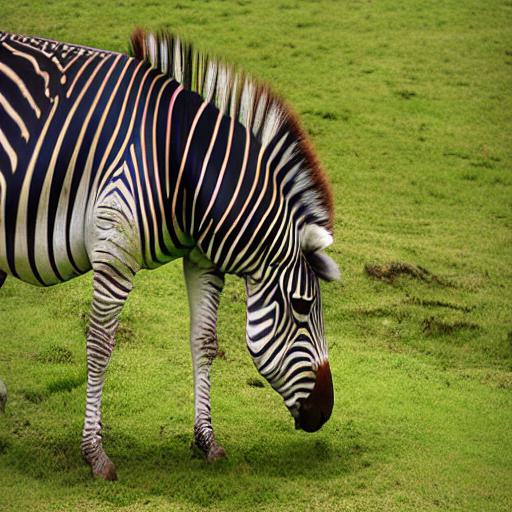}
     \includegraphics[width=\linewidth]{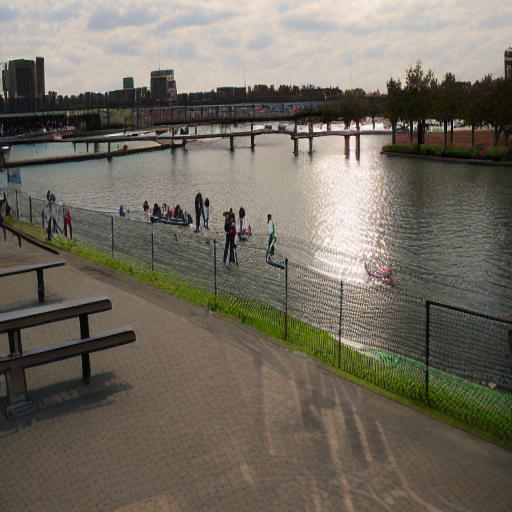}
     \includegraphics[width=\linewidth]{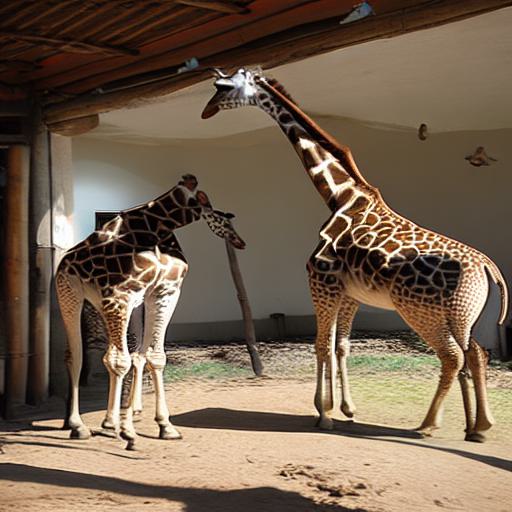}
     \includegraphics[width=\linewidth]{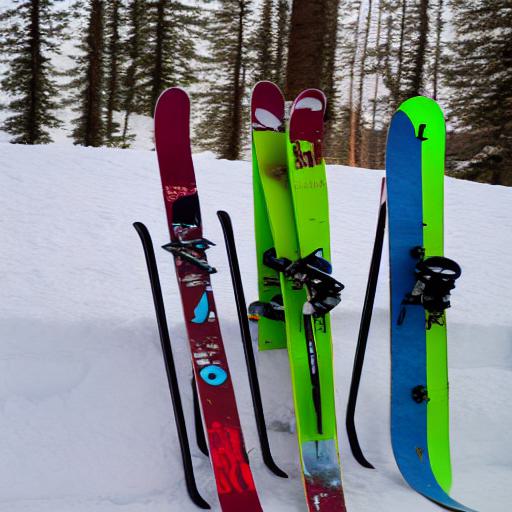}
     \includegraphics[width=\linewidth]{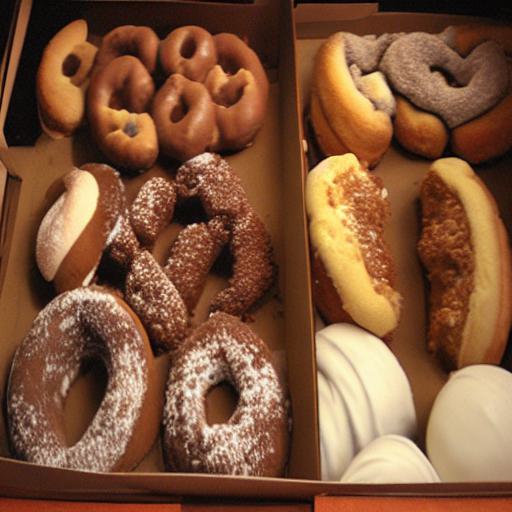}
     \includegraphics[width=\linewidth]{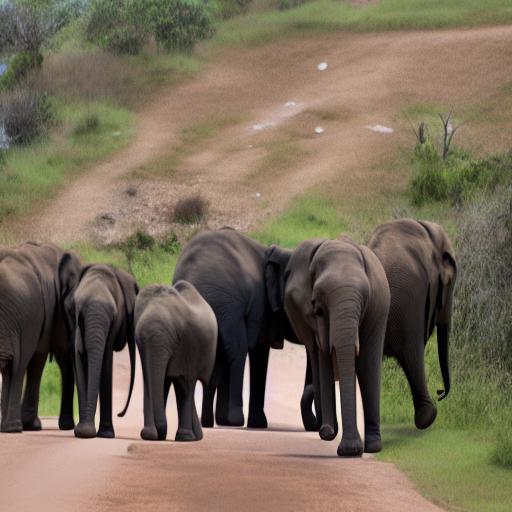}
     \includegraphics[width=\linewidth]{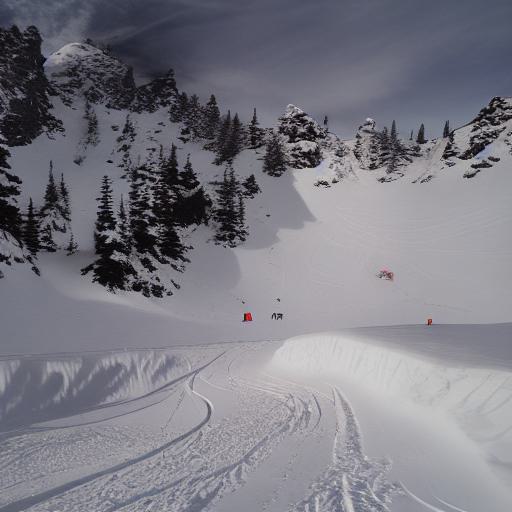}
     \end{minipage}
     }
     \hspace{-2.25mm}
     \subfloat[Subj.2]{
     \begin{minipage}{0.1375\linewidth}
     \includegraphics[width=\linewidth]{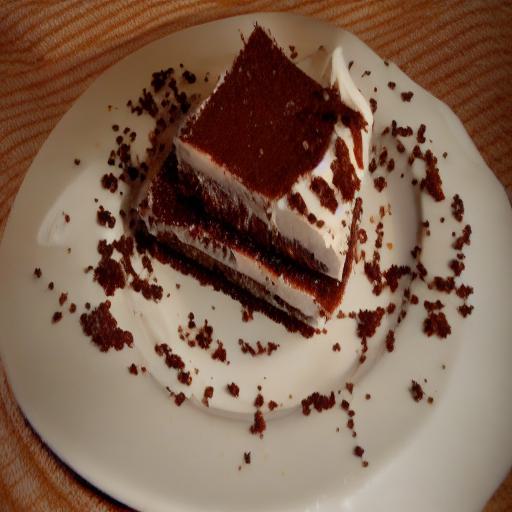}
     \includegraphics[width=\linewidth]{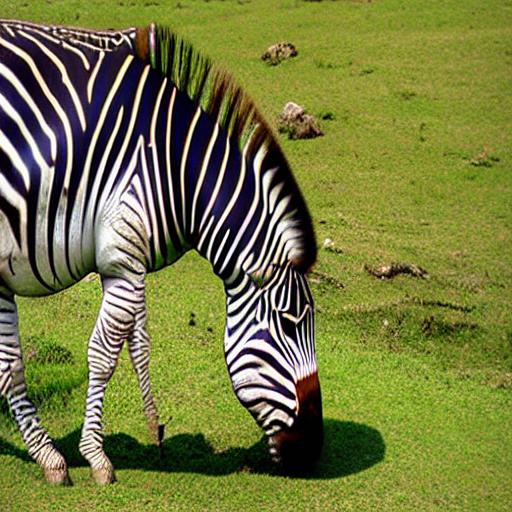}
     \includegraphics[width=\linewidth]{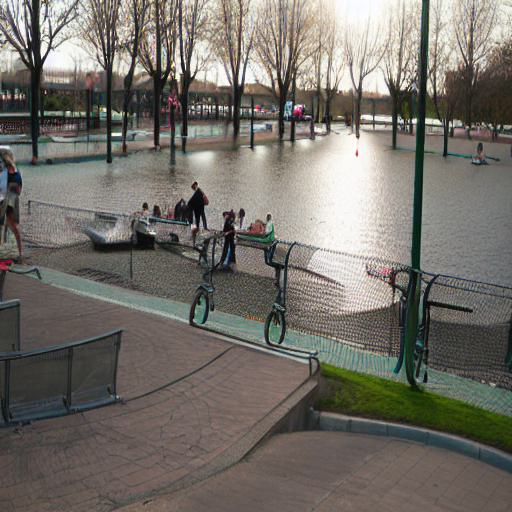}
     \includegraphics[width=\linewidth]{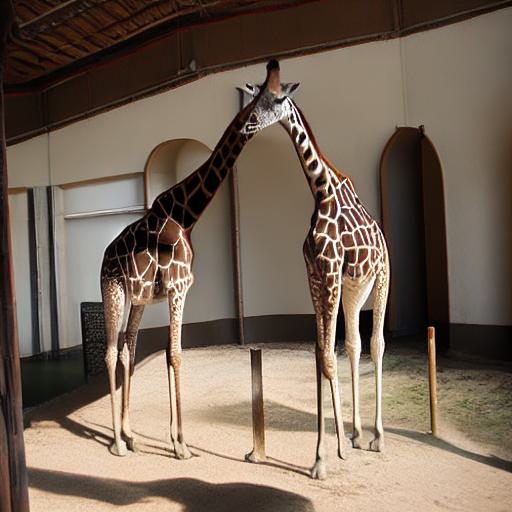}
     \includegraphics[width=\linewidth]{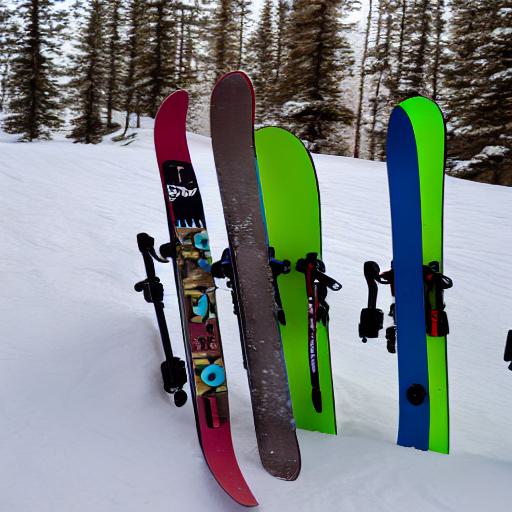}
     \includegraphics[width=\linewidth]{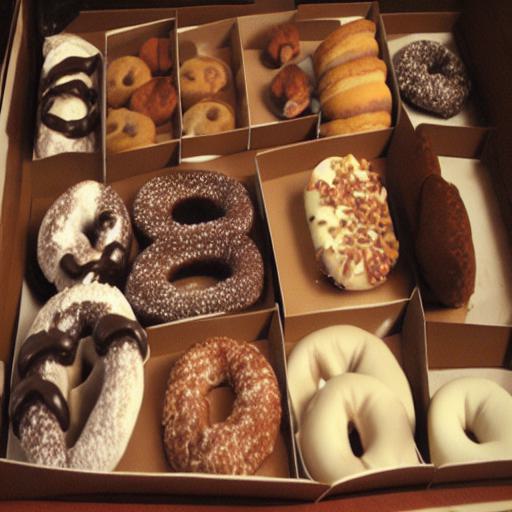}
     \includegraphics[width=\linewidth]{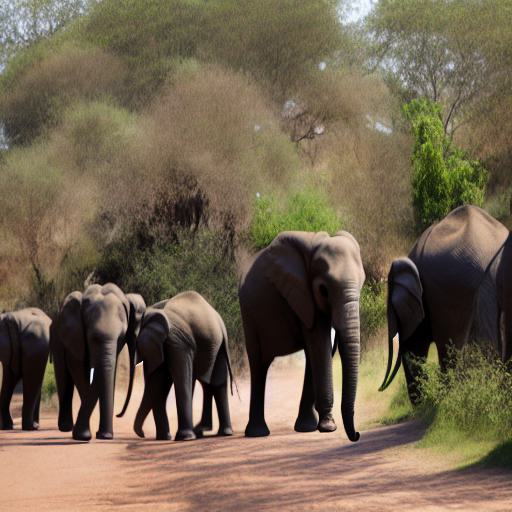}
     \includegraphics[width=\linewidth]{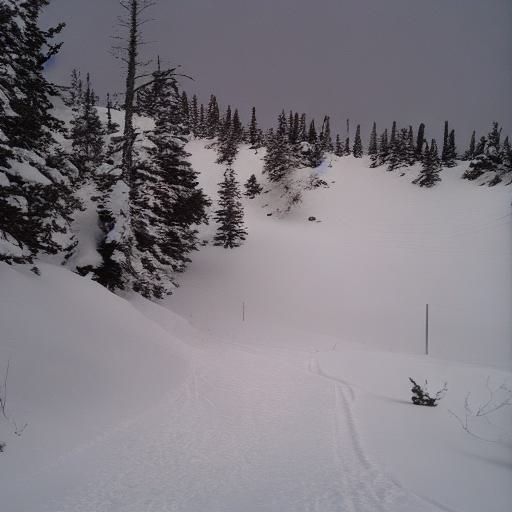}
     \end{minipage}
     }
     \hspace{-2.25mm}
     \subfloat[Subj.5]{
     \begin{minipage}{0.1375\linewidth}
     \includegraphics[width=\linewidth]{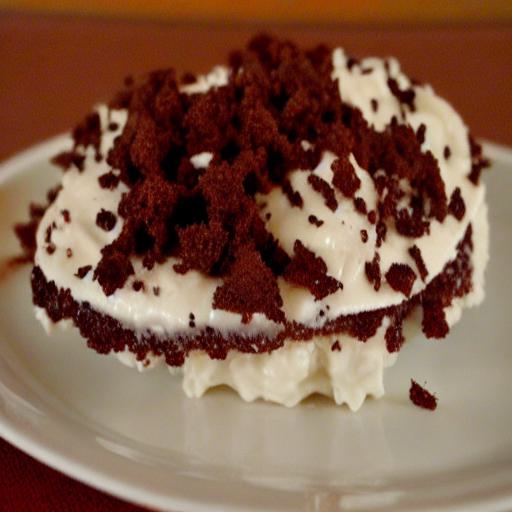}
     \includegraphics[width=\linewidth]{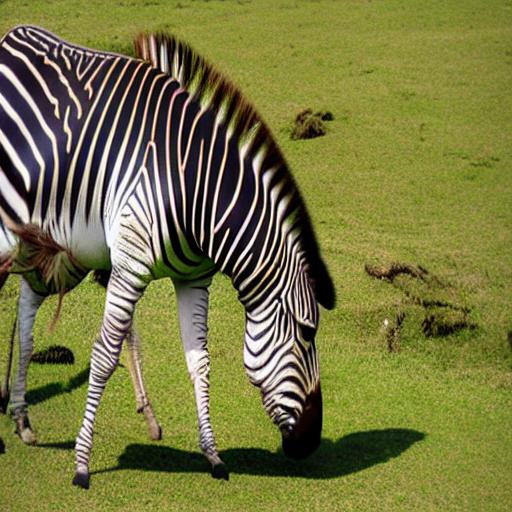}
     \includegraphics[width=\linewidth]{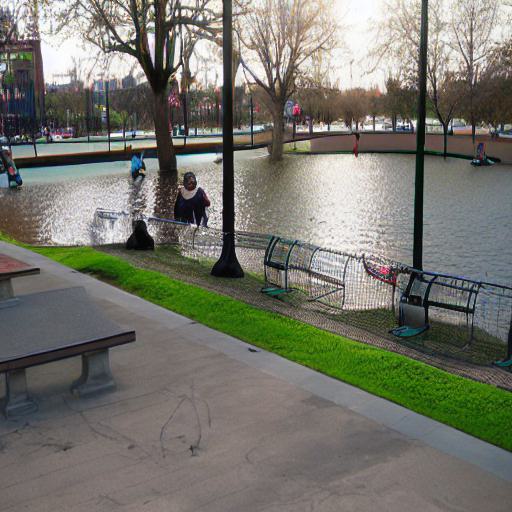}
     \includegraphics[width=\linewidth]{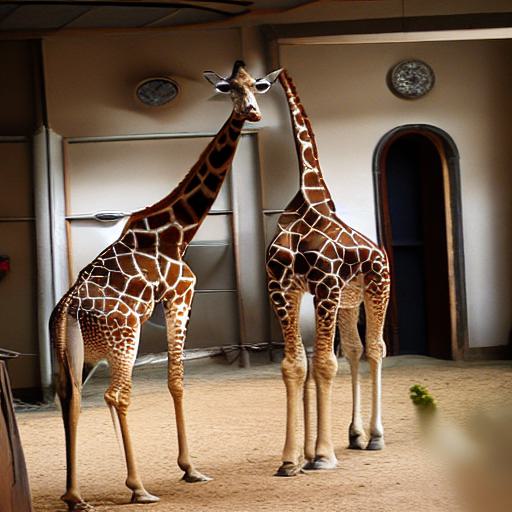}
     \includegraphics[width=\linewidth]{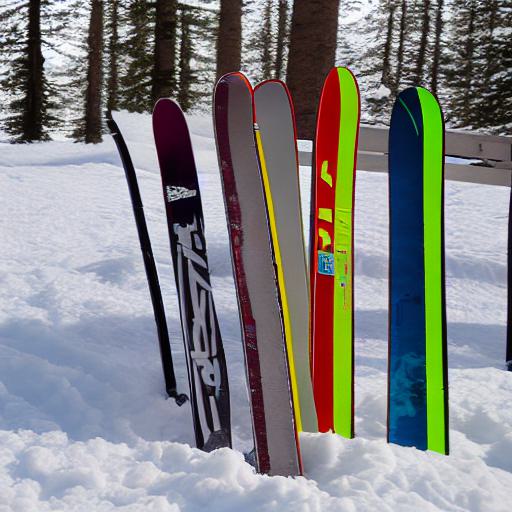}
     \includegraphics[width=\linewidth]{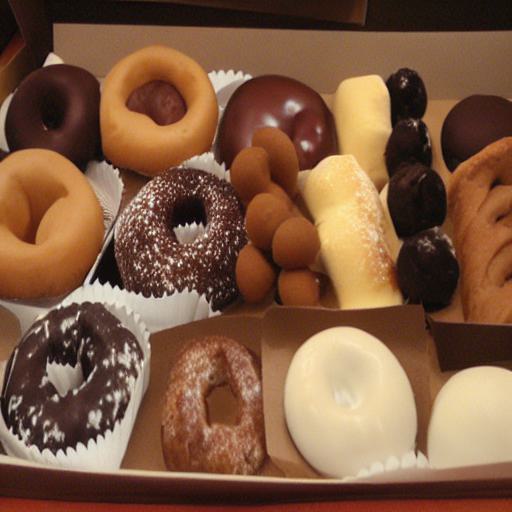}
     \includegraphics[width=\linewidth]{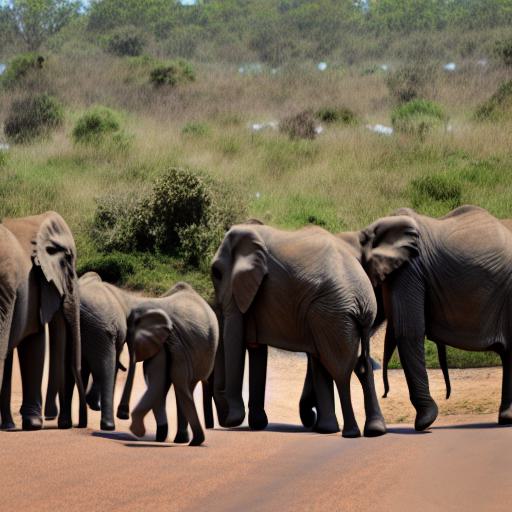}
     \includegraphics[width=\linewidth]{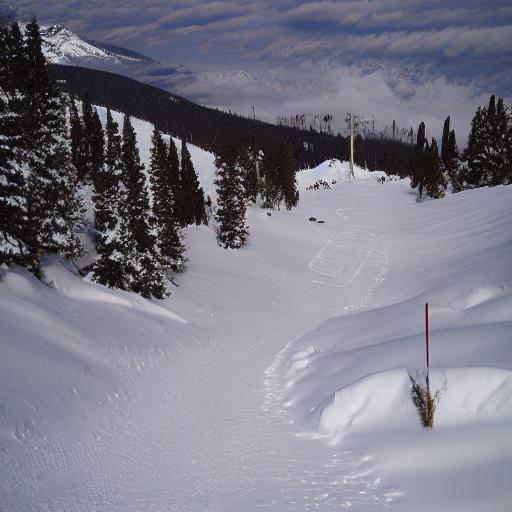}
     \end{minipage}
     }
     \hspace{-2.25mm}
     \subfloat[Subj.7]{
     \begin{minipage}{0.1375\linewidth}
     \includegraphics[width=\linewidth]{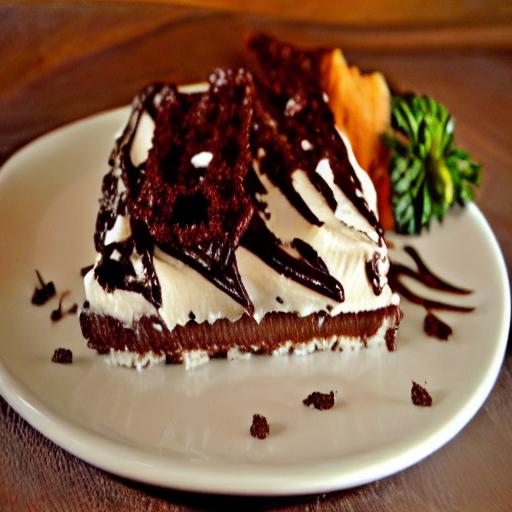}
     \includegraphics[width=\linewidth]{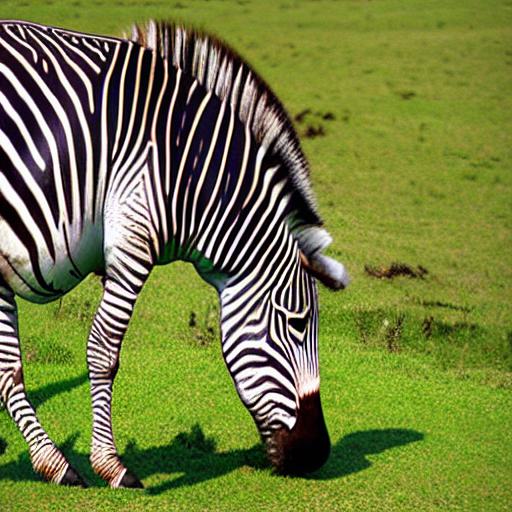}
     \includegraphics[width=\linewidth]{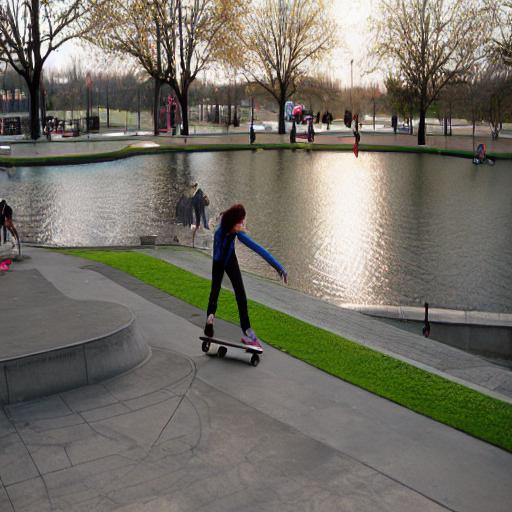}
     \includegraphics[width=\linewidth]{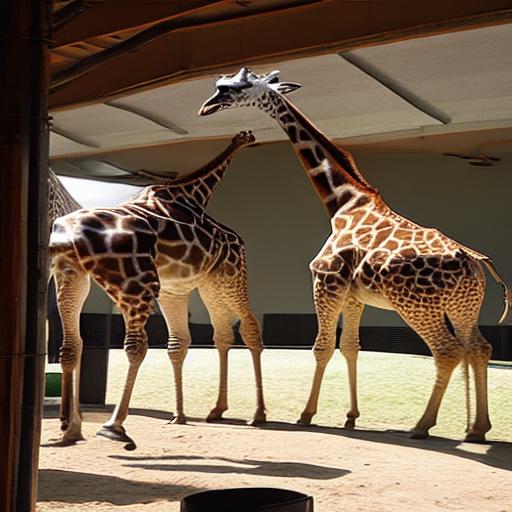}
     \includegraphics[width=\linewidth]{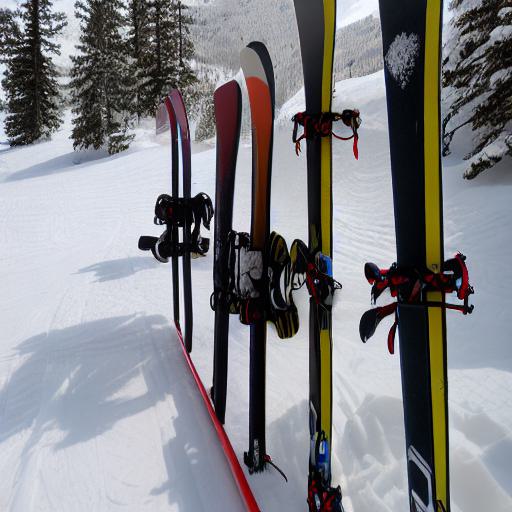}
     \includegraphics[width=\linewidth]{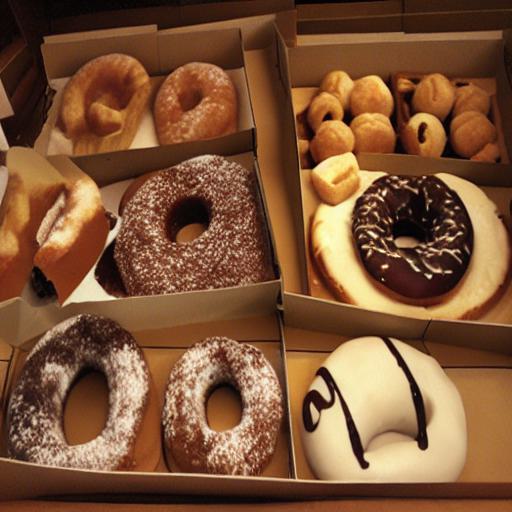}
     \includegraphics[width=\linewidth]{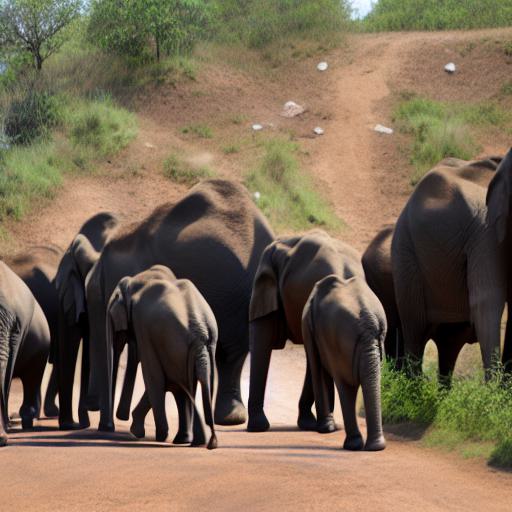}
     \includegraphics[width=\linewidth]{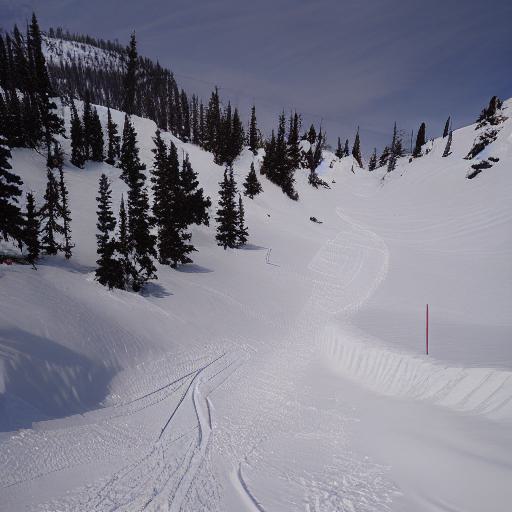}
     \end{minipage}
     }
     \hspace{-2.25mm}
     \subfloat[Stimulus]{
     \begin{minipage}{0.1375\linewidth}
     \includegraphics[width=\linewidth]{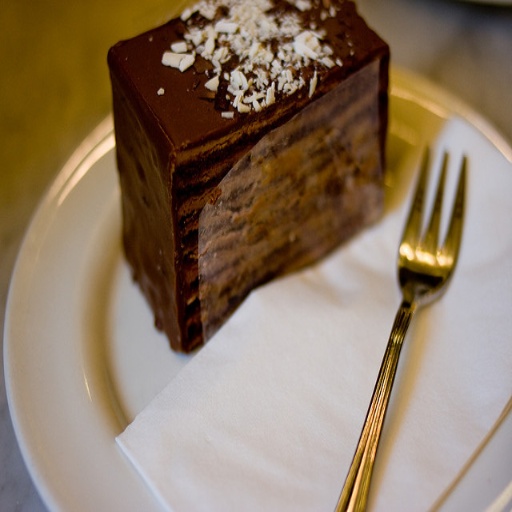}
     \includegraphics[width=\linewidth]{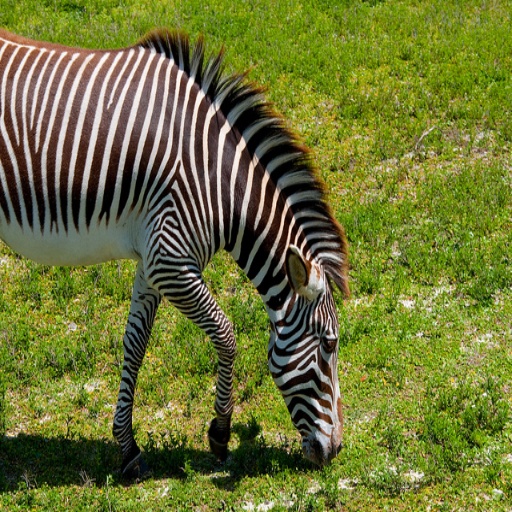}
     \includegraphics[width=\linewidth]{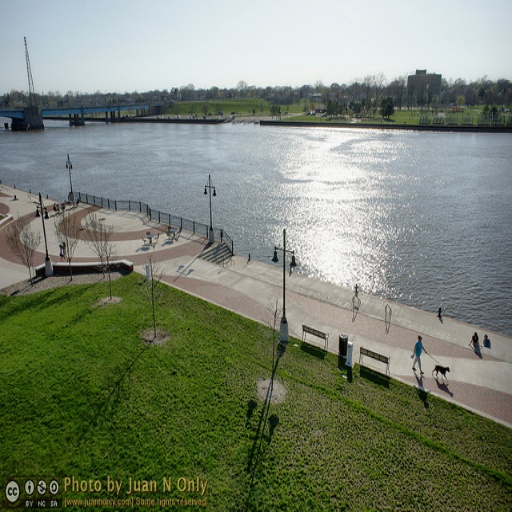}
     \includegraphics[width=\linewidth]{figures_cycle_supp_gt_000000000136.jpg}
     \includegraphics[width=\linewidth]{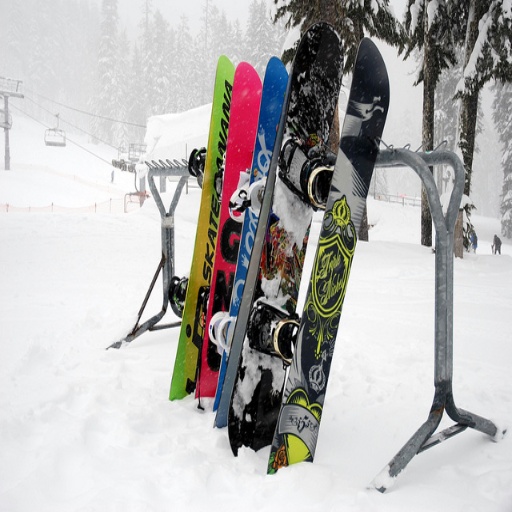}
     \includegraphics[width=\linewidth]{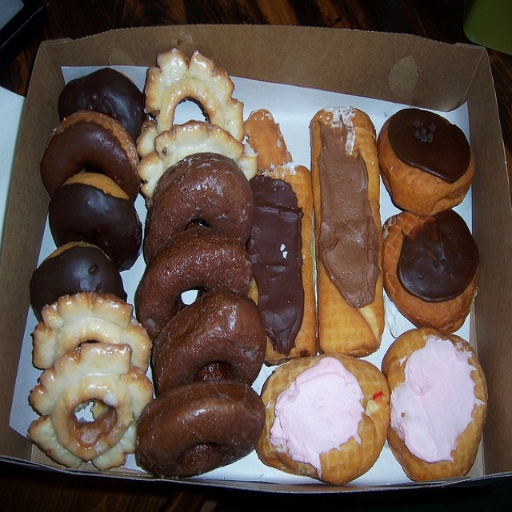}
     \includegraphics[width=\linewidth]{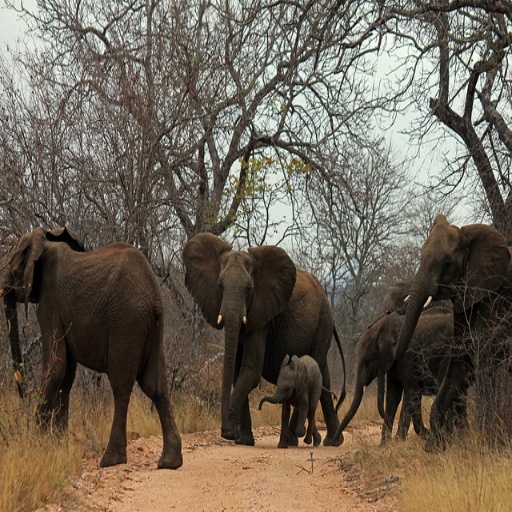}
     \includegraphics[width=\linewidth]{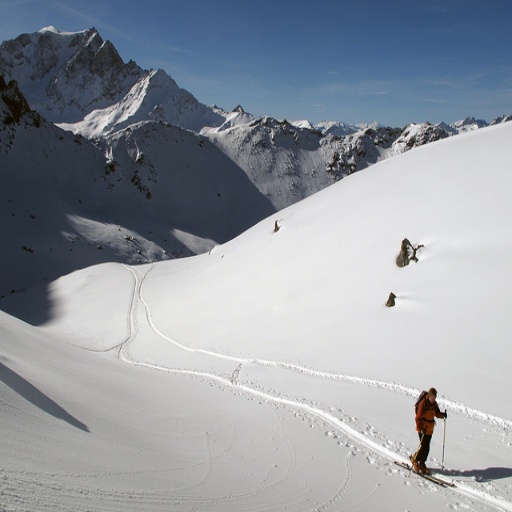}
     \end{minipage}
     }
\vspace{-2mm}
\caption{Our framework also supports the synthesis of fMRI for specific subject based on an unseen image, and the synthesized fMRI voxels can reconstruct the stimulus image faithfully.}
\vspace{-5mm}
\label{fig:figure_cycle_supp}
\end{figure*}

Given an unseen stimulus image, our framework mimics the visual system by synthesizing the corresponding fMRI for a specific subject: $\{S, E, C\} \Rightarrow \hat{V}_{x}$. We then decode the synthesized fMRI 
voxels into representations: $\hat{V}_{x} \Rightarrow \{\doublehat{S}, \doublehat{E}, \doublehat{C}\}$. The reconstructed images are shown in Fig.~\ref{fig:figure_cycle_supp}, where the synthesized fMRI faithfully reconstructs the stimulus image.

\textbf{Comparison with MindBridge on Cross Subject Mind Decoding} 

\begin{figure*}[h]
    \centering
    \captionsetup[subfloat]{labelformat=empty,justification=centering}
    \vspace{-4mm}
    \subfloat[Subj.1]{
     \begin{minipage}{0.12\linewidth} % 原0.1375→0.164
     \includegraphics[width=\linewidth]{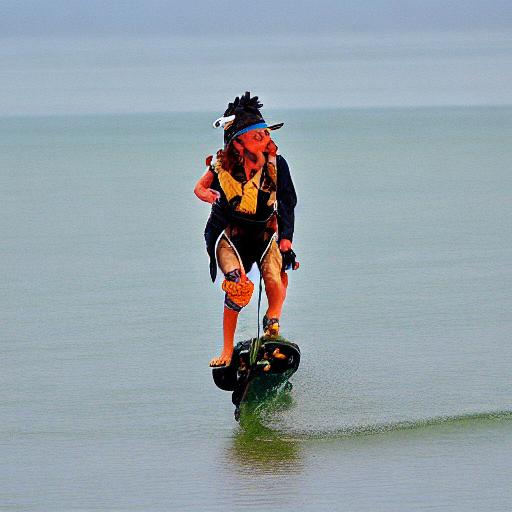}
     \includegraphics[width=\linewidth]{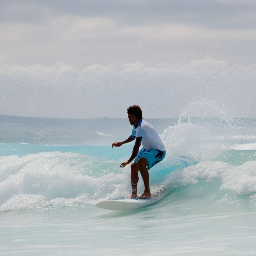}
     \includegraphics[width=\linewidth]{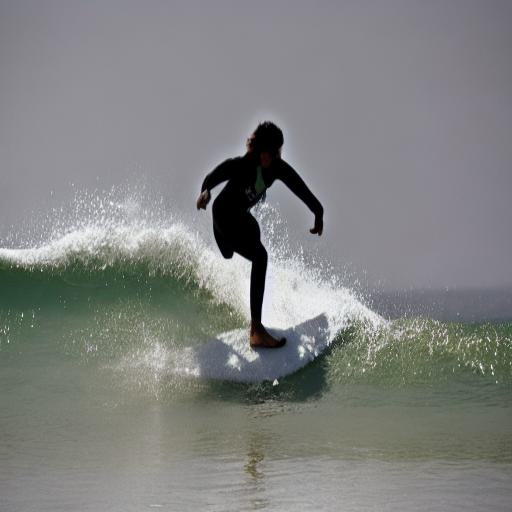}
     \includegraphics[width=\linewidth]{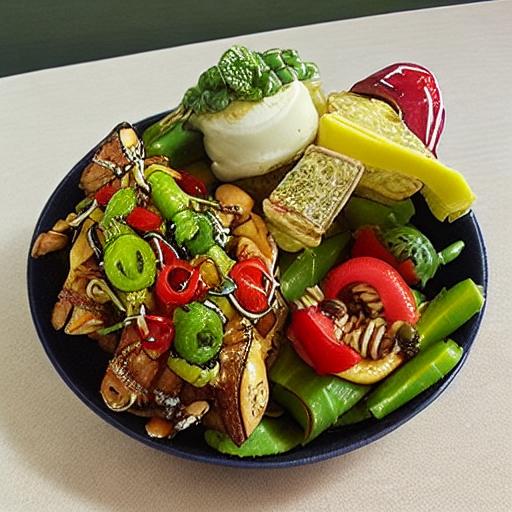}
     \includegraphics[width=\linewidth]{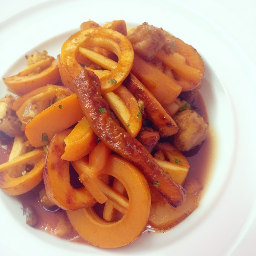}
     \includegraphics[width=\linewidth]{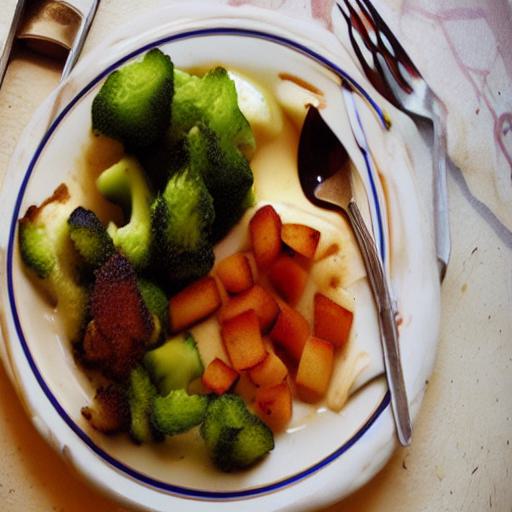}
     \includegraphics[width=\linewidth]{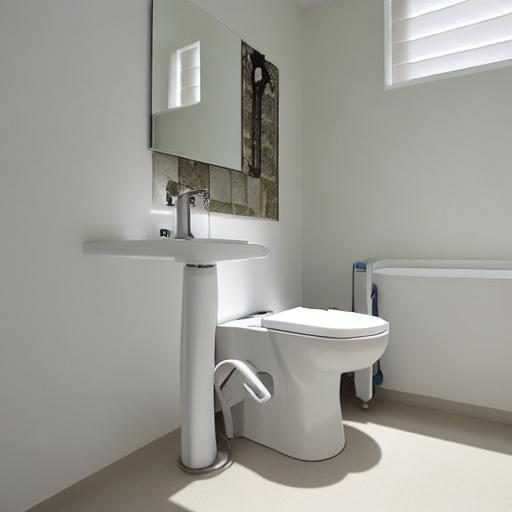}
     \includegraphics[width=\linewidth]{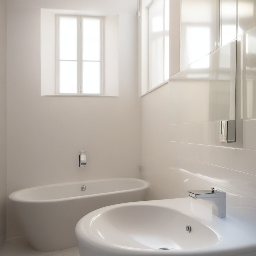}
     \includegraphics[width=\linewidth]{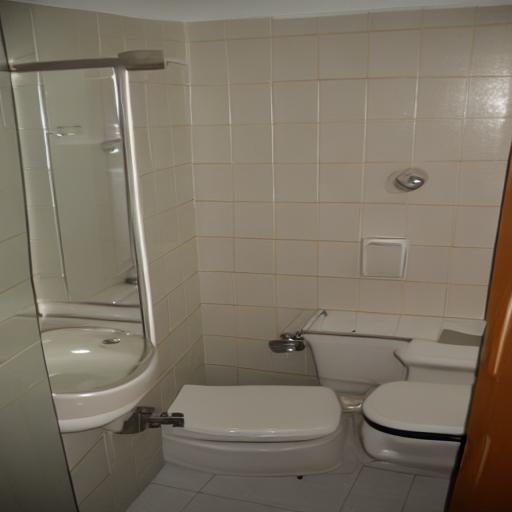}
     \end{minipage}
     }
     \hspace{-3.5mm}
     \subfloat[Subj.2]{
     \begin{minipage}{0.12\linewidth}
     \includegraphics[width=\linewidth]{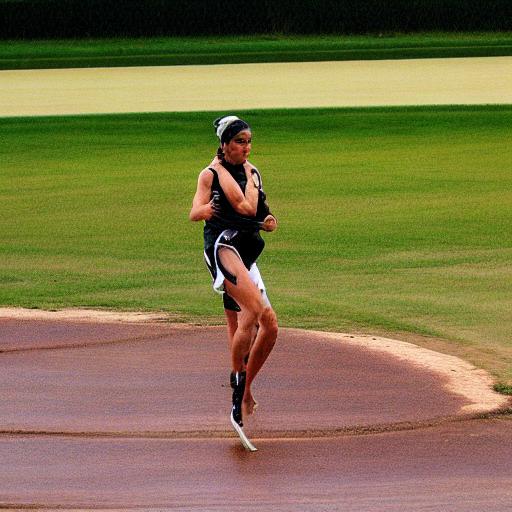}
     \includegraphics[width=\linewidth]{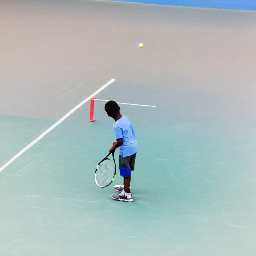}
     \includegraphics[width=\linewidth]{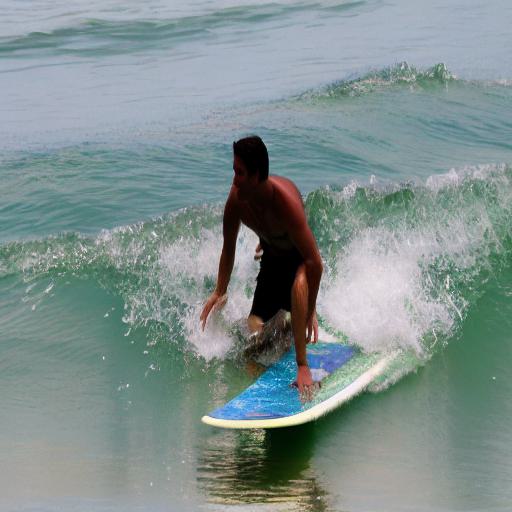}
     \includegraphics[width=\linewidth]{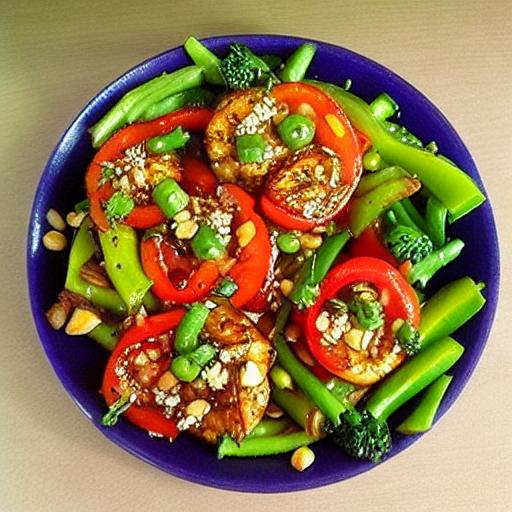}
     \includegraphics[width=\linewidth]{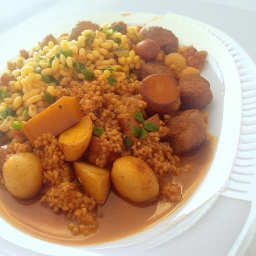}
     \includegraphics[width=\linewidth]{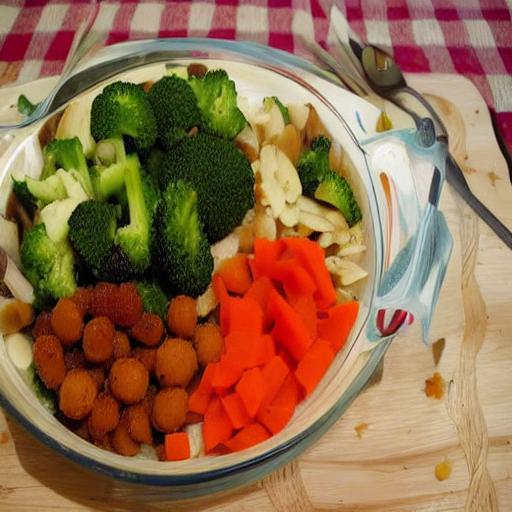}
     \includegraphics[width=\linewidth]{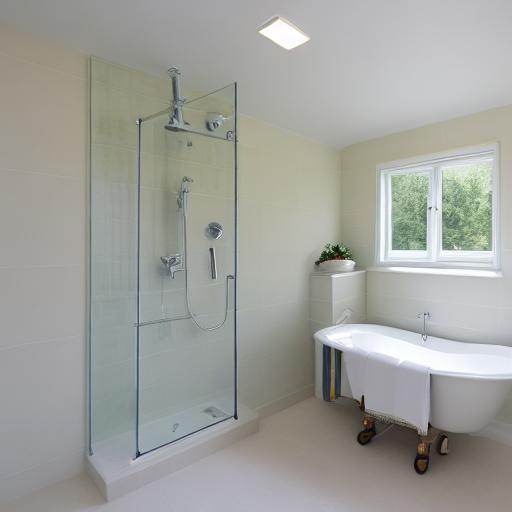}
     \includegraphics[width=\linewidth]{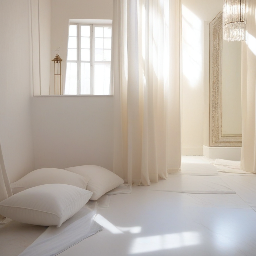}
     \includegraphics[width=\linewidth]{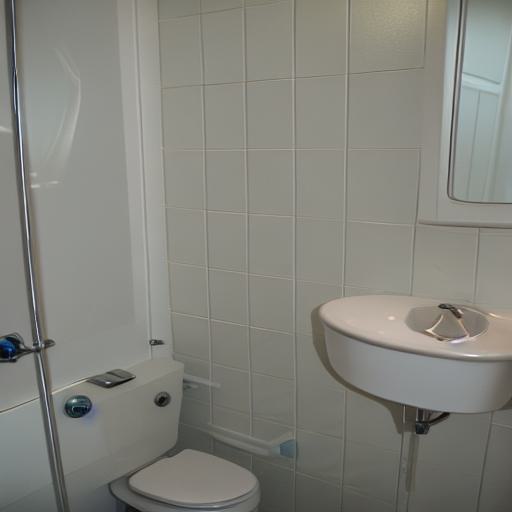}
     \end{minipage}
     }
     \hspace{-3.5mm}
     \subfloat[Subj.5]{
     \begin{minipage}{0.12\linewidth}
     \includegraphics[width=\linewidth]{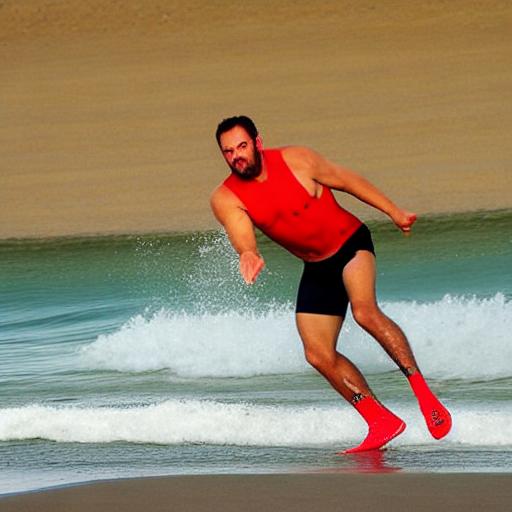}
     \includegraphics[width=\linewidth]{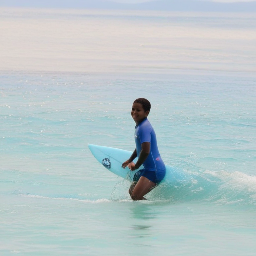}
     \includegraphics[width=\linewidth]{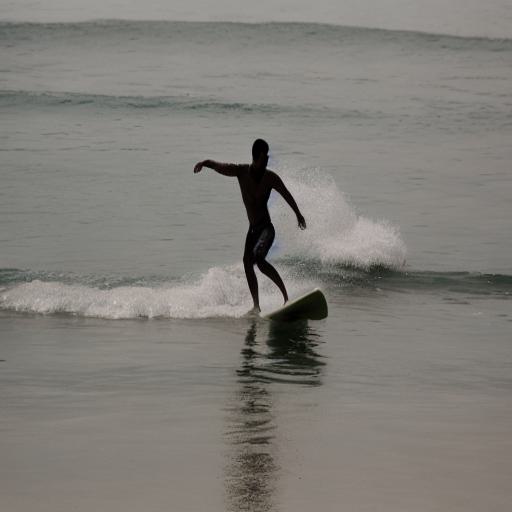}
     \includegraphics[width=\linewidth]{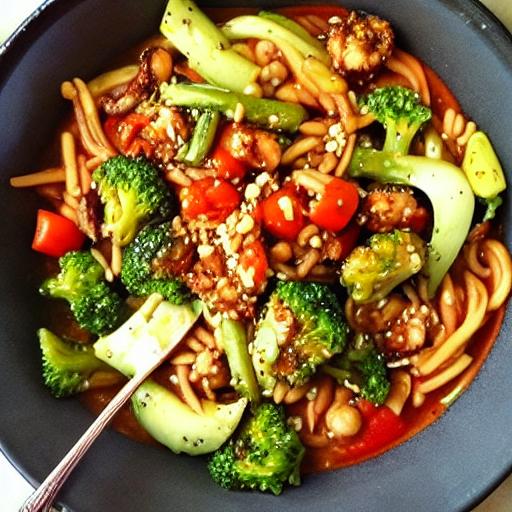}
     \includegraphics[width=\linewidth]{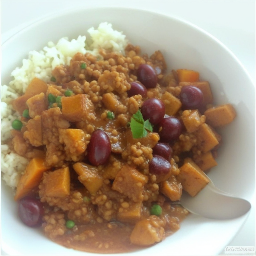}
     \includegraphics[width=\linewidth]{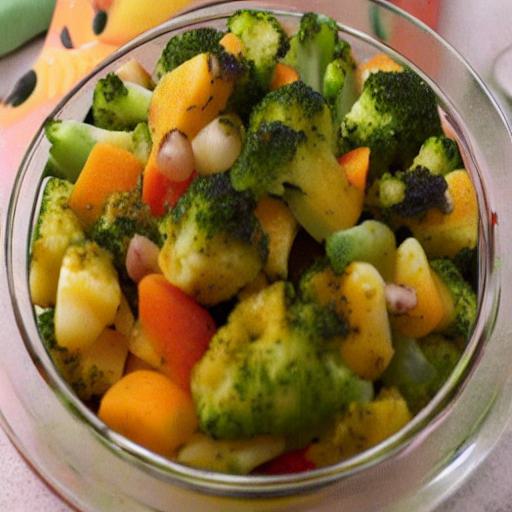}
     \includegraphics[width=\linewidth]{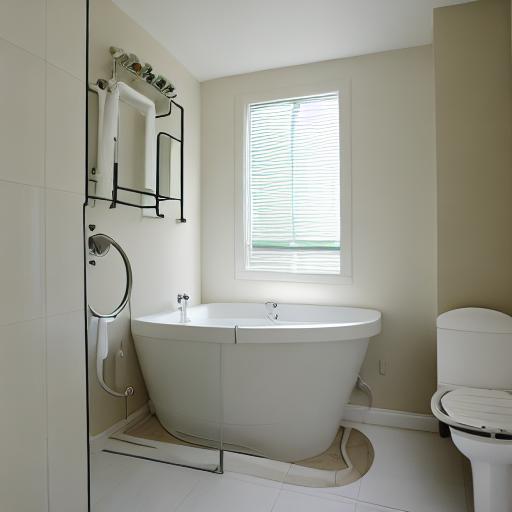}
     \includegraphics[width=\linewidth]{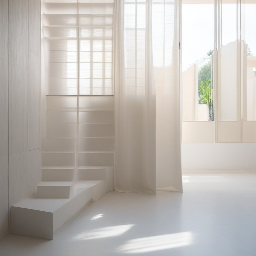}
     \includegraphics[width=\linewidth]{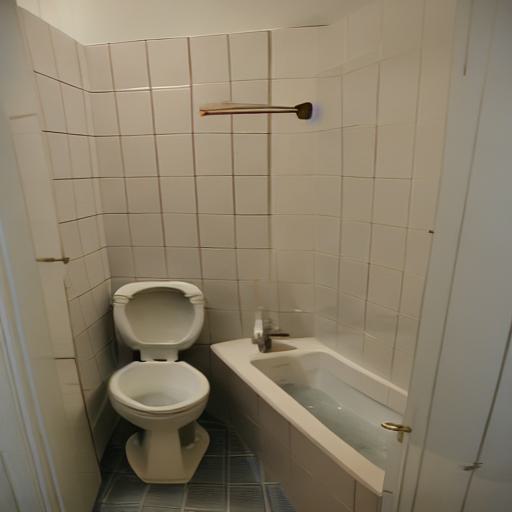}
     \end{minipage}
     }
     \hspace{-3.5mm}
     \subfloat[Subj.7]{
     \begin{minipage}{0.12\linewidth}
     \includegraphics[width=\linewidth]{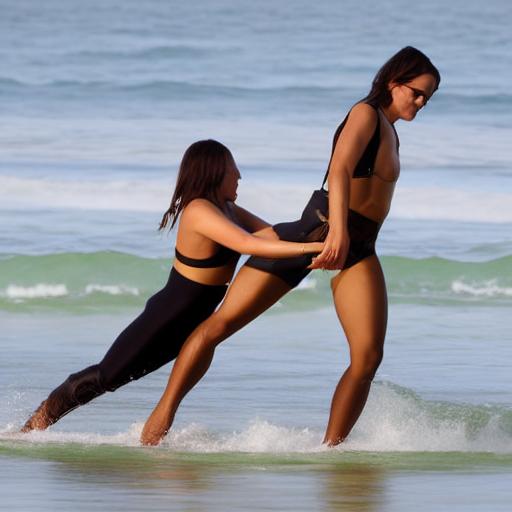}
     \includegraphics[width=\linewidth]{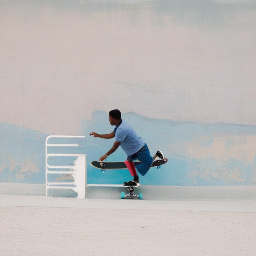}
     \includegraphics[width=\linewidth]{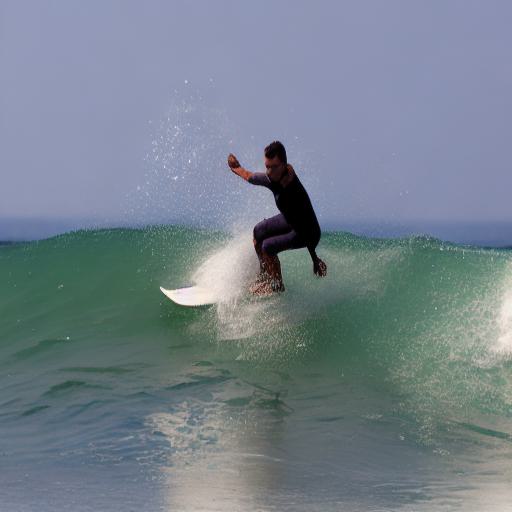}
     \includegraphics[width=\linewidth]{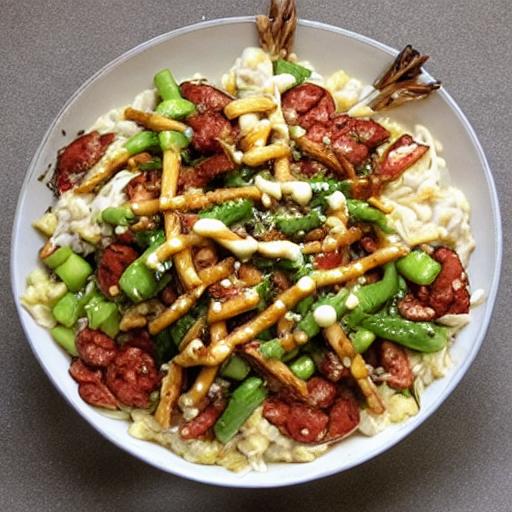}
     \includegraphics[width=\linewidth]{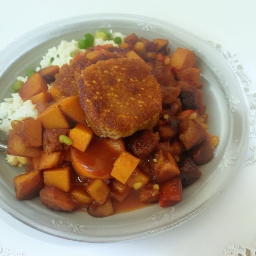}
     \includegraphics[width=\linewidth]{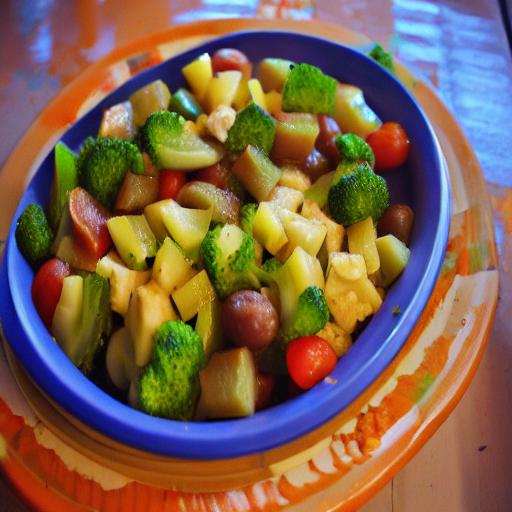}
     \includegraphics[width=\linewidth]{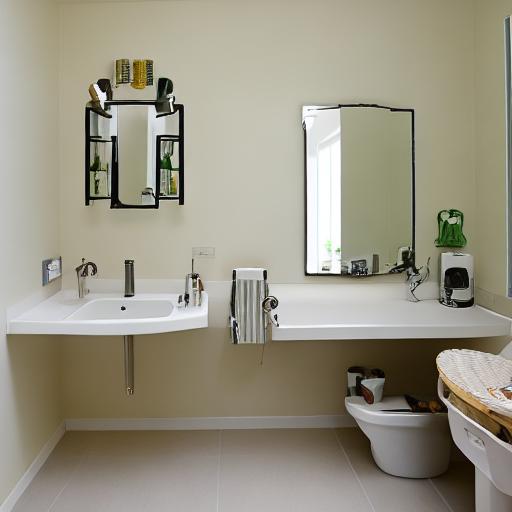}
     \includegraphics[width=\linewidth]{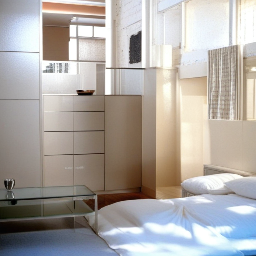}
     \includegraphics[width=\linewidth]{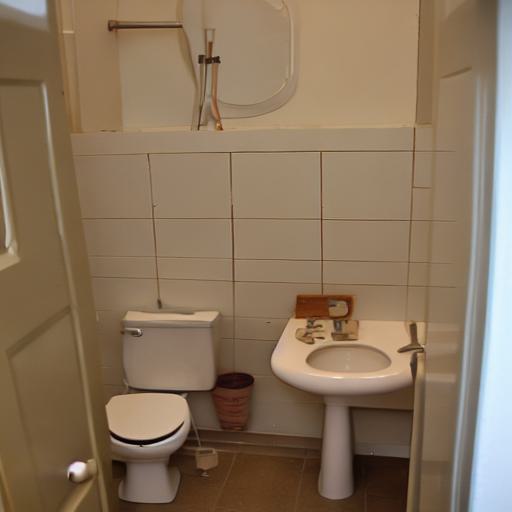}
     \end{minipage}
     }
     \hspace{-3.5mm}
     \subfloat[Stimulus]{
     \begin{minipage}{0.12\linewidth}
     \includegraphics[width=\linewidth]{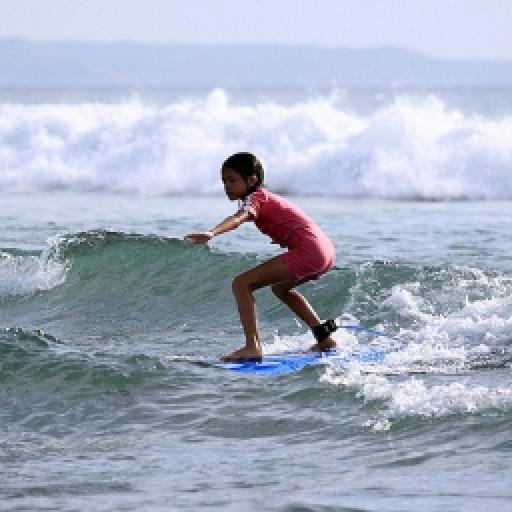}
     \includegraphics[width=\linewidth]{figures_rebuttal_cross_sub_GT_sample000000416.jpg.jpg}
     \includegraphics[width=\linewidth]{figures_rebuttal_cross_sub_GT_sample000000416.jpg.jpg}
     \includegraphics[width=\linewidth]{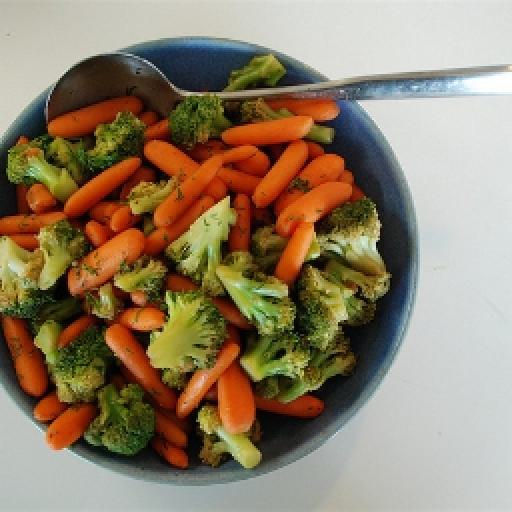}
     \includegraphics[width=\linewidth]{figures_rebuttal_cross_sub_GT_sample000001302.jpg.jpg}
     \includegraphics[width=\linewidth]{figures_rebuttal_cross_sub_GT_sample000001302.jpg.jpg}
     \includegraphics[width=\linewidth]{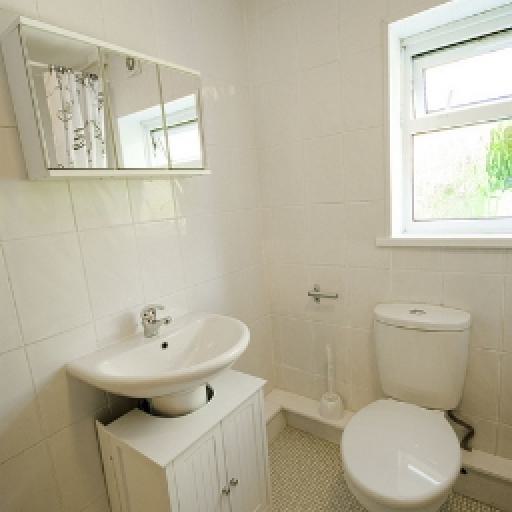}
     \includegraphics[width=\linewidth]{figures_rebuttal_cross_sub_GT_sample000004567.jpg.jpg}
     \includegraphics[width=\linewidth]{figures_rebuttal_cross_sub_GT_sample000004567.jpg.jpg}
     \end{minipage}
     }
     \rotatebox[origin=c]{270}{\hspace{-19cm}\scriptsize{MindEye2} \hspace{10mm} \scriptsize{Mindbridge} \hspace{10mm} \scriptsize{Ours}\hspace{15mm}\scriptsize{MindEye2} \hspace{10mm} \scriptsize{Mindbridge} \hspace{10mm} \scriptsize{Ours}\hspace{15mm}\scriptsize{MindEye2} \hspace{10mm} \scriptsize{Mindbridge} \hspace{12mm} \scriptsize{Ours}}
\caption{Comparison on Cross-subject Mind decoding.}
\label{xsubject}
\end{figure*}

We present a cross-subject comparison with MindBridge and MindEye2 in Fig.~\ref{xsubject}. Our decoded images exhibit greater consistency with the stimulus image across different subjects. For instance, both MindBridge and MindEye2 fail to decode the \textit{``Broccoli''} in the second sample, whereas our method successfully reconstructs it.

\end{document}